\documentclass[multiauthors,onecolumn]{LaTeXclass/cohere}
\usepackage{ragged2e}
\usepackage{lipsum}
\usepackage{pdfpages}
\usepackage{colortbl}
\usepackage{tabulary}
\usepackage{longtable}
\usepackage{array}
\usepackage{placeins}
\usepackage{tablefootnote}
\usepackage{comment}
\usepackage{xcolor}
\usepackage{tcolorbox}
\usepackage{wrapfig}
\usepackage{breqn} %
\usepackage{booktabs}
\usepackage{multirow}

\usepackage{pifont}
\definecolor{deeppurple}{HTML}{9e02f7}
\definecolor{forestgreen}{HTML}{2e7d43}
\usepackage{multirow}
\usepackage{tabularx}
\usepackage{ragged2e} %
\newcolumntype{Y}{>{\RaggedRight\arraybackslash}X} %

\usepackage{cleveref}
\crefname{section}{§}{§§}
\Crefname{section}{§}{§§}
\usepackage{wrapfig}
\usepackage{tcolorbox}
\usepackage{amsmath,amssymb,amsfonts,mathtools,amsthm}
\usepackage{mdframed}
\usepackage{fontspec}
\usepackage{fancyvrb}
\usepackage{fvextra}
\usepackage{float} 
\usepackage{comment}
\usepackage{xcolor}
\usepackage{tcolorbox}
\usepackage{kotex} %
\usepackage{url}
\usepackage[table]{xcolor}
\usepackage{booktabs}
\usepackage{microtype} 
\usepackage{makecell}

\newtcolorbox{promptbox}[1]{
    enhanced,
    breakable,
    colback=gray!5,
    colframe=gray!70,
    title={#1},
    width=\textwidth,
    fontupper=\fontsize{9pt}{12pt},
    fonttitle=\bfseries,
    boxrule=0.5pt,
    arc=1pt
}

\definecolor{lightgray}{gray}{0.45}

\definecolor{sectiongray}{gray}{0.2}
\definecolor{boxgray}{gray}{0.25}
\definecolor{lightgray}{gray}{0.65}

\definecolor{HeaderBG}{RGB}{35,35,35}   %
\definecolor{HeaderFG}{RGB}{255,255,255}
\definecolor{FrameGray}{gray}{0.25}
\definecolor{DarkBlue}{HTML}{2D4CB9}
\newcommand{\tinyaya}{\textsc{Tiny Aya}}
\newcommand{\tinyayamulti}{\textsc{Tiny Aya L2-Thinker}}
\newcommand{\tinyayaen}{\textsc{Tiny Aya En-Thinker}}

\newcommand{\deepseek}{\textsc{DeepSeek-V3}}

\newcommand{\magistralsmall}{\textsc{Magistral-Small-24B}}
\newcommand{\mthinker}{\textsc{M-Thinker-7B}}

\newcommand{\qwensmallnew}{\textsc{Qwen3.5-4B}}

\newcommand{\commandtranslate}{\textsc{command-a-translate}}
\newcommand{\gptoss}{\textsc{gpt-oss-120b}}
\newcommand{\commandaplus}{\textsc{Command A+}}

\newcommand{\mif}{Marco-Bench-MIF}
\newcommand{\mist}{MIST-OEG}
\newcommand{\macaron}{Macaron-MCQ}

\definecolor{linkslate}{HTML}{3A6E7E}  %
\newcommand{\hflogo}{\raisebox{-0.22\height}{\includegraphics[height=1.15em]{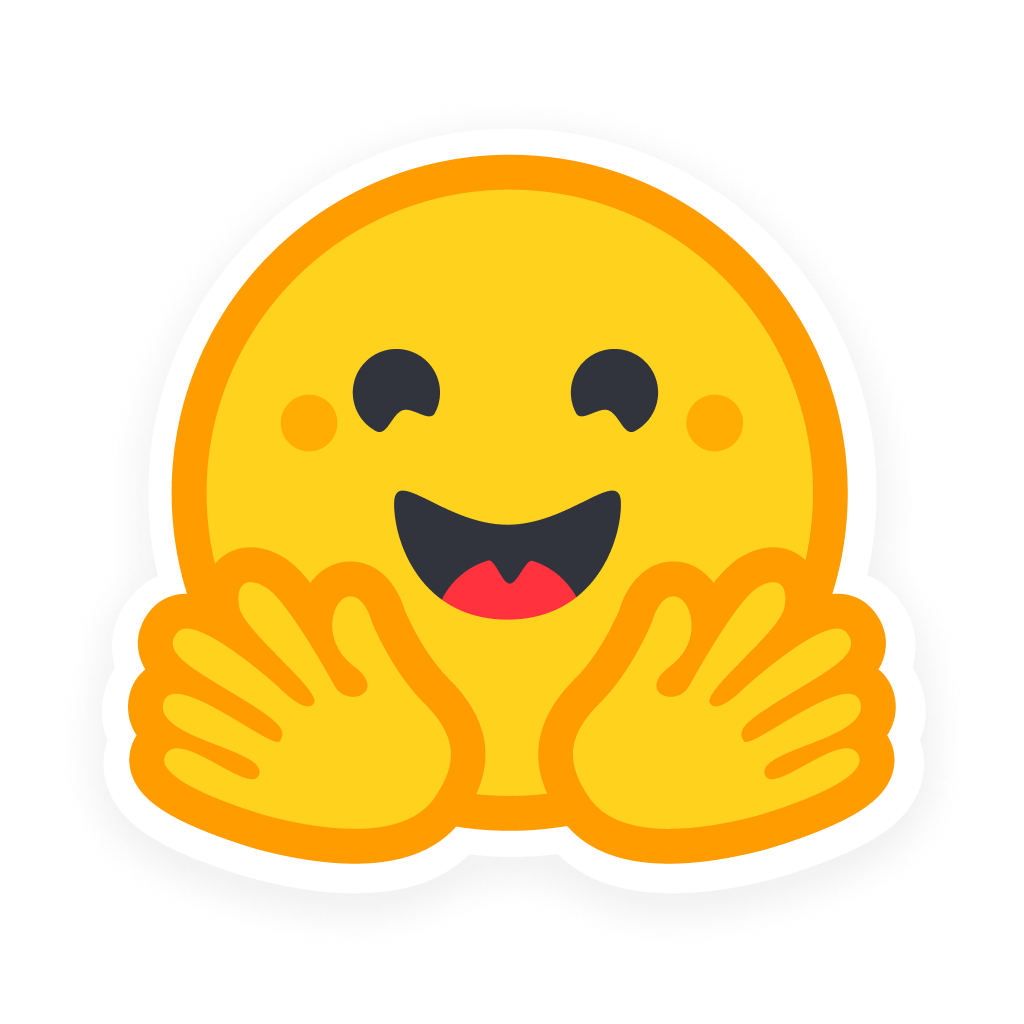}}}
\newcommand{\hflink}[2]{\href{#1}{\textcolor{linkslate}{\texttt{#2}}}}  %
\newcommand{\hfline}[2]{\hflogo\,\hspace{0.15em}\textbf{#1:}\hspace{0.35em}#2}

\title{Building Multilingual Bridges: 
}
\subtitle{Data Mixing as the Pillar of Generalization for In-Language Reasoning}

\author{name={Mehrnaz Mofakhami}, affiliation={1}}
\author{name={Ananya Sahu}, affiliation={1}}
\author{name={Alejandro R. Salamanca}, affiliation={1}}
\author{name={Daniel D'souza}, affiliation={1}}
\author{name={Alexandre Bérard}, affiliation={2}}
\author{name={Thomas Euyang}, affiliation={1}}
\author{name={Marzieh Fadaee\psa}, affiliation={1}}
\author{name={Julia Kreutzer\psa}, affiliation={1}}

\affiliations{
    \item[1] Cohere Labs
    \item[2] Cohere
}

\corresponding[*]{\{\url{mehrnaz.mofakhami},
\url{marzieh},
\url{juliakreutzer}\}\url{{@cohere.com}}}

\abstract{

Reasoning language models have made substantial advances on a variety of complex tasks, yet their capabilities remain overwhelmingly English-centric: models primarily reason in English regardless of the language they are prompted in.
This is inaccessible for non-English-speaking users, risks losing the intent of the original question, and forgoes knowledge more readily expressed in the target language. 
In this work, we advance \textbf{L2 reasoning}, the ability of a model to reason consistently in the language of the user's prompt, thus building an \textit{in-language bridge} between the prompt and the answer.
We approach this problem from a data-centric angle, investigating how to optimize data composition and scheduling in SFT for reasoning generalization. 
Building \tinyayamulti{} at 3.35B scale, 
we achieve an L2 reasoning rate above 93\% across 60 languages on 6 benchmarks spanning math, commonsense reasoning, instruction following, open-ended generation, and cultural reasoning while keeping performance strong. 
We show the path to generalizing L2 reasoning to held-out languages goes through
broader language coverage, readily available multilingual non-reasoning data, and a sufficient English reasoning backbone.
These findings indicate that reasoning is a language-agnostic behavior that can be transferred across typologically diverse languages through careful data mixing and without requiring reasoning supervision in every target language.
We release our model weights and multilingual reasoning data to support further research on accessible, in-language reasoning. 

\vspace{0.6em}
{\setlength{\parindent}{0pt}%
\hfline{Models}{\hflink{https://huggingface.co/CohereLabs/tiny-aya-l2-thinker}{tiny-aya-l2-thinker}}, {\hflink{https://huggingface.co/CohereLabs/tiny-aya-en-thinker}{tiny-aya-en-thinker}}, {\hflink{https://huggingface.co/CohereLabs/tiny-aya-base-32K}{tiny-aya-base-32K}}\\[0.35em]
\hfline{Dataset}{\hflink{https://huggingface.co/datasets/CohereLabs/tiny-aya-l2-thinker-multilingual-reasoning}{tiny-aya-l2-thinker-multilingual-reasoning}}\par}

}
\begin{document}
\section{Introduction}
Reasoning has become a central paradigm for improving the capabilities of language models. 
By generating intermediate sequences of tokens as a bridge to generating an answer, reasoning models can solve problems that require multi-step deduction, calculation, exploration, and verification.
However, the development of these capabilities has been overwhelmingly English-centric~\citep{ghosh-etal-2025-survey}, even in models that are robustly responding to non-English prompts in the respective languages~\citep{bakouch2025smollm3,qwen3.5,team2025gemma,deepseekai2026deepseekv4highlyefficientmilliontoken}. %
The large majority of open-source reasoning datasets for fine-tuning are English-only~\citep{guha2025openthoughts, NemotronPostTrainingDatasetV1} and only very few datasets contain native-language long-form reasoning traces~\citep{m-thinker, lightbluedataset}, with a dominance of math problems. For many large open-weights reasoning models, the linguistic composition of the reasoning training data is often not disclosed~\citep{deepseek-r1, qwen3technicalreport}, making it difficult to determine whether multilingual reasoning supervision is represented at all and if so, to what extent. 
Multilingual models are nonetheless expected to transfer reasoning capabilities acquired primarily through English supervision to other languages. Although substantial cross-lingual transfer occurs, it does not necessarily result in native-language reasoning~\citep{wang-etal-2025-language-mixing,skorobogat2026roundtriptranslationrevealsfrontier}, i.e., reasoning traces that match the language of the prompt, which we refer to as \emph{L2 reasoning}.\footnote{We prefer this term over the more generic ``multilingual reasoning'', which captures both reasoning \textit{for} multilingual prompts and reasoning \textit{in} the user's language.} This creates an important gap between multilingual understanding and multilingual reasoning: a model accepts a prompt in one language while carrying out its reasoning to answer in another language, most commonly English~\citep{wang2025polymath}.

\textbf{The Implications of the Multilingual Reasoning Gap.}
Switching from a non-English prompt to English reasoning introduces the risk of losing intent, nuance, or framing that is specific to the source language context in the process of internal translation, a problem coined as ``lost in translation'' by ~\citet{saji2026reasoninglinguafrancadoubleedged}, 
leading to lower accuracy. 
It has been shown that crosslingual understanding is a major bottleneck for multilingual reasoning even if models have strong translation abilities~\citep{kang-etal-2026-multilingual,ko-etal-2025-understand}, as models switch back and forth between the prompt's content in the target language and reasoning logic in English~\citep{wang-etal-2025-language-mixing, yong2025crosslingualreasoning}. 
Furthermore, there are types of problems where the required knowledge is more readily accessible in the target language due to language-specific associations built in pre-training. For example, ~\cite{sahu2026culturefunnelcantalign} show that cultural knowledge about relevant regions occurs more frequently in the language that is spoken in that region. As a result, questions that draw on culturally embedded knowledge, locally specific terminology, domain expertise or notions of harm, may benefit from native-language reasoning \citep{tam2025languagemattersmultilingualinput}.
While English reasoning has now become a choice of convenience,\footnote{It is an open question whether English would be the objectively optimal reasoning language, or generally if any one individual language would be optimal if only task performance is concerned~\citep{deepseek-r1,kambhampati2026position,huang2026beyond}.} it critically limits the ability to inspect, evaluate, and interact with the reasoning models' intermediate reasoning to those users who speak English---contributing to the deepening of the AI language gap \citep{joshi2020state,peppin2025multilingualdivideimpactglobal,ranathunga2022some}. %
The inspection and understanding of reasoning traces by users is critical in high-stakes domains like healthcare and medical reasoning~\citep{onyame2026cure, ferrazzi2026multilingualmedicalreasoning}, 
and establishing L2 reasoning would also offer an opportunity for more broadly advancing generalization of current reasoning methods, as it allows for meaningful native-language data inspection and reasoning trace analysis ~\citep{marjanovic2026deepseekr,lee2026reasonopsoperatorsegmentationllm} or studies of faithfulness in reasoning~\citep{chen2025reasoningmodelsdontsay}.

\begin{figure}
    \centering
    \includegraphics[width=0.8\linewidth]{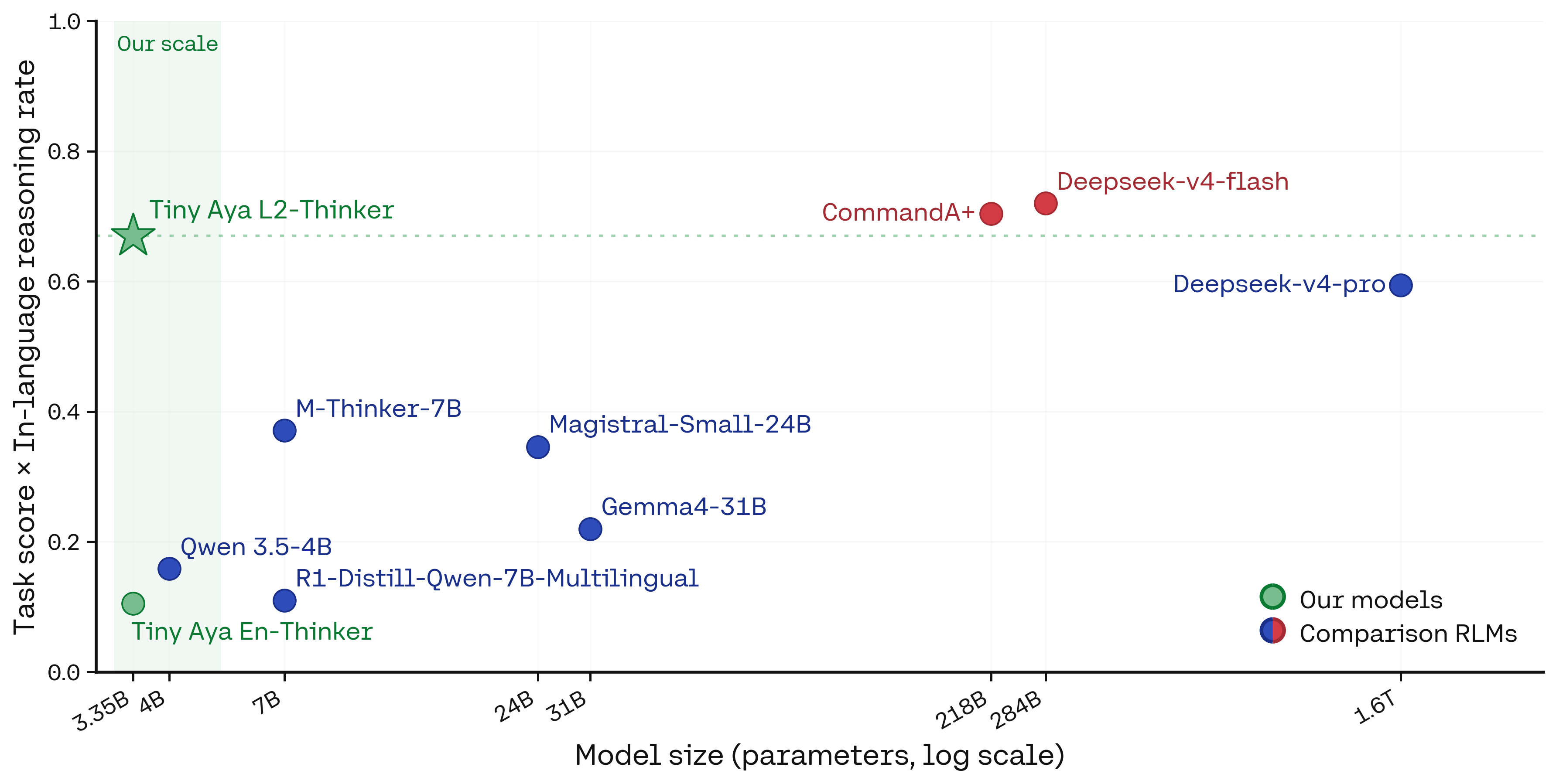}
    \caption{Performance and L2 Reasoning Rate in a combined metric averaged over four benchmarks: \mist{}, MGSM, \mif{}, and GlobalPIQA.  For each benchmark, scores are averaged over all languages (not all might be supported by each model). GPT-4.1 (\texttt{gpt-4-1-04-2025}) is used as the judge across four rubrics of \mist{}. All models are evaluated by prepending an L2 reasoning phrase to the user prompt to incentivize L2 reasoning. \mthinker{}, \textsc{R1-Distill-Qwen-7B-Multilingual}, \magistralsmall{}, and \commandaplus{} are trained for L2 reasoning. \textbf{Even the best L2 reasoning models at large scale fail to achieve perfect in-language reasoning, which shows how challenging the end-goal is. \tinyayamulti{} (ours), at 3.35B scale, achieves a decent score while being substantially smaller.}
    }
    \label{fig:scale_mistoeg}
\end{figure}

We argue that English-only reasoning should not be accepted as the status quo, and challenge existing assumptions that L2 reasoning can only be achieved by trading off performance (the ``\emph{multilinguality tax}'')~\citep{Qi_2025, yong2025crosslingualreasoning}.
This is a rare take, as to date, there are only a few existing open models that support L2 reasoning directly, and even fewer model releases by frontier labs that advertise this feature; one exception being Magistral~\citep{mistralai2025magistral}
For higher-resourced languages, it is often possible to enforce L2 reasoning to a limited extent at test-time (``language forcing'' (LF), e.g. via adding ``Think in language X'' as a prefix to the user prompt)~\citep{yong2025crosslingualreasoning,Qi_2025}, but with less success for lower-resourced languages~\citep{wang2025polymath}. 
For open-weights and smaller models with weaker instruction and cross-lingual generalization skills, this switch is even more difficult~\citep{yang2025parallelscalinglawunveiling}. As~\Cref{fig:scale_mistoeg} illustrates, models at smaller scales (3--7B) such as \qwensmallnew{}~\citep{qwen3.5}, \mthinker{}~\citep{m-thinker}, or \textsc{R1-Distill-Qwen-7B-Multilingual}\footnote{\url{https://huggingface.co/lightblue/DeepSeek-R1-Distill-Qwen-7B-Multilingual}} struggle to remain strong in both task performance and L2 reasoning rate (defined in \Cref{sec:evaluation}), and even the strongest L2 reasoning models such as \commandaplus{} and \textsc{DeepSeek-v4-Flash} don't achieve perfect in-language reasoning, underscoring how challenging the end-goal is.

To make the model accessible for low-resource deployments and use cases, we experiment with the massively multilingual 3.35B \tinyaya{} as a base model~\citep{salamanca2026tinyayabridgingscale}, which we extend to a dual-mode L2 reasoning model covering 45 languages---to our knowledge, the broadest coverage among open-source models optimized for L2 reasoning.
At this scale, crosslingual generalization and instruction following are key obstacles~\citep{murthy2025kcifknowledgeconditionedinstructionfollowing,lim2025languagespecificlatentprocesshinders,zeng2025marcobenchmif},  which we address by optimizing how we mix in auxiliary data from English reasoning and multilingual non-reasoning.
With our resulting model \tinyayamulti{}, we show that when building multilinguality right from the start---not as a posthoc adaptation of an English reasoning model---\textbf{L2 reasoning can be achieved with minimal 
loss in performance, and for many languages at once.}
Owing to the efficient tokenizer of \tinyaya{}~\citep{abagyan-tokenizer-2025} and our efficient reasoning training data~(\Cref{sec:data}), \tinyayamulti{} has far less degenerate repetition than \qwensmallnew{} and uses thinking tokens more efficiently. Moreover, the broad language and task coverage of our training data further yields much smaller cross-language variance in L2 reasoning rate than \mthinker{} and \magistralsmall{} (results in \Cref{sec:results}).

\begin{figure}
    \centering
    \includegraphics[width=\linewidth]{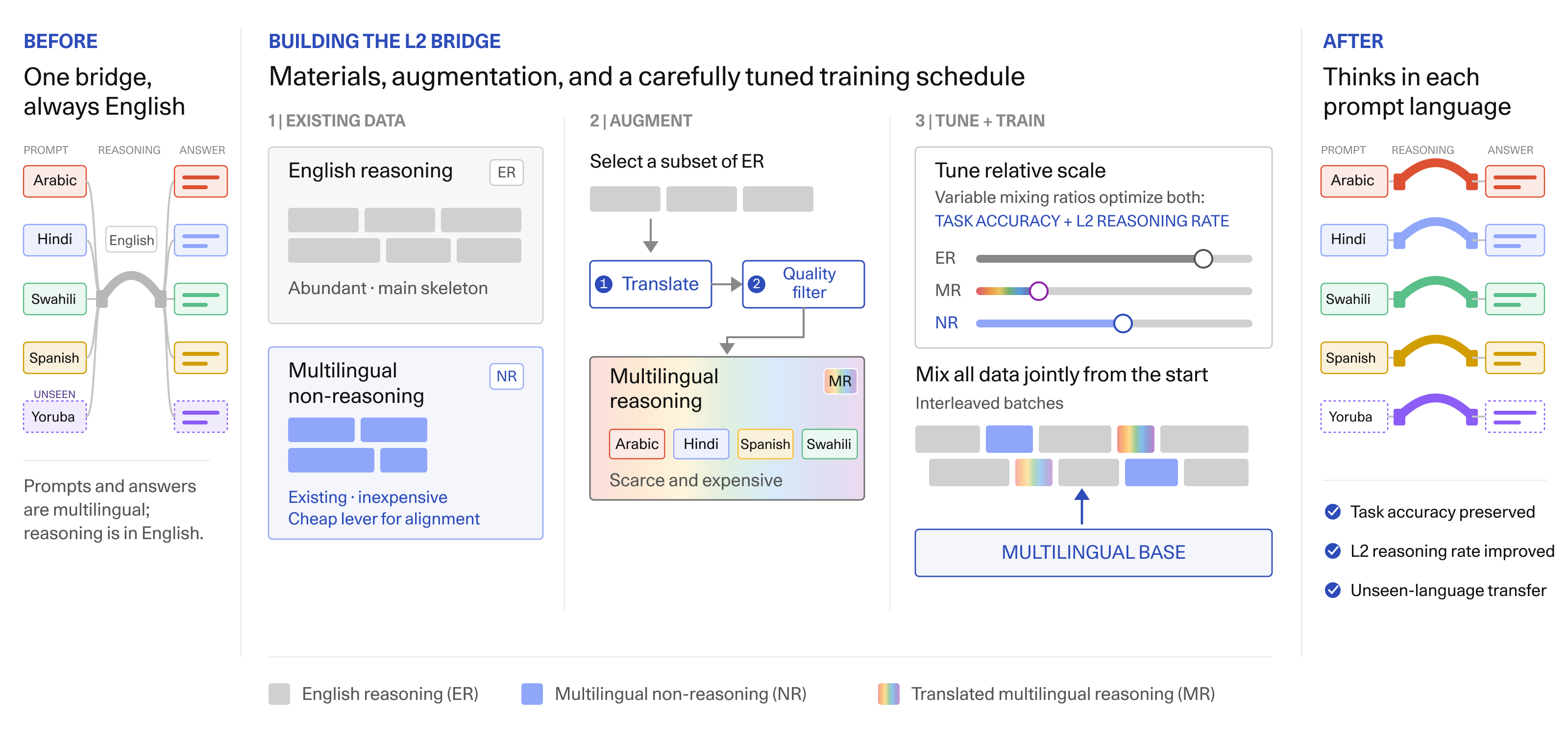}
    \caption{Our contribution: Creating target-language bridges between prompts and responses by optimizing multilingual reasoning capabilities through data augmentation, data mixing and carefully calibrated fine-tuning. 
    }
    \label{fig:bridge}
\end{figure}

The core question we ask is, \emph{how can we optimize jointly for multilingual task accuracy and L2 reasoning, while generalizing to as many languages as possible?} We take a data-centric approach, and show how SFT data should be composed and scheduled so that reasoning \emph{behavior} generalizes across languages. Through careful evaluations across multiple benchmarks covering multiple facets of reasoning, we find that
\emph{(i) more languages help L2 reasoning transfer and do not cause interference} (\Cref{sec:lang_scale}): broadening L2 supervision
maintains accuracy and in-language reasoning on covered languages while increasing L2 reasoning rates on those never supervised, so a single jointly
trained model transfers better than per-region specialists, and the curse of multilinguality \citep{conneau-etal-2020-unsupervised} does not apply to reasoning.
We further identify \emph{(ii) non-reasoning data as
a cheap lever for cross-lingual transfer} (\Cref{sec:NR-sweep}), as multilingual instruction data benefits both L2 reasoning rate and accuracy on unseen languages,
while \emph{English reasoning data remains the necessary backbone especially for difficult math tasks} (\Cref{sec:en_sweep}). 
Lastly, we find that \emph{(iii) how supervision is combined matters} (\Cref{sec:datamix}), with data mixing offering better trade-offs between task accuracy and L2 reasoning rates than sequential adaptation or weight merging. \Cref{fig:bridge} sketches the building blocks of our pipeline at a high level, showing how we go from English-dominated reasoning traces to target-language bridges between prompts and responses.
 
 Together, these results suggest that reasoning is largely language-agnostic for reasoning language models (RLMs) as it is for humans~\citep{Kean2026languageofthought}, and that the language a model uses to bridge between user query and final answer can be controlled with proper training and data mixing. We achieve this with only a small amount of multilingual reasoning data, bringing multilingual reasoning within reach even for languages where such data is scarce.
We release the model weights, and the multilingual reasoning data to support further work on accessible, in-language reasoning. 
We hope that these releases can spark further research into how reasoning can be even less English-centric and more natural and diverse across languages.

\section{Methodology}\label{sec:methodology}

Our goal is to build L2 reasoning without requiring large amounts of training data for every language, and learn to reason beyond easily verifiable domains like math. We focus on the supervised finetuning (SFT) stage, where the model learns from given reference reasoning traces. 
We leverage established post-training techniques, 
and describe (1) how we go from English reasoning teachers to L2 reasoning traces, and (2) which techniques combine multiple data sources.

\subsection{Data augmentation via translation}\label{sec:data-augmentation}
The scarcity of in-language reasoning traces and strong reasoning systems that would produce those makes it impractical to train a separate reasoning system for every target language.
Following established practices in closing data gaps in multilingual instruction following~\citep{muennighoff-etal-2023-crosslingual,ustun-etal-2024-aya}, we leverage automatic translation to go from abundant English reasoning traces to translated target language reasoning traces for training. 
This allows us to obtain multilingual reasoning supervision while preserving the underlying problem and solution structure of the original trajectory.
This approach is also known as ``translate-train''~\citep{pmlr-v119-hu20b} and stands in contrast to approaches that target translation at test time~\citep{huang2023languagescreatedequalllms}. While English reasoning models commonly perform translation as part of their reasoning, training-time translation has the advantage of allowing for more control of the quality of this translation by e.g. optimizing the choice of translation model (particularly relevant for small models that might not be the strongest translators), or filtering translation inputs or outputs.

Compared to translating instruction following data, expected costs for translating reasoning traces are, however, significantly higher, and quality can be expected to be lower. Reasoning traces often span tens of thousands of tokens, and translation models are not typically trained (or tested) on reference translations of reasoning traces, nor specialized on typical reasoning domains like math and code that require strong domain expertise~\citep{wang2025polymath}. Thus, errors might easily accumulate~\citep{kocmi-etal-2025-findings}, and small mistakes can have disproportionate effects on the logical coherence and factual correctness of the reasoning trace. For lower-resourced languages, costs and quality degradations might further increase due to inefficient tokenization~\citep{ahia-etal-2023-languages}.

We consequently treat \emph{translated reasoning as a scarce resource}, focusing on 44 diverse languages but with a small set of samples ($<5K$) for each (details in \Cref{sec:data}).
While scaling up translation further is possible in principle, we prioritize measuring and optimizing for crosslingual generalization only with this small seed set. 
The quality of our target language data depends on the source and the quality of translation, so we carefully filter available English prompts to remove sources that are particularly susceptible to translation artifacts: We require the prompt, reasoning trace, and response to be consistently identified as English, and remove examples containing intra-document code-switching.
We additionally remove trajectories that explicitly discuss translation or name target languages (e.g., \textit{translat, tradu, übersetz}) since translating such content can introduce inconsistencies.
Finally, we remove prompts containing constraints that are difficult to preserve reliably under translation such as exact word counts, length bounds, and capitalization or formatting requirements.

\subsection{Multi-Task post-training} 

\citet{muennighoff-etal-2023-crosslingual} discovered that by multi-task learning~\citep{caruana1997multitask}, i.e., joint training with mixed data, crosslingual generalization can be achieved in fine-tuning, even for languages absent from the finetuning data. 
We build on this observation and ask whether the same principle applies to reasoning language: can multilingual supervision teach a model to condition its reasoning language on the input language, including for languages for which no L2 reasoning traces were provided?

We combine three sources of supervision.
English reasoning (ER) provides abundant reasoning traces and serves as the primary source of task-solving capability.
Multilingual reasoning (MR) data directly supervises reasoning in target languages and multilingual non-reasoning (NR) data provides ordinary instruction-following examples in many languages without an accompanying reasoning trace.
The latter is substantially cheaper to obtain and encourages language alignment to transfer into the reasoning mode. We adopt a \emph{dual-mode} setup: reasoning examples contain explicit reasoning traces, while non-reasoning examples contain an empty reasoning block and instruct the model to answer directly. We combine these sources through \emph{joint training} within each batch---learning to reason in English, reason in other languages, and follow instructions at the same time.

\section{Experimental Setup}\label{sec:experiments}
\subsection{Base Model}
\tinyaya{} is a family of small-scale (3.35B) multilingual 
language models supporting 70+ languages covering five world regions: Asia-Pacific, Europe, Africa, West Asia, and South Asia, making it well-suited for controlled studies of multilingual reasoning across typologically diverse languages. We use the \tinyaya{} Base 
model as starting point for all experiments \citep{salamanca2026tinyayabridgingscale}, with the first modification being the extension of its context length from 8K to 32K, as described below.

\subsection{Long-Context Training}
We extend the model's context length to 32K tokens by resuming cooldown training midway through a linear learning-rate schedule initialized at \(1.25 \times 10^{-4}\). During this stage, we interleave 8K and 32K context sequences at a 3:1 ratio. We balanced the training mixture across data domains~\citep{fu2024data} and context-length buckets to maintain data diversity and ensure a stable transition to long-context modeling. 

\subsection{Training Data}
\label{sec:data}
\textbf{English reasoning (ER) data.}
We consider two English reasoning mixes, a smaller one for more efficient ablations and preliminary experiments, and one larger one for building the final model. 
The \emph{simple} mix pairs AM-Thinking prompts~\citep{ji2025amthinking} with reasoning traces and responses generated by \gptoss{}~\citep{openai2025gptoss120bgptoss20bmodel}, spanning mathematics, science, and general reasoning domains ($\sim$0.66M, $\sim$0.17M, and $\sim$0.89M samples respectively, totalling $\sim$1.7M samples). 
The extended mix augments these with the math and science subsets of Dolci-Think-SFT-32B~\citep{olmo2025olmo3} ($\sim$75K) and Open-Thoughts-114K~\citep{guha2025openthoughts} ($\sim$380K), which contribute substantially longer traces. 
 The controlled experiments in \Cref{sec:lang_scale} and \Cref{sec:NR-sweep} use the \emph{simple} mix: its shorter traces keep training time controlled and make the many-way data-composition and strategy comparisons tractable. We reserve the extended mix for the later scaling experiments in \Cref{sec:en_sweep} and our final model.

\begin{figure}[h]
    \centering
    \includegraphics[width=0.9\linewidth]{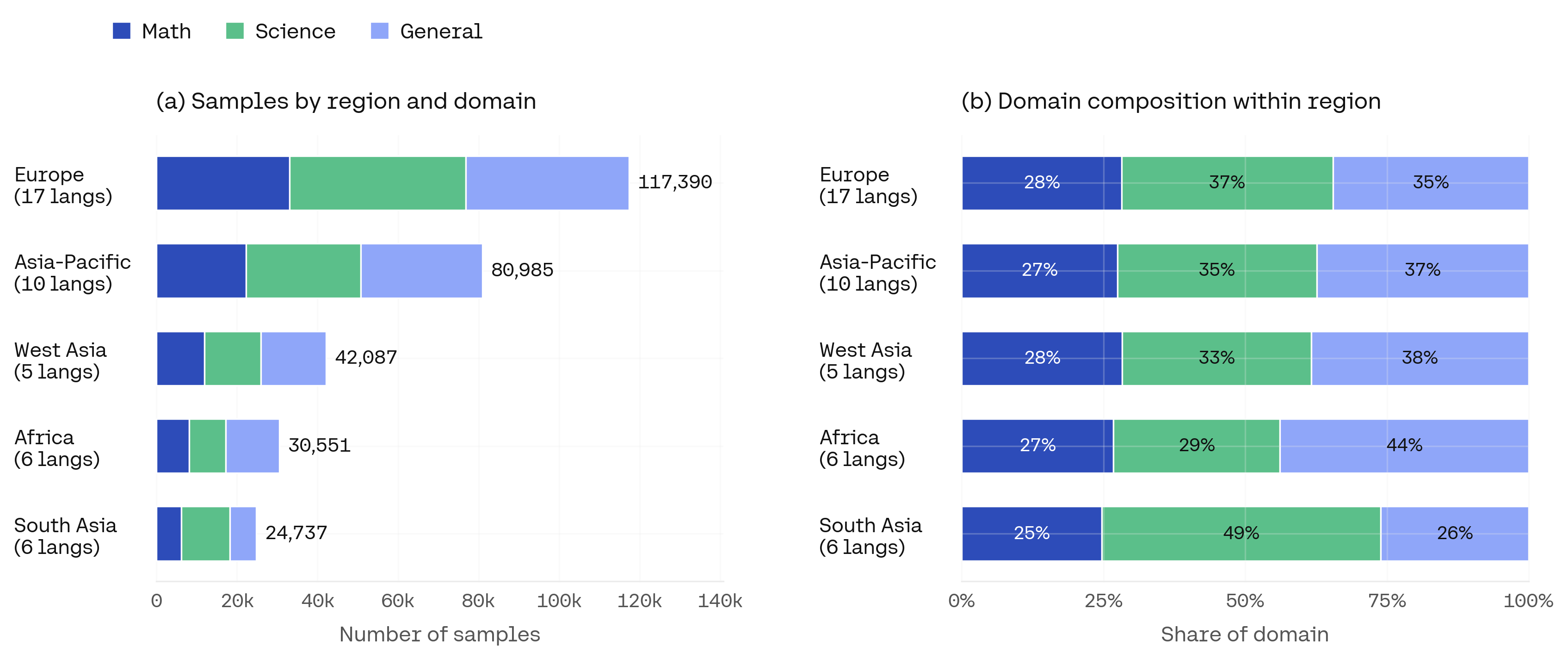}
    \caption{Data composition of our multilingual reasoning data per \tinyaya{} regions (Europe, Asia-Pacific, West Asia, Africa, South Asia) and data categories (Math, Science, and General)}
    \label{fig:MR-data-composition}
\end{figure}

\textbf{Multilingual reasoning (MR) data.} To supervise L2 reasoning, we translate the English reasoning data into diverse target languages selected from the list of languages that \tinyaya{} supports, with the main goal of creating a rich subset that captures language family, script and resource levels. 
We translate the data using \commandtranslate~\citep{kocmi-etal-2025-command}  for its supported languages and \deepseek ~\citep{liu2024deepseek} for the rest, sampling sources independently per language to maximize input diversity, and cap each language at 5K samples\footnote{For the controlled experiments in \Cref{sec:lang_scale} and \Cref{sec:NR-sweep}, we cap the number of samples at 5K per language ($\sim$1.5K per domain); however, in the final \tinyayamulti{} model we use all of our available translated data that includes more samples for some high-resource languages: ar, fr, de, ja, ko.}.
For preliminary experiments and ablations in~\Cref{sec:controlled_exps} we work with a subset of ten languages, two selected by region: Europe (German, French), Asia-Pacific (Japanese, Korean), South Asia (Hindi, Bengali), West Asia (Arabic, Persian), and Africa (Swahili, Zulu).
For the final MR dataset, we translate reasoning traces into an additional 34 languages across the five regions (breakdown in \Cref{fig:MR-data-composition}), teaching our  \tinyayamulti{} model to reason in 45 languages (incl. English).\footnote{Language list: Amharic, Bulgarian, Bengali, Catalan, Czech, English, Greek, Basque, Persian, Finnish, Filipino, Irish, Hausa, Hebrew, Hindi, Hungarian, Indonesian, Igbo, Italian, Javanese, Khmer, Lithuanian, Malay, Maltese, Norwegian, Punjabi, Polish, Russian, Slovak, Swahili, Tamil, Telugu, Thai, Turkish, Ukrainian, Urdu, Vietnamese, Yoruba, Chinese, Zulu, Arabic, German, French, Japanese, Korean} Detailed statistics are given in \Cref{tab:translation-stats}.\footnote{We release the data at \url{https://huggingface.co/datasets/CohereLabs/tiny-aya-l2-thinker-multilingual-reasoning}}

\textbf{Multilingual non-reasoning (NR) data.} 
Fine-tuning exclusively on less multilingual, and less domain-diverse reasoning data risks losing specific capabilities such as target-language generation and instruction following in \tinyaya{}'s languages. 
To counteract this, we incorporate multilingual non-reasoning instruction-following data into our training mix. It consists of a total of $\sim$4.9M
samples spanning all of \tinyaya{}'s 67 languages 
to prevent catastrophic forgetting of languages that are not included in the MR data. Our non-reasoning data combines translated general instruction data (e.g.\ from Dolci Instruct SFT~\citep{olmo2025olmo3}) and curated and filtered public translation data~\citep{kocmi-etal-2025-command,proietti-etal-2025-estimating}, region-specific instruction data~\citep{mora2025art}, the Aya Collection~\citep{ayadata2024}, and prompts from \citet{khairi-etal-2025-life}, together with a small amount of synthesized data targeting common MCQA formatting (disjoint from all our benchmarks).

NR examples are included with empty thinking blocks and marked with special tokens that signal the model to respond directly without reasoning. This allows the model to retain general multilingual capabilities alongside its reasoning skills, without requiring full reasoning supervision. In Section~\ref{sec:NR-sweep}, we study how the proportion of NR data in the mixture affects multilingual understanding and cross-lingual L2-thinking generalization.

\subsection{Evaluation}\label{sec:evaluation} 

\textbf{Benchmarks.}
We evaluate multilingual reasoning across a diverse set of capabilities relevant for multilingual use cases, including both parallel (translated) and non-parallel benchmarks with a wide language coverage:
\begin{itemize}
    \item \textbf{Math at multiple difficulty levels}: To cover the to-date most dominant direction of evaluations~\citep{ghosh-etal-2025-survey}, we include the \textbf{GlobalMGSM}~\citep{salamanca2026tinyayabridgingscale} extension of MGSM~\citep{shi2023language-mgsm}---referred to as MGSM throughout the text---%
    and \textbf{PolyMath}~\citep{wang2025polymath}  for evaluating math reasoning. MGSM includes grade-school arithmetic math problems and PolyMath includes more advanced math problems in four difficulty levels (the first level being equivalent to MGSM); both are translated from English. They require writing answers to math prompts in multiple languages, but the correct answers that are used for evaluation are mostly identical across languages since they are numerical.
    \item \textbf{Multicultural and multilingual reasoning}: In addition to evaluating multilingual proficiency, we assess cultural reasoning abilities utilizing \textbf{GlobalPIQA}~\citep{chang2026globalpiqa} (non-parallel subset) and the multilingual variant of \textbf{\macaron{}}~\citep{elsetohy-etal-2026-macaron}. GlobalPIQA evaluates physical commonsense reasoning, which requires the model to make a binary choice between two seemingly plausible alternatives to a highly language/culture-specific prompt. \macaron{} consists of natively written culturally grounded questions spanning 22 cultural aspects and 7 reasoning types (mathematical, commonsense, causal, temporal, logical, spatial, and multihop) across 20 languages. The questions require strong language proficiency, reasoning abilities and cultural understanding, to select the correct answer from four plausible options. 
    \item \textbf{Localized instruction following}: The ability to follow instructions in any language is key to allowing meaningful user interactions. This aspect is usually overlooked when training reasoning models. \textbf{\mif{}}~\citep{zeng2025marcobenchmif} tests this ability with localized prompts that require fulfilling constraints for desired output formats and contents in the model response.
    \item \textbf{Linguistically diverse open-ended generation}:
    To assess language proficiency in writing, we include the open-ended generation sub-task from the WMT 2025 Multilingual Instruction Shared Task
    (\textbf{\mist{}})~\citep{kocmi-etal-2025-mist}. It contains localized open-ended questions such as brainstorming, creative or professional writing, or information requests, which require aspects like naturalness and coherence that are not captured in any of the other benchmarks. We use the same open-ended generation rubric-based LLM-as-a-judge setup as~\citep{salamanca2026tinyayabridgingscale} that extends the one from~\citep{kocmi-etal-2025-mist} by the dimension of accuracy. We use GPT-4.1 (\texttt{gpt-4-1-04-2025}) as the judge across four rubrics.        
\end{itemize}

\textbf{Language selection.} Where necessary, we automatically translate benchmark instruction prompts into the target languages so that they do not contain any English instructions anymore.
We exclude benchmark languages that the base model \tinyaya{} does not support due to our post-training scope. This still covers a diverse list of 60 languages including high- and low-resource languages. For ablations, we distinguish \emph{seen} languages (those languages from each benchmark present in MR data) from \emph{unseen} ones (absent from MR but present in NR data). More details are in \Cref{app:benchmark-languages}.

\textbf{Metrics.} Our main concerns are \emph{whether} the model answers correctly, and \emph{which language} it reasons in: we want both high task accuracy and high L2 reasoning rate. Furthermore, we want to achieve efficient use of a fixed 32K token budget. We track these complementary aspects of a successful L2 reasoning model with the following metrics:

    \quad\textbf{Task accuracy} measures to what degree the final answer is correct under the metric definition of the respective benchmark. This is the commonly tracked metric in reasoning research.
    
    \quad\textbf{L2 reasoning rate} measures the percentage of the samples where the model's reasoning trace is predominantly written in the prompt language, according to a FastText language-identification \citep{joulin2016fasttext} classifier, with GlotLID~\citep{kargaran-etal-2023-glotlid} as a fall-back for FastText's unsupported languages.

    \quad\textbf{Repetition rate} 
    measures the fraction of repeated $4$-grams in a reasoning trace~\citep{li2023repetition,yao-etal-2025-understanding}, capturing potentially degenerate repetitions, and we use it as a proxy for \emph{doomlooping}. When models ``doomloop'', i.e. never recover from looping through the same sequence of tokens, they often exhaust the token budget preventing them from arriving at a final answer, which in turn hurts task accuracy. For a trace with distinct $4$-grams
$g \in \mathcal{G}$ occurring with counts $c_g$, we define the
\emph{$4$-gram repetition score} as:
\begin{equation}
 \frac{\sum_{g:\, c_g > 1} (c_g - 1)}{\sum_{g} c_g}.
\end{equation}

We also track the reasoning length in number of tokens, where reasoning traces are tokenized based on each model's own tokenizer. 
    
\subsection{Baselines \& External Comparisons}

\textbf{Same scale comparison with inference-time language forcing.} Our primary comparison is against inference-time language forcing methods on \qwensmallnew{}~\citep{qwen3.5}, a massively multilingual reasoning model at our scale supporting 201 languages trained only on English reasoning---reflecting the status quo of multilingual reasoning at the 3--4B scale. 

For \qwensmallnew{}, we use two approaches for forcing in-language reasoning at inference time, either via a user message prefix or as prefix for the reasoning trace. For the former, we prepend the phrase ``\texttt{Think in the same language as the prompt.}'' to the user request (user prefix LF); this is a low-cost approach that any user can apply. As an alternative, we prefill the start of the reasoning trace with the equivalent of ``\texttt{Here's a thinking process to solve the problem:}''---a phrase that appears frequently at the beginning of \qwensmallnew{}'s reasoning traces---translated into the prompt's language (thinking prefix LF). Downstream users cannot pre-fill the reasoning trace when access is gated behind a platform, so this second approach must in practice be handled by the model provider and requires an additional language-identification step of the user prompt to pick the prefix in the right language. This makes it less attractive in practice, as it requires language-specific processing and prevents genuine exploration by the model within these first reasoning trace tokens.

\textbf{English-only reasoning.} For a direct data-controlled comparison, we also train a \tinyaya{}-derived English reasoning model. It receives the same training data as its L2 counterpart, but the reasoning traces are entirely in English.  

\textbf{Larger L2-reasoning optimized models.}
In addition, we compare against two existing larger models that were specifically optimized towards L2 reasoning (see \Cref{sec:related}): 

\mthinker{}~\citep{m-thinker}\footnote{We chose the 7B model (\url{https://huggingface.co/XueZhang-bjtu/M-Thinker-7B-Iter2}) over the 1.5B one based on preliminary probes where 1.5B performed very poorly.} is optimized on five languages (ja/ko/fr/pt/th) via SFT and RL with L2 reasoning-tailored rewards on math, based on DeepSeek-R1-Distill-Qwen-7B~\citep{deepseek-r1}\footnote{\url{https://huggingface.co/deepseek-ai/DeepSeek-R1-Distill-Qwen-7B}}, and \magistralsmall{} supports 24 languages while being trained on in-language reasoning traces for a few of them (fr/es/it/de/zh/ru) via language consistency reward \citep{mistralai2025magistral}.
Decoding settings are described in \Cref{app:decoding}.

\section{Results}\label{sec:results}

\definecolor{rowhl}{HTML}{ECEAE3}
\definecolor{l2hl}{HTML}{E4E8F2}
\newcommand{\lcolgap}{%
  &&&\cellcolor{l2hl}&&\cellcolor{l2hl}&&\cellcolor{l2hl}&&\cellcolor{l2hl}&&\cellcolor{l2hl}&&\cellcolor{l2hl}\\
}
\begin{table*}[t]
\centering
\scriptsize{%
\setlength{\tabcolsep}{2pt}
\resizebox{\textwidth}{!}{%
\begin{tabular}{ll ll ll ll ll ll ll}
\toprule
\multirow{2}{*}{\textbf{Model}} & \multirow{2}{*}{\textbf{Langs.}} & \multicolumn{2}{c}{\textbf{MGSM}} & \multicolumn{2}{c}{\textbf{PolyMath}} & \multicolumn{2}{c}{\textbf{\mist{}}} & \multicolumn{2}{c}{\textbf{\mif{}}} & \multicolumn{2}{c}{\textbf{GlobalPIQA}} & \multicolumn{2}{c}{\textbf{\macaron{}}} \\
\cmidrule(lr){3-4}\cmidrule(lr){5-6}\cmidrule(lr){7-8}\cmidrule(lr){9-10}\cmidrule(lr){11-12}\cmidrule(lr){13-14}
 & & Acc & \cellcolor{l2hl}L2\% & Acc & \cellcolor{l2hl}L2\% & Acc & \cellcolor{l2hl}L2\% & Acc & \cellcolor{l2hl}L2\% & Acc & \cellcolor{l2hl}L2\% & Acc & \cellcolor{l2hl}L2\% \\
\midrule
\multicolumn{14}{l}{\textbf{English Reasoning}} \\
\midrule
  & English & $92.8$ & \cellcolor{l2hl}$100.0$ & $17.3$ & \cellcolor{l2hl}$99.3$ & $95.3$ & \cellcolor{l2hl}$100.0$ & $71.2$ & \cellcolor{l2hl}$97.6$ & $75.0$ & \cellcolor{l2hl}$100.0$ & -- & \cellcolor{l2hl}-- \\
  \multirow{-2}{*}{\makecell[l]{\tinyayaen{}\\ (Ours)}} & Other   & $\mathbf{70.8}_{\pm14.9}$ & \cellcolor{l2hl}$0.8_{\pm3.1}$ & $18.6_{\pm1.4}$ & \cellcolor{l2hl}$0.1_{\pm0.1}$ & $85.3_{\pm5.4}$ & \cellcolor{l2hl}$10.5_{\pm8.5}$ & $\mathbf{52.8}_{\pm5.0}$ & \cellcolor{l2hl}$7.9_{\pm6.4}$ & $72.3_{\pm9.5}$ & \cellcolor{l2hl}$28.4_{\pm28.1}$ & $42.6_{\pm10.1}$ & \cellcolor{l2hl}$15.2_{\pm16.0}$ \\
  \lcolgap
    & English & $96.4$ & \cellcolor{l2hl}$100.0$ & $24.6$ & \cellcolor{l2hl}$99.3$ & $90.8$ & \cellcolor{l2hl}$98.0$ & $37.0$ & \cellcolor{l2hl}$95.7$ & $88.0$ & \cellcolor{l2hl}$100.0$ & -- & \cellcolor{l2hl}-- \\
  \multirow{-2}{*}{\qwensmallnew{}} & Other   & $52.2_{\pm24.7}$ & \cellcolor{l2hl}$31.0_{\pm22.7}$ & $39.8_{\pm5.6}$ & \cellcolor{l2hl}$0.2_{\pm0.5}$ & $79.9_{\pm10.4}$ & \cellcolor{l2hl}$15.2_{\pm7.8}$ & $23.8_{\pm8.0}$ & \cellcolor{l2hl}$31.4_{\pm7.7}$ & $\mathbf{78.6}_{\pm10.5}$ & \cellcolor{l2hl}$7.9_{\pm11.1}$ & $37.0_{\pm14.0}$ & \cellcolor{l2hl}$12.8_{\pm12.6}$ \\
\midrule
\multicolumn{14}{l}{\textbf{L2 Reasoning}} \\
\midrule
  & English & $93.6$ & \cellcolor{l2hl}$100.0$ & $16.9$ & \cellcolor{l2hl}$98.9$ & $95.0$ & \cellcolor{l2hl}$100.0$ & $72.7$ & \cellcolor{l2hl}$98.5$ & $70.0$ & \cellcolor{l2hl}$100.0$ & -- & \cellcolor{l2hl}-- \\
  \multirow{-2}{*}{\makecell[l]{\tinyayamulti{} \\(Ours)}} & Other   & $68.0_{\pm14.1}$ & \cellcolor{l2hl}$\mathbf{96.5}_{\pm9.6}$ & $11.1_{\pm2.5}$ & \cellcolor{l2hl}$\mathbf{94.9}_{\pm7.7}$ & $\mathbf{85.8}_{\pm4.8}$ & \cellcolor{l2hl}$95.2_{\pm15.2}$ & $50.7_{\pm5.3}$ & \cellcolor{l2hl}$\mathbf{96.8}_{\pm5.5}$ & $70.7_{\pm9.7}$ & \cellcolor{l2hl}$\mathbf{98.3}_{\pm8.4}$ & $39.8_{\pm9.2}$ & \cellcolor{l2hl}$\mathbf{93.8}_{\pm15.2}$ \\
  \lcolgap
  & English & $93.8$ & \cellcolor{l2hl}$99.6$ & $21.9$ & \cellcolor{l2hl}$99.2$ & $91.3$ & \cellcolor{l2hl}$99.0$ & $34.6$ & \cellcolor{l2hl}$97.8$ & $90.0$ & \cellcolor{l2hl}$100.0$ & -- & \cellcolor{l2hl}-- \\
  \multirow{-2}{*}{\makecell[l]{\qwensmallnew{}\\ (user prefix LF)}} & Other   & $51.2_{\pm25.2}$ & \cellcolor{l2hl}$33.9_{\pm24.8}$ & $\mathbf{40.3}_{\pm5.2}$ & \cellcolor{l2hl}$0.2_{\pm0.5}$ & $81.5_{\pm8.3}$ & \cellcolor{l2hl}$29.4_{\pm11.3}$ & $21.2_{\pm7.5}$ & \cellcolor{l2hl}$33.3_{\pm8.0}$ & $78.1_{\pm10.9}$ & \cellcolor{l2hl}$12.2_{\pm13.2}$ & $30.2_{\pm19.2}$ & \cellcolor{l2hl}$13.5_{\pm12.0}$ \\
  \lcolgap
  & English & $92.0$ & \cellcolor{l2hl}$100.0$ & $23.7$ & \cellcolor{l2hl}$100.0$ & $57.3$ & \cellcolor{l2hl}$100.0$ & $34.6$ & \cellcolor{l2hl}$95.6$ & $90.0$ & \cellcolor{l2hl}$100.0$ & -- & \cellcolor{l2hl}-- \\
  \multirow{-2}{*}{\makecell[l]{\qwensmallnew{}\\ (thinking prefix LF)}} & Other   & $50.6_{\pm34.4}$ & \cellcolor{l2hl}$94.3_{\pm12.2}$ & $27.6_{\pm9.3}$ & \cellcolor{l2hl}$87.6_{\pm15.6}$ & $67.7_{\pm15.4}$ & \cellcolor{l2hl}$93.3_{\pm12.4}$ & $33.8_{\pm14.3}$ & \cellcolor{l2hl}$82.3_{\pm9.0}$ & $77.0_{\pm13.1}$ & \cellcolor{l2hl}$94.7_{\pm15.7}$ & $37.7_{\pm19.7}$ & \cellcolor{l2hl}$91.1_{\pm20.7}$ \\
  \lcolgap
  \multispan{14}\unskip{\color[HTML]{C4C1B8}\xleaders\hbox{\kern 1.6pt \vrule height 0.35pt width 1.8pt \kern 1.6pt}\hfill}\kern0pt \\
  \lcolgap
  & English & $84.8$ & \cellcolor{l2hl}$97.6$ & $42.6$ & \cellcolor{l2hl}$98.3$ & $90.7$ & \cellcolor{l2hl}$99.0$ & $57.1$ & \cellcolor{l2hl}$97.2$ & $71.0$ & \cellcolor{l2hl}$100.0$ & -- & \cellcolor{l2hl}-- \\
  \multirow{-2}{*}{\mthinker{}} & Other   & $38.0_{\pm31.6}$ & \cellcolor{l2hl}$87.8_{\pm24.0}$ & $32.6_{\pm7.4}$ & \cellcolor{l2hl}$88.0_{\pm29.5}$ & $30.0_{\pm16.4}$ & \cellcolor{l2hl}$\mathbf{96.5}_{\pm12.2}$ & $28.2_{\pm7.3}$ & \cellcolor{l2hl}$94.1_{\pm16.6}$ & $59.2_{\pm9.0}$ & \cellcolor{l2hl}$92.2_{\pm21.6}$ & $32.7_{\pm8.8}$ & \cellcolor{l2hl}$87.7_{\pm29.9}$ \\
  \lcolgap
  & English & $98.0$ & \cellcolor{l2hl}$99.6$ & $32.1$ & \cellcolor{l2hl}$100.0$ & $97.8$ & \cellcolor{l2hl}$100.0$ & $78.9$ & \cellcolor{l2hl}$96.1$ & $88.0$ & \cellcolor{l2hl}$99.0$ & -- & \cellcolor{l2hl}-- \\
  \multirow{-2}{*}{\magistralsmall{}} & Other   & $64.2_{\pm31.2}$ & \cellcolor{l2hl}$33.6_{\pm39.1}$ & $15.9_{\pm8.1}$ & \cellcolor{l2hl}$39.4_{\pm32.0}$ & $80.8_{\pm12.5}$ & \cellcolor{l2hl}$60.4_{\pm32.3}$ & $50.9_{\pm16.5}$ & \cellcolor{l2hl}$48.9_{\pm30.4}$ & $77.1_{\pm10.5}$ & \cellcolor{l2hl}$49.0_{\pm43.3}$ & $\mathbf{49.6}_{\pm13.8}$ & \cellcolor{l2hl}$14.7_{\pm26.4}$ \\
\bottomrule
\end{tabular}%
}
}
\caption{Task accuracy (Acc) and L2 reasoning rate (L2\%) on English and averaged over all covered languages besides English ($\pm$ std across those languages) for baselines and our \tinyayamulti{} model trained on 45 reasoning languages. Models after the dashed separator have larger scale than our model but we still include them for comparison: 7B/24B vs. 3.35B. \macaron{} is country-based and does not have English in the original implementation. LF stands for language forcing. The task accuracy metric for \mist{} is a $1$--$7$ judge score rescaled to [0,100]; the PolyMath column blends the three difficulty levels per language as $(2\,\text{medium} + 4\,\text{high} + 8\,\text{top}) / 14$. The highest value among non-english averages is bold for Acc and L2\%.} 
\label{tab:main-results}
\end{table*}

\begin{figure*}[t]
    \centering
    \includegraphics[width=\linewidth]{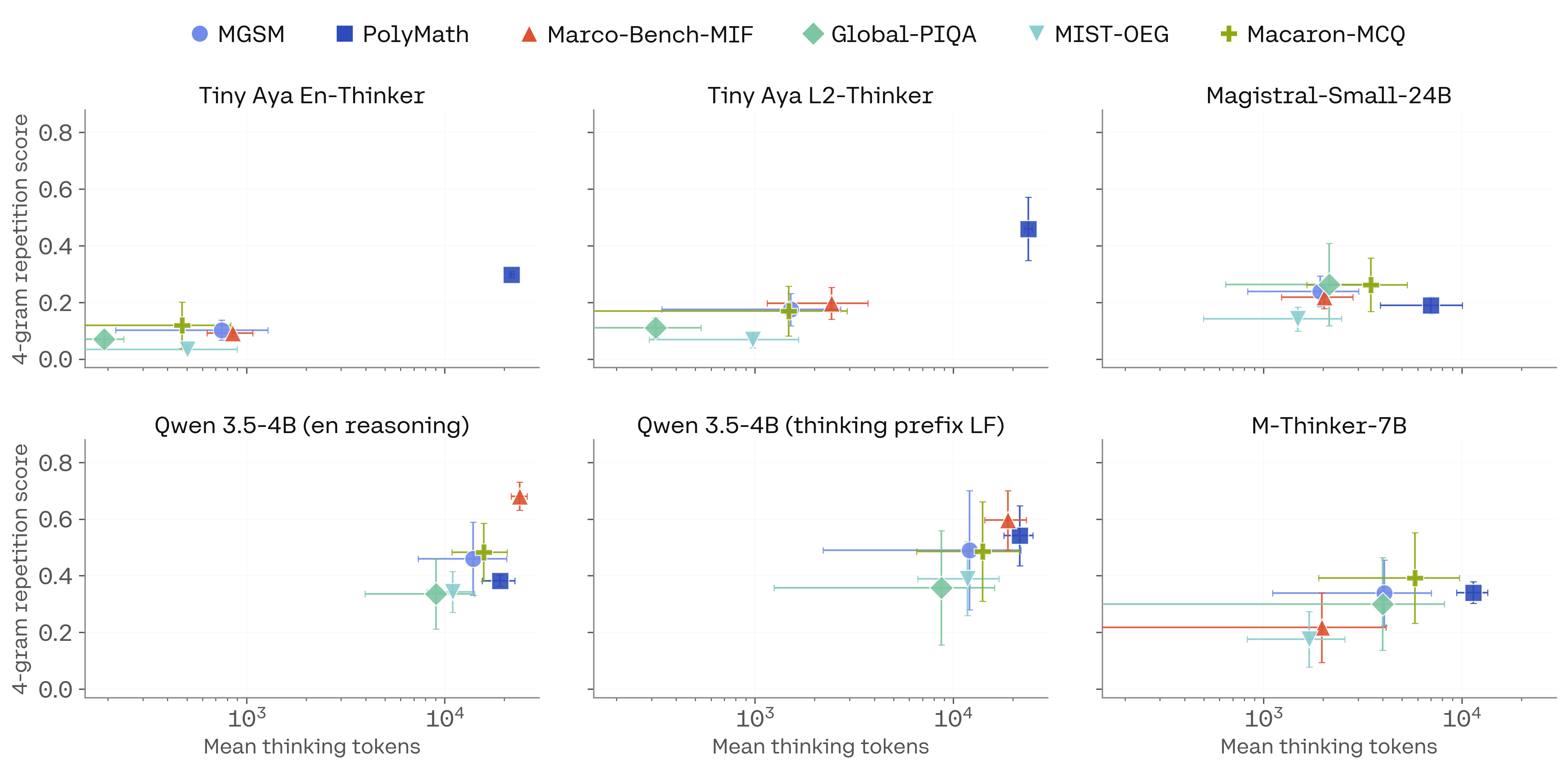}
    \caption{4-gram repetition score as our doomlooping proxy versus mean number of thinking tokens (log scale) across six benchmarks; error bars are ±1 s.d. across all the benchmark's languages. Longer traces track more repetition: Qwen3.5-4B (LF) sits at the high-token, high-repetition extreme, while \tinyayamulti{} reasons in far fewer tokens except for the competition-level PolyMath task, and shows markedly less repetition. Scores are length normalized, averaged across 1024-token windows.  
    }
    \label{fig:doomlooping}
\end{figure*}
 
We group systems by whether they are meant to reason in the user's language. One group reasons in English by design: \qwensmallnew{} and \tinyayaen{}, an English-only reasoner trained on the same backbone as our model (same base and data, but with English reasoning traces). The other group targets L2 reasoning, either at inference time through language forcing (\qwensmallnew{}~user/thinking LF) or through training (\tinyayamulti{}, \mthinker{}, \magistralsmall{}). 

For each system we report metrics on two language groups in \Cref{tab:main-results}: \emph{English}, for English prompts as a control, and \emph{Other}, averaged over the remaining benchmark languages, with the standard deviation across those languages (\Cref{app:benchmark-languages} for more details). Per-language numbers are detailed in tables~\ref{tab:app-mgsm} to \ref{tab:app-macaron_mcq} in the Appendix.

\subsection{Forcing the reasoning language is not a substitute for training}\label{sec:lang-forcing}
Thinking prefix LF lifts \qwensmallnew{}'s L2 reasoning rate far above user prefix LF, yet even this stronger variant falls short of \tinyayamulti{}. Trained for in-language reasoning on multilingual data, \tinyayamulti{} reaches a higher L2 rate than \qwensmallnew{} with thinking prefix LF on every benchmark and is more stable across languages. It is also more accurate on several tasks, most clearly on the open-ended generation benchmarks \mist{} and \mif{}, where linguistic understanding matters most.

Forcing is fragile by construction as it relies on templates the model may have never encountered in training, so its success depends on the model's instruction-following and on how sensitive it is to edits of the prompt or reasoning prefix~\citep{m-thinker}. Prefilling the forcing phrase as a generation prefix can also steer the model away from its more likely generation paths and cost accuracy: on \qwensmallnew{}, thinking prefix LF lowers other-language accuracy on four of six benchmarks, by roughly $12$ points on both PolyMath and \mist{}.
When L2 reasoning is instead introduced during training, it is naturally integrated into the model's learned behavior. \tinyayamulti{} reasons in the target language on more than 93\% of traces on every benchmark, across both seen and unseen languages, and does so with consistently low variance. It exceeds \mthinker{} (87.7–96.5\%) on five of six benchmarks at half the size, and exceeds \magistralsmall{} on all tasks by a considerable margin. In Table \ref{tab:main-results} we report both language-forcing variants for \qwensmallnew{}, but in the remainder of the paper we report only the stronger method, thinking prefix LF.

\subsection{In-language reasoning comes at minimal cost to accuracy}
Switching from English to in-language reasoning barely moves \tinyayamulti{}'s other-language accuracy relative to its English-reasoning counterpart: it drops by at most 2--3\% on five of the six tasks and gives ground only on PolyMath (11.1\% vs. 18.6\%). English performance is preserved too: the model continues to reason in English on English prompts and matches the English reasoner on most benchmarks, the one real drop being 5 points on GlobalPIQA. In-language reasoning therefore costs at most a few points outside competition-level math, in exchange for an L2 rate that climbs from near zero to above 93\%.

PolyMath is the one benchmark where \qwensmallnew{} leads on accuracy ($40.3$\% for \qwensmallnew{} (user prefix LF) vs.\ our $11.1$\%), but the gap decomposes: $21.7$ points separate \qwensmallnew{} from our own English-reasoning model ($18.6$), and only $7.5$ separate that model from \tinyayamulti{}. This reflects the absence of an RL stage for our models, since competition-level math benefits from RL refinement \citep{shen2026understandingreasoningpretrainingposttraining}, which also smoothes translation artifacts \citep{wang2025polymath}. 
\mthinker{} shows the cost of chasing that accuracy too narrowly: math-targeted RL on five languages raises its PolyMath score to $32.6$, but the model collapses on open-ended generation ($30$ on MIST and $28.2$ on \mif{}, against $85.8$ and $50.7$ for our model). \tinyayamulti{} instead reasons concisely in the prompt language across every evaluated language, making it a well-rounded multilingual reasoner rather than a specialist.

\subsection{Long traces signal doomlooping, not deeper reasoning}
Longer reasoning is not undesirable in itself: harder problems such as advanced math captured in PolyMath warrant more deliberation, but traces that stay long regardless of difficulty signal degenerate generations rather than deeper reasoning. \Cref{fig:doomlooping} bears this out: across models, the 4-gram repetition score---our proxy metric used to detect doomlooping~\citep{li2023repetition,yao-etal-2025-understanding}---rises with the mean number of used thinking tokens. This shows that when traces get longer, there are also more repetitions within these traces. Some repetition is unavoidable as sequences grow against a fixed vocabulary, but longer traces should ideally add new contents rather than redundantly repeating existing contents. \qwensmallnew{} with thinking prefix LF sits at the high-token, high-repetition extreme, with particular outliers on \mif{} and \mist{} (above 0.6 and 0.4 4-gram repetition scores respectively). Recalling from Table~\ref{tab:main-results}, these are the tasks for which \qwensmallnew{} has the weakest performance, showing that the extra tokens are not buying performance. Manual inspection confirms that many of these traces are in fact doomlooping,  (see \Cref{sec:reasoning_errors} for an example). \cite{scaleorreason} also find that longer traces correlate with failure and  tend to be incorrect.
\tinyayamulti{}, trained at the same 32K context length as \qwensmallnew{}, controls length far better, staying short and low-repetition on easier benchmarks while spending more tokens on PolyMath.

\subsection{Efficiency and Coverage on Low-resource Languages}

\begin{figure}[t]
    \centering
    
    \includegraphics[width=1.0\linewidth]{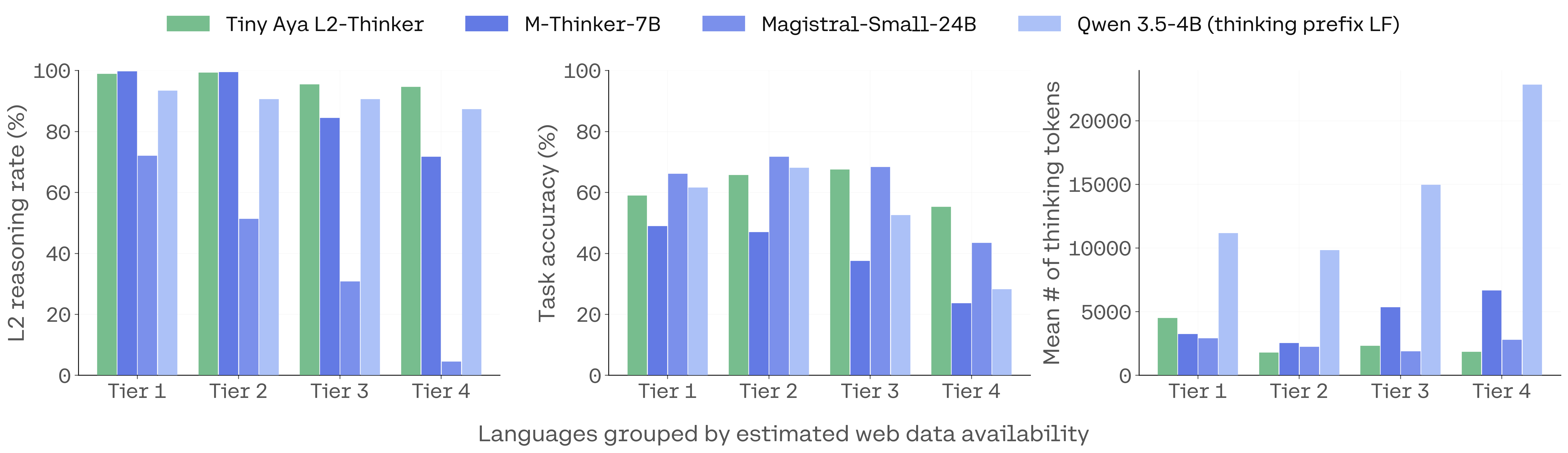}
    \caption{\textbf{Efficiency and coverage of L2 reasoning LLMs across tiers of language resourcedness.}
Tiers range from the highest-resourced (Tier~1) to the lowest-resourced (Tier~4) languages. \tinyayamulti{} maintains broad L2 reasoning coverage across all four tiers, whereas \mthinker{} and \magistralsmall{} achieve lower L2 reasoning rates when evaluated on lower-resourced languages. \tinyayamulti{} is also more token-efficient, using under $2{,}000$ thinking tokens on average, while \qwensmallnew{} uses far more, with token counts increasing toward the lower-resourced tiers. Values are averaged over the languages in each tier and over the benchmarks that include them. \\[2pt]
\begin{tabular}{@{}l@{\quad}*{14}{l@{~}}r@{}}
\textbf{Tier~1:} & cs, & de, & en, & es, & fr, & id, & it, & ja, & nl, & pl, & pt, & ru, & tr, & vi, & zh\\
\textbf{Tier~2:} & ar, & bg, & ca, & el, & fa, & fi, & he, & hu, & ko, & no, & ro, & sk, & sv, & th, & uk\\
\textbf{Tier~3:} & bn, & et, & eu, & gl, & hi, & hr, & lt, & mr, & ms, & ne, & sl, & sr, & ta, & te, & ur\\
\textbf{Tier~4:} & am, & cy, & gu, & ha, & ig, & jv, & km, & my, & sn, & sw, & tl, & wo, & xh, & yo, & zu\\
\end{tabular}
}
\label{fig:efficiency-coverage}
\end{figure}

\Cref{fig:efficiency-coverage} draws a three-way trade-off among coverage (L2 reasoning rate), useful performance (task accuracy), and efficiency (mean number of reasoning trace tokens), across four tiers of resourcedness (1: highest, 4: lowest). We rank languages by Common Crawl page count and bin them into four groups as a proxy for available web data for pretraining: tier~1 (20--870M) for the highest-resource languages, tier~2 (5--20M), tier~3 (400K--5M), and tier~4 (5K--400K) for the lowest-resource languages. 
The comparison is restricted to models that actually produce in-language reasoning to some degree: \tinyayamulti{}, \mthinker{}, and \magistralsmall{} via training, and \qwensmallnew{} (thinking prefix LF) via inference-time language forcing. For each tier, we first average the metric across benchmarks covering each language before averaging uniformly over languages within each tier.

None of these metrics is sufficient without the other two, and they should be considered all together. \tinyayamulti{} is the only model that remains strong on all three axes as resourcedness falls. Its L2 rate stays near ceiling on tiers 1–2 (99\%) and declines only modestly on lower-resourced languages (94.3\% on tier~3, 94.5\% on tier~4), while \magistralsmall{}’s in-language reasoning collapses (72\% → 5\%). \mthinker{}'s performance drops as we move toward lower-resourced tiers on all metrics: L2 reasoning rate falls from 99.6\% to 71.6\%, task accuracy from 48.8\% to 23.5\%, and mean thinking length roughly doubles from 3.2k to 6.6k tokens.  While thinking prefix language forcing keeps \qwensmallnew{}'s L2 rate high even on lower-resourced tiers (93.3\% on tier~1, 87.2\% on tier~4; still below \tinyayamulti{}), its task accuracy drops quickly. While on tier~1 \qwensmallnew{} has higher accuracy than \tinyayamulti{} (61.5\% vs.\ 58.8\%), on tier~4 it stands behind \tinyayamulti{} by more than 25\%, and the reasoning traces get unnecessarily long for lower-resourced languages. 

Overall, our model achieves both broader language coverage (45 vs.\ 6 languages) and stronger performance than \mthinker{}, a model twice its size. The variance in L2 reasoning rates for \tinyayamulti{} is roughly half that of \mthinker{}, indicating more robust L2 thinking. Compared to \qwensmallnew{}, \tinyayamulti{} attains comparable performance while achieving higher L2 reasoning rate, and more efficient traces.

\section{Analysis and Building Blocks}\label{sec:controlled_exps}
We organize the analysis around two questions.
The first question is how to achieve cross-lingual L2 reasoning with minimal target-language reasoning supervision.
We analyze the impact of three main pillars with a set of controlled experiments---broader language coverage (\Cref{sec:lang_scale}), a small fraction of multilingual non-reasoning data (\Cref{sec:NR-sweep}), and a sufficient English reasoning backbone (\Cref{sec:en_sweep})---each of which transfers L2 reasoning to languages carrying little or no in-language reasoning data. 
The second question is how to navigate the trade-off between task accuracy and L2 reasoning rate: comparing joint mixing against sequential adaptation and model merging (\Cref{sec:training_strategies_tradeoff,sec:fallback-lang}); we find that mixing gives the best trade-off and the most predictable failures.

\subsection{More languages improve transfer of reasoning behavior}
\label{sec:lang_scale}

\begin{figure}[t]
\centering
\includegraphics[width=\linewidth]{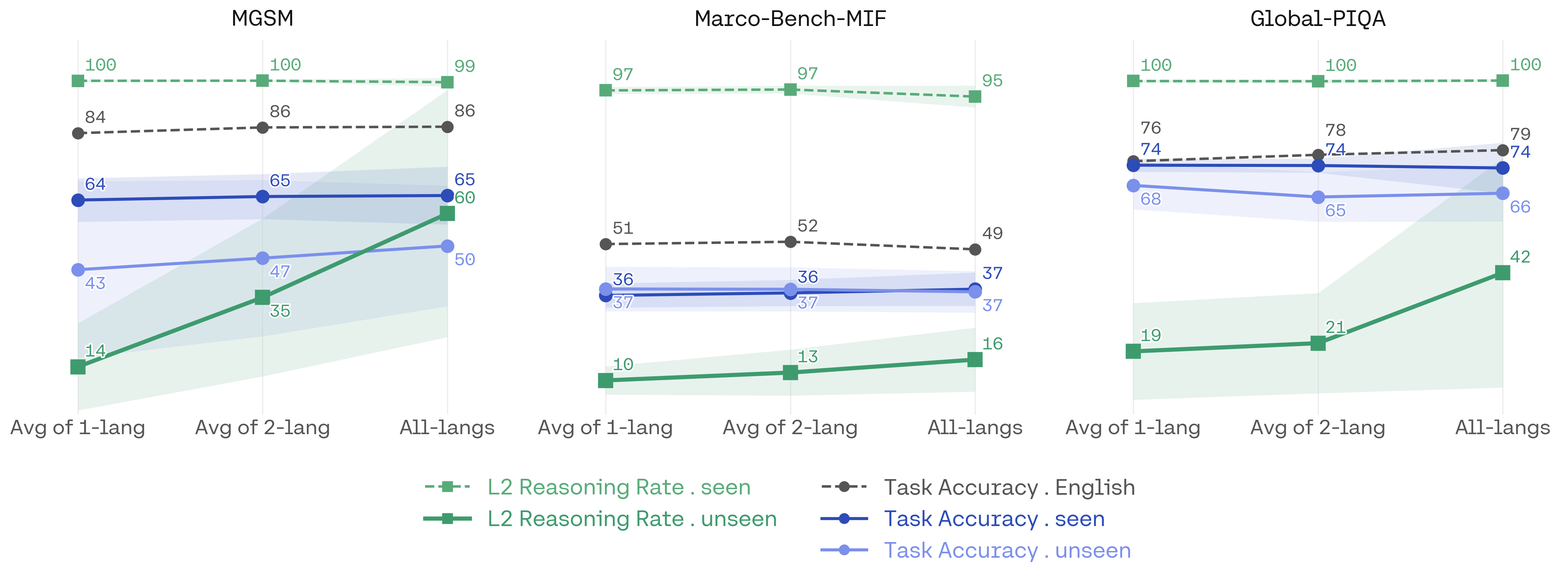}
\caption{Effect of language coverage across MGSM, \mif{}, and GlobalPIQA. Each panel shows, as coverage grows from the 1-Lang specialists to the 2-Lang regional models to the single All-Langs model, English/seen/unseen task accuracy and seen/unseen L2 reasoning rate. Task accuracy and seen LangID are flat within error bars (no interference), while unseen-language L2 reasoning rate rises monotonically, indicating that broader language coverage improves generalization to unseen languages.
}
\label{fig:region-avg}
\end{figure}

Adding languages to a model is often expected to create a trade-off: broader language coverage may improve transfer, but competing languages may interfere with capabilities already acquired or lead to code-switching.
For L2 reasoning, this raises a specific question: does exposing a model to more reasoning languages improve its ability to reason in new languages, or does it dilute the language-specific behavior it has already learned? 

We compare three levels of language coverage: language-specific specialists trained with one target language, regional models trained with two languages, and a single model jointly trained on all ten target languages:
\textbf{1-Lang} specialists (English + one target language; ten models, two per region across Europe, Asia-Pacific, South Asia, West Asia, and Africa), \textbf{2-Lang} regional models (English + the two target languages of one region; five models), and a single \textbf{All-Langs} model trained on English + all ten languages jointly. 
We evaluate both task accuracy and L2 reasoning rate separating languages that received L2 reasoning supervision from held-out languages.

To ensure that the comparison reflects language coverage rather than differences in the evaluation distribution, we first average within each region and then across the five regions, so every region contributes equally regardless of how many of its languages a given benchmark happens to cover. 
For each benchmark, the seen and unseen language sets are fixed in advance, since the languages available for evaluation differ across the benchmarks.
The 10 training languages of this experiment are German, French, Japanese, Korean, Hindi, Bengali, Arabic, Persian, Swahili, and Zulu;
the benchmark-specific seen and unseen splits are given in Table~\ref{tab:seen-unseen}.

We find that expanding language coverage improves transfer without producing the expected interference.
Across all three benchmarks in \Cref{fig:region-avg}, English accuracy, task accuracy on seen languages, task accuracy on unseen languages, and L2 reasoning rate on seen languages remain essentially stable as coverage increases.
The behavior that changes is \emph{L2 reasoning rate} on unseen languages, which rises monotonically with coverage on all three benchmarks: $14\to35\to60$ on MGSM, $10\to13\to16$ on \mif{}, and $19\to21\to42$ on GlobalPIQA. 
The transfer is thus selective: added languages improve in-language reasoning specifically on languages that never received it as supervision, while leaving already-acquired capabilities intact. 
We observe that L2 reasoning behaves less like a fixed capacity that must be divided among languages and more like a transferable behavioral pattern whose generalization improves with broader linguistic coverage. This finding corroborates ~\citet{yang2025parallelscalinglawunveiling}'s observation that reinforcement learning of mathematical reasoning on multiple languages jointly helps transfer reasoning capabilities to other languages.

\subsection{A small proportion of non-reasoning data yields positive transfer across metrics}
\label{sec:NR-sweep}

The previous experiments show that multilingual reasoning can transfer across languages while still relying on multilingual reasoning traces.
We then ask whether the language alignment needed for L2 reasoning can be learned from cheaper supervision that contains no in-language reasoning traces.

\begin{figure}[t]
\centering
\includegraphics[width=\linewidth]{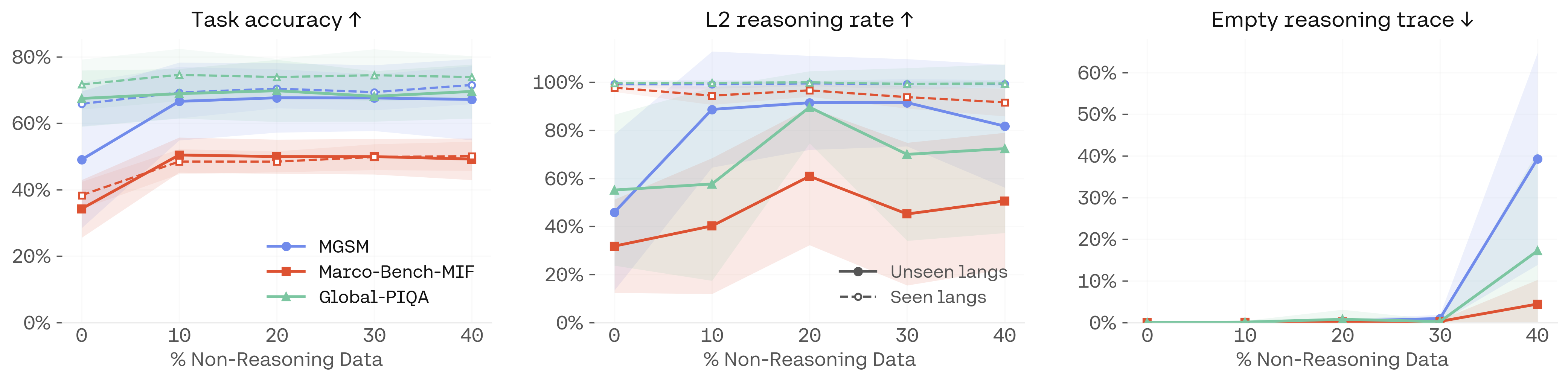}
\caption{Effect of the multilingual non-reasoning data fraction (L2 reasoning fixed at 10\%; the remainder is English reasoning). \emph{Left:} task accuracy. \emph{Middle:} rate of reasoning in the target language (L2 reasoning rate). \emph{Right:} empty thinking-trace rate. Solid lines are averaged over unseen languages, dashed over seen languages; one line per benchmark; bands show $\pm1$ std across languages. A small non-reasoning fraction sharply improves both in-language reasoning and accuracy on unseen languages, with a sweet spot around 20--30\% before the model starts skipping thinking at around 40\%.}
\label{fig:nr-sweep}
\end{figure}

\textbf{Batch-wise mixing.} We sweep the fraction of multilingual non-reasoning data while holding L2 reasoning fixed at 10\% and trading the remainder against English reasoning, spanning the NR mix at 0\%, 10\%, 20\%, 30\%, 40\%. The mixture proportions are enforced at the \emph{batch level}, and we compare checkpoints at matched training steps so that differences reflect data composition rather than training length. \Cref{fig:nr-sweep} reports, as a function of the non-reasoning fraction, task accuracy, the rate at which the model reasons in the target language (L2 reasoning rate), and the fraction of empty thinking traces, each averaged over the \emph{unseen} languages (solid) and \emph{seen} languages (dashed), with one line per benchmark.

\textbf{A little non-reasoning data goes a long way.} The largest change comes from the very first increment. On MGSM unseen languages, moving from 0\% to 10\% raises the rate of in-language reasoning from $46\%$ to $89\%$ and, strikingly, lifts task accuracy from $49\%$ to $67\%$ at the same time. Without any non-reasoning data the model has high English-reasoning capacity but reasons in English on many unseen prompts; adding a small non-reasoning SFT slice couples input language to output language and this anchoring \emph{transfers into reasoning mode}, improving both metrics together, indicating a cross-mode transfer. Non-reasoning data, however, helps only in moderation, with a sweet spot at around 20--30\%. Beyond that, the model increasingly skips reasoning altogether, as shown by the empty reasoning trace fraction in \Cref{fig:nr-sweep}. 

\subsection{English reasoning data as the backbone for mathematical reasoning}
\label{sec:en_sweep}

In this experiment, we measure the effect of English reasoning data on L2 reasoning, tracing how the dynamic evolves as we increase the English backbone. We begin with a multilingual-only model trained on Multilingual Reasoning (MR) and Non-Reasoning (NR) data, then add our extended English mix at fractions of 10\%, 25\%, 50\%, 75\%, and 100\%.

\begin{figure}[t]
    \centering
    \includegraphics[width=0.9\linewidth]{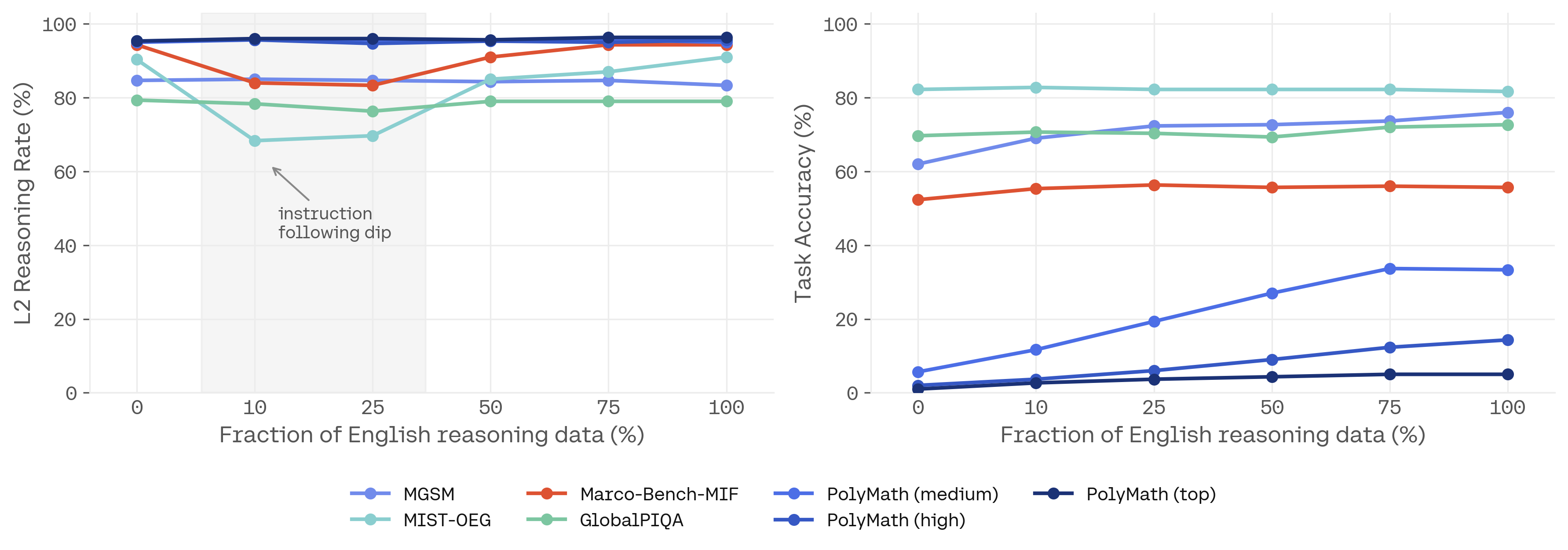}
    \caption{English-reasoning sweep, averaged over all languages (English and our 44 translated languages). Starting from a multilingual-only model (0\%), we add our extended English reasoning mix at increasing fractions (10\%, 25\%, 50\%, 75\%, 100\%). \textbf{Left:} L2 reasoning rate. Math reasoning benchmarks (MGSM, PolyMath) stay stable, while non-math benchmarks (\mist{}, \mif{}, GlobalPIQA) show a dip at 10–25\% (shaded region) before recovering as the English backbone grows. \textbf{Right:} Task accuracy (\%, with MIST rescaled from its 1–7 scale). Math benchmarks, especially PolyMath (medium), gain substantially from a heavier English backbone via language transfer, whereas open-ended generation tasks are largely insensitive to the English fraction.}
    \label{fig:en-sweep}
\end{figure}

We observe that the two task families respond differently in~\Cref{fig:en-sweep}. For math benchmarks (MGSM, PolyMath), the L2 reasoning rate stays nearly flat while task accuracy climbs with a heavier English backbone. We attribute this to knowledge transfer: math reasoning benefits most from English data, so more of it lifts performance without disrupting the language of the thinking trace. 

Open-ended generation tasks (\mist{}, \mif{}), by contrast, gain almost nothing from additional English data; 10\% of the mix behaves the same as 100\%. These open-ended benchmarks reveal a subtler dynamic in the L2 reasoning rate. The multilingual-only model starts with a high L2 reasoning rate; adding a small fraction of English data drives it down, and it takes a heavier English backbone to recover to the original level. We attribute this recovery to instruction following: as English data increases, the model becomes better at following instructions in general, so when prompted to reason in L2 it digests and complies with that instruction more reliably.

These three pillars each push the accuracy-L2 reasoning rate outward, but all assume a single model trained by joint mixing; once the recipe is fixed, how English and L2 supervision are combined \emph{in time} forces a genuine choice between them. We justify that choice by comparing mixing against two strategies that fragment the process---sequentially adapting an English-only reasoner, and merging separately trained specialists---and examining how each behaves when it fails to reason in the target language. For a controlled comparison, we conduct this experiment in the ten-language, five-region setup of \Cref{sec:lang_scale}, reusing the specialist models trained there; its fixed seen/unseen structure keeps the analysis clean and directly comparable to the coverage results above.

\subsection{Data mixing beats sequential adaptation and merging}\label{sec:datamix}

Strong English reasoning models are already available at various scales, so the cost of joint training from scratch and the need to tune data balances might appear unattractive for language expansion. We test whether lower-effort (1) merging of specialist single-language reasoners or (2) continued training on L2 data from an English reasoner are promising alternatives. We refer to \Cref{app:ablations} for the details of this ablation but summarize the key findings here.

When we revisit the language coverage experiment (\Cref{sec:lang_scale}) with merging the 5 specialized models and sequential L2 training starting from English, we find that merging might be best for task accuracy, but collapses to reasoning entirely in English. Sequential training has the opposite effect: it excels on L2 reasoning but loses task accuracy in the process, and introduces unexpected language confusion in the reasoning trace. Data mixing represents a middle ground between both, offering the best trade-off (\Cref{fig:training_strategies_pareto}), plus a consistent fallback to English reasoning when L2 reasoning fails (\Cref{fig:unseen-thinking-fallback}).

We also find that merging the final \tinyayamulti{} with \tinyayaen{} or sequential training of \tinyayaen{} on only L2 reasoning do not add any further benefits beyond the optimized data mixing for \tinyayamulti{}. 
Overall, this analysis indicates that initial integration of multilingual reasoning—rather than subsequent retrofitting—yields significant advantages in closing the multilingual reasoning gap.

\subsection{Comparing contributions of each data pillar}

\begin{figure}[h]
    \centering
    \includegraphics[width=0.9\linewidth]{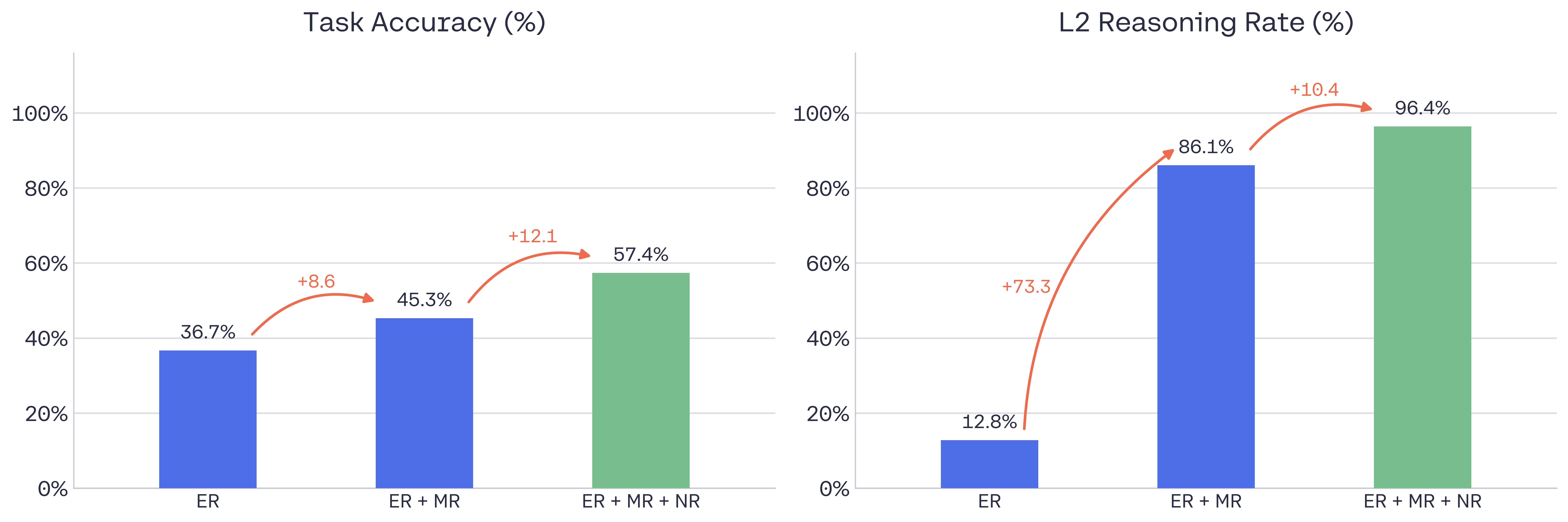}
    \caption{\textbf{Task accuracy and L2 reasoning rate both evolve as we add relevant data pillars: ER (English Reasoning), MR (Multilingual Reasoning), and NR (Non-Reasoning)}. ER was trained only with English Reasoning data. Adding Multilingual Reasoning data raises L2 Reasoning rate sharply as expected; and adding Non-Reasoning data further improves both metrics, shaping our final model, \textbf{\tinyayamulti{}}. Means are averaged across MGSM, PolyMath, \mist{}, \mif{}, and GlobalPIQA.}
    \label{fig:data_pillar_ablation}
\end{figure}

As a final remark, \Cref{fig:data_pillar_ablation} draws a high-level view of how each of our three data pillars (ER, MR, and NR) contributes to advancing both performance and L2 reasoning rate. Trained on English reasoning data alone (ER), the model reaches only 36.7\% task accuracy and reasons in the target language just 12.8\% of the time: it solves a portion of problems but almost always thinks in English. Adding multilingual reasoning data (MR) raises the L2 reasoning rate sharply (to 86.1\%), showing that multilingual reasoning traces play a major role in eliciting in-language reasoning. Adding non-reasoning data (NR) on top improves both axes: it raises the L2 reasoning rate further by helping the model generalize to languages for which it has seen no reasoning traces, and it lifts task accuracy by strengthening the model's multilingual understanding. 
 
\section{Related Work}\label{sec:related}

The language gap in multilingual reasoning has been approached from three different angles: leaving the reasoning trace in English (in whole or in part) while improving how non-English inputs are handled, steering the thinking language at inference time, and training models to reason natively in the target language.

\subsection{Improving English reasoning models on multilingual inputs}\label{sec:related_enreas}

One family of methods keeps reasoning in English by design and instead improves the model's ability to process non-English inputs. \citet{yoon2024langbridgemultilingualreasoningmultilingual} connect a frozen multilingual encoder to a frozen English-centric reasoner through a small set of trainable parameters, trained without any multilingual supervision; \citet{huang2024mindmerger} merge the reasoner's internal capability with an external multilingual model via a learned mapping layer, and \citet{ruan2025layalignenhancingmultilingualreasoning}  extend this by fusing all encoder layers rather than only the top one. \citet{uemura2026merlinmultistagecurriculumalignment} add a multi-stage curriculum and report gains on low-resource language benchmarks.

A parallel line recomposes experts in parameter space rather than through a learned bridge: \citet{bandarkar2025layer} introduce layer swapping for zero-shot cross-lingual transfer by interleaving the layers of a math expert and a language expert, and \citet{li-etal-2026-enhancing} make the interpolation between a reasoning model and a multilingual model steerable at inference. These methods consistently improve accuracy on mathematical reasoning tasks, but they are explicit that the intermediate reasoning remains predominantly English, and we are not aware of any that report the language of the reasoning trace among their metrics. The capability being transferred is therefore reasoning accuracy, not reasoning \emph{in} the user's language, which is the behavior we study.

A related thread accepts mixed-language reasoning as the target rather than an error mode. \citet{son2026pushingmultilingualreasoningmodels}  propose Language-Mixed Chain-of-Thought (CoT), in which an English scaffold is retained while entities and key terms stay in the target language, and show that this outperforms both English-only and target-only traces for Korean; \citet{lin2026thinkmultilingualhardercodeswitch} curate a code-switched reasoning corpus and train models to code-switch deliberately in a data-efficient way. These works share our premise that the language of the trace is a learned, controllable behavior, but they optimize for a hybrid trace, whereas we target traces that a monolingual user of the prompt language can read end to end.

\subsection{Controlling reasoning language at inference}

The distinction between reasoning in the input language and reasoning in English
dates back to user-language-CoT versus English-CoT prompting \citep{shi2023language-mgsm}, and to cross-lingual thought prompting, which explicitly instructs the model to restate the problem in English before reasoning \citep{huang2023languagescreatedequalllms}. More recent work applies the same idea in the opposite direction, prompting reasoning models to think in a given language (``language forcing'')~\citep{yong2025crosslingualreasoning}. \citet{tam2025languagemattersmultilingualinput} show that large reasoning models default to a dominant language and that constraining them to the input language degrades accuracy on reasoning-intensive tasks, while helping on culturally grounded ones. \citet{Qi_2025} formalize this as a trade-off: prompt-based control raises the rate at which models reason in the requested language and makes traces auditable, but costs accuracy, and the effect is strongest for lower-resource languages. \citet{luo2025mmathmultilingualbenchmarkmathematical} document the complementary failure, off-target generation, and introduce two prompting baselines that seed the reasoning block in the target language, either with a discourse marker or by restating the question. \citet{wang2025polymath} report the resulting picture at scale across 18 languages: general-purpose models keep input--output language consistency above 95\%, whereas reasoning models are substantially lower, with the reasoning process the weakest point.

Inference-time control therefore succeeds only where a model can already sustain extended reasoning in the target language, and prompting redistributes probability mass toward that behavior rather than instilling it. The asymmetry reported by \citet{wang2025polymath} makes this concrete: models comply with the directive in their final answer far more reliably than in the trace that precedes it, so the limiting factor is not instruction-following but the capacity to hold a chain of reasoning in-language, and this fails in the lower-resource regime we care about. We adopt language forcing applied to an English-only reasoner as our primary baseline.

\subsection{Teaching L2 reasoning}

An earlier generation of work translates English reasoning data---mostly mathematical--- and updates the model, or parts of it, on the result. \citet{chen-etal-2024-breaking} build a translated GSM8K training set and show that multilingual training also benefits English; \citet{zhu2024questiontranslationtrainingbetter} instead train question translation as an auxiliary task before English reasoning fine-tuning; \citet{lai2024mcotmultilingualinstructiontuning} translate and reformat CoT data across eleven languages and introduce \emph{reasoning consistency}---whether a model reaches the same answer for the same question posed in different languages. We note that this is agreement of \emph{outcomes} under translation, and is orthogonal to the question of which language the trace itself is written in; the two can be high and low respectively, as language forcing makes clear. A recurring theme in this line is catastrophic forgetting, which motivates either updating only the layers responsible for multilinguality \citep{fan-etal-2025-slam}, moving to preference optimization over translated pairs \citep{she-etal-2024-mapo}, or careful distillation \citep{payoungkhamdee-etal-2024-empirical}. More recently, \citet{gurgurov2026reasonxlshiftingllmreasoning} scale translate-train to a cross-domain parallel corpus in five languages and report that adapting a model to reason in one language largely preserves performance in others, evidence that the surface reasoning language can be changed without disturbing the underlying language-agnostic representations.

We share the translate-train premise of this line
(\S\ref{sec:data-augmentation}), but differ in what is translated, how much of it,
and what the translation is for. These works translate short-form mathematical
CoT into a handful of languages to \emph{constitute} the training set; we
translate long-form reasoning traces, whose length and domain specificity make
translation both costlier and more error-prone (\S\ref{sec:data-augmentation}),
across 44 languages and three domains, and we treat the result as a deliberately
scarce seed---under 5K samples per language---whose purpose is to measure how far
L2 reasoning generalizes beyond the languages it covers. This changes what has to
be controlled: we filter the English source for properties that survive
translation poorly (\S\ref{app:translation}), pair the translated traces
with multilingual non-reasoning data in a dual-mode mixture rather than
fine-tuning on translations alone, and, because catastrophic forgetting is
answered here by mixture composition rather than by restricting the update or the
objective, we test that choice directly against the alternatives this literature
adopted (\S\ref{sec:datamix}). Where these works report reasoning
consistency, we report the language of the trace itself.

More recent work optimizes the reasoning language directly during post-training, with the main focus on RL with verifiable rewards. \citet{m-thinker} reward strict language consistency over both thought and answer via reinforcement learning (RL), together with an alignment reward scoring non-English traces against the model's own English trace. Similarly, \citet{hwang2025learngloballyspeaklocally} pair multilingual alignment with a consistency reward, and argue for benchmark scoring that should include the reasoning trace. \citet{ki2026makesgoodmultilingualreasoning} question that mechanism from the evaluation side, decomposing multilingual traces into measurable features and finding that the association between a given feature and accuracy varies substantially across languages and can even reverse, so that similarity to an English reference trace is a competitive but not universally correct target. We correspondingly supervise reasoning directly in the target language instead of scoring it against an English counterpart. \citet{lee2025makingqwen3thinkkorean} take the narrower route of language-targeted RL for a single language (Korean), while at the other end of the spectrum one can remove the constraint on reasoning language altogether \citep{gao2026explangimprovedexplorationexploitation}. \citet{park2026crosslingualcollapselanguagecentricfoundation} show that under verifiable-reward RL the CoT collapses toward English as accuracy rises, that the collapse is severe for lower-resource languages and largely irreversible, and that a language-consistency reward mitigates the drift only at a measurable accuracy cost.

\citet{huang2026beyond} report that enforcing language consistency hurts crosslingual generalization, and \citet{barua2026long} train separate models per dataset and language and do not examine mixing. Both establish their result in the per-language specialist regime; \citet{yang2025parallelscalinglawunveiling} point the other way, establishing a parallel scaling law under which jointly training multiple reasoning languages at once has a diminishing but beneficial effect on crosslingual generalization. Since the regime, not the objective, is what separates these results, we vary language coverage directly---one language, one region, all languages---holding everything else fixed, and find joint training transfers better than per-region specialists (\S\ref{sec:lang_scale}), extending \citeauthor{yang2025parallelscalinglawunveiling}'s observation from RL to SFT.

A third perspective argues that the multilingual reasoning gap is not really about the reasoning language at all. \citet{kang-etal-2026-multilingual} attribute it to a failure to comprehend the source language, and show that translating only the inputs a model is likely to misunderstand recovers most of the benefit of translating everything; \citet{ko-etal-2025-understand} reach a similar conclusion for Korean mathematics, using English as an anchor for solving before translating back. Together these motivate why L2 reasoning is not free, and why we need to track and optimize task accuracy alongside L2 reasoning rate throughout.

Finally, a line of work changes the reasoning language by recomposing separately trained models in parameter space, rather than by training one model on mixed data. \citet{lasbordes2026layerswap} extend the layer-swapping idea of \S\ref{sec:related_enreas} to long-CoT reasoning models, transferring a contiguous block of an English reasoning specialist into a native specialist trained from the same base. Following this line of work, we merge our own per-language specialists and find that averaging cancels each specialist's language conditioning, collapsing the trace back to English at otherwise competitive accuracy (\S\ref{sec:datamix}).

\subsection{Existing reasoning data}\label{sec:related-work-reasoning-data}
The English bias becomes concrete in the language composition of the datasets that drive open reasoning models. The most widely adopted corpora contain reasoning traces exclusively in English: OpenThoughts3~\citep{guha2025openthoughts} and the first release of NVIDIA's Nemotron post-training data~\citep{NemotronPostTrainingDatasetV1, bercovich2025llamanemotronefficientreasoningmodels} are English-only. Tellingly, where multilingual data has been introduced, localization typically stops short of the reasoning trace itself: the v2 Nemotron release translates prompts and responses into five languages (Spanish, French, German, Italian, and Japanese) yet leaves the chain-of-thought, the very component that determines whether a user can follow the model's reasoning, in English~\citep{nvidia2025nvidianemotronnano2}. The same asymmetry holds for frontier models: DeepSeek-R1 is optimized for Chinese and English and may revert to English reasoning for queries in other languages~\citep{deepseek-r1}, whereas Qwen3 does not release its post-training data, leaving the language composition of its reasoning traces undisclosed altogether~\citep{qwen3technicalreport}. Only a handful of resources supervise reasoning directly in the target language: the M-Thinker SFT set spans five languages~\citep{m-thinker}, and Lightblue's multilingual R1 corpus provides traces in roughly thirty~\citep{lightbluedataset}, whose resulting models we adopt as a baseline. Non-English reasoning traces thus remain scarce, and even the efforts that localize the surrounding data most often leave the reasoning itself in English. We reduce this gap by releasing our multilingual reasoning data covering 44 languages besides English and three domains.

\section{Conclusion}
Reasoning is becoming a central capability of language models, yet the development of reasoning systems has remained overwhelmingly English-centric.
In this work, we ask how we can optimize multilingual reasoning with minimal in-language reasoning supervision. Although dual-mode training is not the default mode in many existing large reasoning models, we show that pairing a small fraction of non-reasoning data with a large enough English reasoning dataset is a crucial pillar for achieving multilingual reasoning despite the scarcity of multilingual reasoning data.

Taken together, these findings point to a different way of thinking about multilingual reasoning.
\textbf{Reasoning capability} and \textbf{reasoning language} are related but separable: a model can acquire the capability to solve complex problems from large-scale reasoning supervision while learning the language in which that reasoning is expressed through multilingual supervision.
This distinction suggests that multilingual reasoning need not be built language by language.
Instead, reasoning language can be treated as a transferable behavioral property that can be learned from a relatively small amount of multilingual reasoning data, reinforced through broader multilingual instruction, and generalized to languages without direct reasoning supervision.
This provides a more scalable path toward reasoning systems that are not only capable across languages, but also able to \textit{reason in the language of the people who use them}.

\section{Limitations}

\textbf{Dependence on translated reasoning traces.} Our multilingual reasoning supervision is created by translating English reasoning traces rather than creating target-language original reasoning from multilingual teachers or human annotators.
While we filter for translation quality, this pipeline inevitably anchors reasoning style, cultural framing, and problem decomposition to English templates.
For low-resource languages translation quality degrades and may not reflect how speakers naturally approach multi-step deduction. 

\textbf{Evaluation through automatic proxies.} We measure L2 reasoning rate and open-ended generation quality automatically and do not conduct human evaluations of reasoning trace quality, fluency, or cultural appropriateness. 
Consequently, our metrics capture linguistic compliance rather than the depth, coherence, or usefulness of the reasoning itself.

\textbf{Prompt sensitivity.} Our reported L2 reasoning rates rely on prepending an explicit instruction to the user prompt (\textit{``Think in the same language as the prompt''}) that the model sees in training, and without this trigger the model may revert to English.
This sensitivity implies that deployment contexts with constrained system prompting, ambiguous language signals, or mixed-language inputs may see degraded performance.

\section*{Acknowledgements}
We thank David Stap for his thoughtful feedback on the paper, and Maximilian Mozes, Ammar Khairi, Kelly Marchisio, and Nikita Moghe for early discussions that informed this work. 
We are also grateful to Sara Rajaee and 
Saurabh Dash for their technical assistance; and to David Cairuz, Sammie Bae, Diana Abagyan, Björn Bebensee, Sylvie Shi and Pierre Beaulieu for their support on the long context extension procedure.

\bibliography{main}

\newpage

\appendix
\onecolumn
\newpage
\section{Training}\label{app:training}
\subsection{Long Context Extension} 
We extend the context length of the \tinyaya{} Base model \citep{salamanca2026tinyayabridgingscale} from 8K to 32K in a single training stage. We continue pretraining for 12,000 steps on an interleaved mixture of 8K and 32K token sequences in a 3:1 ratio, using a linear learning rate schedule with an initial rate of \(1.25 \times 10^{-4}\). For the longer-context data, we evenly distributed the training mixture across three 8K-token context-length buckets spanning 8K to 32K to support a stable transition to long-context modeling.

\subsection{SFT}
\tinyayamulti{} was finetuned from \tinyaya{} base with a standard next-token cross-entropy objective on 32 NVIDIA H100 GPUs for 40 hours using fully sharded data parallelism. Training runs for 4 epochs over the packed SFT mixture  ($\approx$ 15.7B tokens per epoch) with a global batch size of 32. We used the Adam optimizer ($\beta_1=0.9$, $\beta_2=0.95$) with additive weight decay 0.1 and gradient clipping at 1.0. The learning rate followed a cosine decay schedule with a short 10-step warmup, peaking at $1.25\times 10^{-4}$ and annealing to $1.25\times 10^{-5}$.

\newcolumntype{L}{>{\raggedright\arraybackslash}X}

\begin{table}[h]
\centering
\small
\setlength{\tabcolsep}{5pt}
\renewcommand{\arraystretch}{1.3}
\begin{tabularx}{\textwidth}{@{}llrL@{}}
\toprule
\textbf{Benchmark} & \textbf{Reasoning category} & \textbf{\#Lang.} & \textbf{Languages} \\
\midrule
MGSM & Math reasoning & 35 & \texttt{en}, \texttt{bn}, \texttt{de}, \texttt{es}, \texttt{fr}, \texttt{ja}, \texttt{ru}, \texttt{sw}, \texttt{te}, \texttt{th}, \texttt{zh}, \texttt{ar}, \texttt{ca}, \texttt{cs}, \texttt{cy}, \texttt{el}, \texttt{eu}, \texttt{gl}, \texttt{hu}, \texttt{ko}, \texttt{sr}, \texttt{vi}, \texttt{amh}, \texttt{hau}, \texttt{sna}, \texttt{wol}, \texttt{xho}, \texttt{yor}, \texttt{zul}, \texttt{ur}, \texttt{hi}, \texttt{gu}, \texttt{km}, \texttt{my}, \texttt{ta} \\
\midrule
PolyMath & Math reasoning & 18 & \texttt{en}, \texttt{zh}, \texttt{ar}, \texttt{bn}, \texttt{de}, \texttt{es}, \texttt{fr}, \texttt{id}, \texttt{it}, \texttt{ja}, \texttt{ko}, \texttt{ms}, \texttt{pt}, \texttt{ru}, \texttt{sw}, \texttt{te}, \texttt{th}, \texttt{vi} \\
\midrule
GlobalPIQA & Commonsense reasoning & 59 & \texttt{eng\_latn}, \texttt{amh\_ethi}, \texttt{arb\_arab}, \texttt{ben\_beng}, \texttt{bul\_cyrl}, \texttt{cat\_latn}, \texttt{ces\_latn}, \texttt{cmn\_hans}, \texttt{cmn\_hant}, \texttt{deu\_latn}, \texttt{ekk\_latn}, \texttt{ell\_grek}, \texttt{fin\_latn}, \texttt{fra\_latn\_cana}, \texttt{fra\_latn\_fran}, \texttt{glg\_latn}, \texttt{guj\_gujr}, \texttt{hau\_latn}, \texttt{heb\_hebr}, \texttt{hin\_deva}, \texttt{hrv\_latn}, \texttt{hun\_latn}, \texttt{ibo\_latn}, \texttt{ind\_latn}, \texttt{ita\_latn}, \texttt{jav\_latn}, \texttt{jpn\_jpan}, \texttt{kor\_hang}, \texttt{lit\_latn}, \texttt{mar\_deva}, \texttt{nld\_latn}, \texttt{nob\_latn}, \texttt{pes\_arab}, \texttt{pol\_latn}, \texttt{por\_latn\_braz}, \texttt{por\_latn\_port}, \texttt{ron\_latn}, \texttt{rus\_cyrl}, \texttt{slk\_latn}, \texttt{slk\_latn\_sari}, \texttt{slv\_latn}, \texttt{slv\_latn\_cerk}, \texttt{spa\_latn\_mexi}, \texttt{spa\_latn\_peru}, \texttt{spa\_latn\_spai}, \texttt{srp\_cyrl}, \texttt{swe\_latn}, \texttt{swh\_latn}, \texttt{tam\_taml}, \texttt{tel\_telu}, \texttt{tgl\_latn}, \texttt{tha\_thai}, \texttt{tur\_latn}, \texttt{ukr\_cyrl}, \texttt{urd\_arab}, \texttt{vie\_latn}, \texttt{yor\_latn}, \texttt{zsm\_latn}, \texttt{zul\_latn} \\
\midrule
\mif{} & Instruction-following reasoning & 29 & \texttt{en}, \texttt{ar}, \texttt{bn}, \texttt{cs}, \texttt{de}, \texttt{el}, \texttt{es}, \texttt{fr}, \texttt{he}, \texttt{hu}, \texttt{id}, \texttt{it}, \texttt{ja}, \texttt{ko}, \texttt{ms}, \texttt{ne}, \texttt{nl}, \texttt{pl}, \texttt{pt}, \texttt{ro}, \texttt{ru}, \texttt{sw}, \texttt{th}, \texttt{tr}, \texttt{uk}, \texttt{ur}, \texttt{vi}, \texttt{yo}, \texttt{zh} \\
\midrule
\mist{} & Open-ended generation & 25 & \texttt{en}, \texttt{ar}, \texttt{bn}, \texttt{cs}, \texttt{de}, \texttt{el}, \texttt{et}, \texttt{fa}, \texttt{hi}, \texttt{hr}, \texttt{id}, \texttt{it}, \texttt{ja}, \texttt{ko}, \texttt{lt}, \texttt{mr}, \texttt{ro}, \texttt{ru}, \texttt{sr}, \texttt{sv}, \texttt{th}, \texttt{tr}, \texttt{uk}, \texttt{vi}, \texttt{zh} \\
\midrule
\macaron{} & Cultural reasoning & 20 & \texttt{pt\_BR}, \texttt{zh\_CN}, \texttt{ar\_EGY}, \texttt{am}, \texttt{ka}, \texttt{el}, \texttt{hi}, \texttt{id}, \texttt{it}, \texttt{ja}, \texttt{ky}, \texttt{es\_MX}, \texttt{ar\_MAR}, \texttt{yo}, \texttt{tl}, \texttt{zu}, \texttt{th}, \texttt{ar\_TUN}, \texttt{tr}, \texttt{ar\_YEM} \\
\bottomrule
\end{tabularx}
\caption{Benchmarks used in our evaluation suite and the languages or language varieties included for each. Counts include English. \macaron{} covers 20 country-language varieties corresponding to 17 base languages. PolyMath is evaluated at three difficulty levels (medium, high, top) over the same language set.}
\label{tab:benchmarks-stats}
\end{table}

\definecolor{cohereblue}{HTML}{2D4CB9}
\definecolor{cohereink}{HTML}{111111}
\definecolor{coheremuted}{HTML}{555555}
\definecolor{coherefill}{HTML}{F4F6F8}
\definecolor{cohereline}{HTML}{D8D8D8}

\newcommand{\headcell}[1]{\textcolor{coheremuted}{\footnotesize\bfseries #1}}
\newcommand{\regionrow}[1]{%
  \multicolumn{6}{@{}l}{%
    \textcolor{cohereblue}{\footnotesize\bfseries\textls[80]{\MakeUppercase{#1}}}%
  } \\[1.5pt]}
\newcommand{\code}[1]{\textcolor{coheremuted}{\ttfamily\footnotesize #1}}
\newcommand{\total}[1]{\textcolor{cohereblue}{\bfseries #1}}
\begin{table*}[t]
\centering
\setlength{\tabcolsep}{7pt}
\renewcommand{\arraystretch}{1.02}
\small
\begin{tabular}{@{}llrrrr@{}}
\toprule
\headcell{Language} & \headcell{Code} & \headcell{Math} & \headcell{Science}
  & \headcell{General} & \headcell{Total} \\
\midrule
\regionrow{Europe}
Basque & \code{eu} & 827 & 1,535 & 1,283 & \total{3,645} \\
\rowcolor{coherefill}Bulgarian & \code{bg} & 879 & 2,078 & 1,672 & \total{4,629} \\
Catalan & \code{ca} & 1,043 & 2,076 & 1,644 & \total{4,763} \\
\rowcolor{coherefill}Czech & \code{cs} & 1,284 & 2,001 & 1,582 & \total{4,867} \\
Finnish & \code{fi} & 1,095 & 1,991 & 660 & \total{3,746} \\
\rowcolor{coherefill}French & \code{fr} & 8,494 & 6,258 & 9,103 & \total{23,855} \\
German & \code{de} & 7,890 & 6,393 & 8,461 & \total{22,744} \\
\rowcolor{coherefill}Greek & \code{el} & 1,185 & 2,123 & 1,604 & \total{4,912} \\
Hungarian & \code{hu} & 701 & 1,843 & 1,505 & \total{4,049} \\
\rowcolor{coherefill}Irish & \code{ga} & 907 & 1,926 & 1,604 & \total{4,437} \\
Italian & \code{it} & 1,252 & 1,933 & 1,428 & \total{4,613} \\
\rowcolor{coherefill}Lithuanian & \code{lt} & 946 & 1,835 & 1,527 & \total{4,308} \\
Norwegian & \code{no} & 1,353 & 2,173 & 1,491 & \total{5,017} \\
\rowcolor{coherefill}Polish & \code{pl} & 763 & 1,204 & 946 & \total{2,913} \\
Russian & \code{ru} & 1,288 & 2,210 & 1,646 & \total{5,144} \\
\rowcolor{coherefill}Slovak & \code{sk} & 1,163 & 1,975 & 1,626 & \total{4,764} \\
Ukrainian & \code{uk} & 1,104 & 2,134 & 1,382 & \total{4,620} \\
\addlinespace[3pt]
\regionrow{Asia-Pacific}
Chinese & \code{zh} & 1,130 & 2,106 & 1,556 & \total{4,792} \\
\rowcolor{coherefill}Filipino & \code{fil} & 861 & 2,040 & 287 & \total{3,188} \\
Indonesian & \code{id} & 1,507 & 1,969 & 1,431 & \total{4,907} \\
\rowcolor{coherefill}Japanese & \code{ja} & 5,771 & 5,927 & 11,026 & \total{22,724} \\
Javanese & \code{jv} & 1,913 & 1,579 & 2,307 & \total{5,799} \\
\rowcolor{coherefill}Khmer & \code{km} & 540 & 1,402 & 1,899 & \total{3,841} \\
Korean & \code{ko} & 6,454 & 6,249 & 8,332 & \total{21,035} \\
\rowcolor{coherefill}Malay & \code{ms} & 1,430 & 1,976 & 159 & \total{3,565} \\
Thai & \code{th} & 1,162 & 2,107 & 1,267 & \total{4,536} \\
\rowcolor{coherefill}Vietnamese & \code{vi} & 1,259 & 2,156 & 1,704 & \total{5,119} \\
\addlinespace[3pt]
\regionrow{West Asia}
Arabic & \code{ar} & 7,963 & 6,273 & 11,270 & \total{25,506} \\
\rowcolor{coherefill}Hebrew & \code{he} & 1,105 & 2,075 & 1,309 & \total{4,489} \\
Maltese & \code{mt} & 676 & 1,436 & 1,081 & \total{3,193} \\
\rowcolor{coherefill}Persian & \code{fa} & 1,148 & 2,036 & 1,333 & \total{4,517} \\
Turkish & \code{tr} & 931 & 1,883 & 1,019 & \total{3,833} \\
\addlinespace[3pt]
\regionrow{Africa}
Amharic & \code{am} & 630 & 1,240 & 2,385 & \total{4,255} \\
\rowcolor{coherefill}Hausa & \code{ha} & 1,630 & 1,371 & 2,050 & \total{5,051} \\
Igbo & \code{ig} & 1,544 & 1,473 & 2,157 & \total{5,174} \\
\rowcolor{coherefill}Swahili & \code{sw} & 1,213 & 1,619 & 2,092 & \total{4,924} \\
Yoruba & \code{yo} & 1,102 & 1,246 & 2,118 & \total{4,466} \\
\rowcolor{coherefill}Zulu & \code{zu} & 1,622 & 1,139 & 1,677 & \total{4,438} \\
\addlinespace[3pt]
\regionrow{South Asia}
Bengali & \code{bn} & 923 & 2,061 & 1,241 & \total{4,225} \\
\rowcolor{coherefill}Hindi & \code{hi} & 1,081 & 2,019 & 1,462 & \total{4,562} \\
Punjabi & \code{pa} & 966 & 1,947 & 978 & \total{3,891} \\
\rowcolor{coherefill}Tamil & \code{ta} & 824 & 2,047 & 947 & \total{3,818} \\
Telugu & \code{te} & 869 & 2,062 & 942 & \total{3,873} \\
\rowcolor{coherefill}Urdu & \code{ur} & 869 & 1,885 & 887 & \total{3,641} \\
\midrule
\textbf{Total} & & \textbf{79,297} & \textbf{103,011}
  & \textbf{104,080} & \textbf{286,388} \\
\bottomrule
\end{tabular}
\caption{Number of samples for our translated multilingual reasoning data per language and domain, separated by regions.}
\label{tab:translation-stats}
\end{table*}

\begin{table*}[t]
\centering
\small
\renewcommand{\arraybackslash}{}
\begin{tabular}{@{}llll@{}}
\toprule
\textbf{Region} & \textbf{Benchmark} & \textbf{Seen (trained)} & \textbf{Unseen (held-out)} \\
\midrule
\multirow{3}{*}{Europe}
  & MGSM          & German, French            & Spanish, Welsh \\
  & \mif{} & German, French            & Czech, Spanish \\
  & GlobalPIQA   & German, French            & Czech, Spanish \\
\midrule
\multirow{3}{*}{Asia-Pacific}
  & MGSM          & Japanese, Korean          & Chinese, Thai \\
  & \mif{} & Japanese, Korean          & Thai, Chinese \\
  & GlobalPIQA   & Japanese, Korean          & Chinese, Thai \\
\midrule
\multirow{3}{*}{Africa}
  & MGSM          & Swahili                   & Yoruba \\
  & \mif{} & Swahili                   & Yoruba \\
  & GlobalPIQA   & Swahili                   & Yoruba \\
\midrule
\multirow{3}{*}{South Asia}
  & MGSM          & Hindi, Bengali            & Tamil \\
  & \mif{} & Bengali                   & Urdu \\
  & GlobalPIQA   & Hindi, Bengali            & Tamil, Telugu \\
\midrule
\multirow{3}{*}{West Asia}
  & MGSM          & Arabic                    & Urdu \\
  & \mif{} & Arabic                    & Hebrew, Turkish \\
  & GlobalPIQA   & Arabic, Persian           & Hebrew, Turkish \\
\bottomrule
\end{tabular}
\caption{Seen (present in L2 training) versus unseen (held-out) evaluation
languages, by region and benchmark for experiments with our ten selected training languages;
each benchmark covers a different subset, so we report a region-balanced
seen/unseen split per benchmark. 
} 
\label{tab:seen-unseen}
\end{table*}

\section{Translation Details}\label{app:translation} 

\subsection{Filtering} From the AM thinking dataset \citep{ji2025amthinking} we remove sources that are particularly susceptible to translation artifacts: We require the prompt, reasoning trace, and response to be consistently identified as English, and remove examples containing intra-document code-switching.
We additionally remove trajectories that explicitly discuss translation or name target languages (e.g., \textit{translat, tradu, übersetz}) since translating such content can introduce inconsistencies.
Finally, we remove prompts containing constraints that are difficult to preserve reliably under translation such as exact word counts, length bounds, and capitalization or formatting requirements. 
Language id checks are performed with FastText and GlotLID as a fallback for any language FastText does not cover.

\subsection{Translated Reasoning Data}
\Cref{tab:translation-stats} details the number of translated reasoning examples across domains and 44 languages besides English included in the training of \tinyayamulti{} model, and \Cref{fig:MR-data-composition} illustrates the composition of this data across regions (Europe, Asia-Pacific, West Asia, Africa, South Asia) and domains (Math, Science, General). Please note that for the controlled experiments in \Cref{sec:lang_scale} and \Cref{sec:NR-sweep}, we cap the number of samples at 5K per language; however, in the final \tinyayamulti{} model we use all of our available translated data.

\section{Benchmarking Details}\label{app:benchmark_details}

For Figures \ref{fig:region-avg} and \ref{fig:nr-sweep}, we report results on a subset of seen and unseen languages. Seen languages are those covered by the benchmark that also appear among our ten training languages used in \Cref{sec:lang_scale} and \ref{sec:NR-sweep}; unseen languages are held-out languages covered by the benchmark but absent from training. For each benchmark we select at least one and at most two unseen languages per region to keep regional coverage balanced. Where a region has no natural unseen candidate in a given benchmark, we substitute the closest available language: Urdu stands in for West Asia in MGSM (as the nearest relative of Persian among covered languages) and for South Asia in \mif{}, since no other unseen language from those regions is covered. The full split is given in \Cref{tab:seen-unseen}.

\subsection{Language Coverage}\label{app:benchmark-languages}

Table~\ref{tab:benchmarks-stats} lists the languages from each benchmark covered for our main experiments in \Cref{sec:results}. Note that this is a subset of the original benchmark languages, restricted to those supported by \tinyaya{} and identifiable by either FastText or GlotLID. These include all languages used in the training of \mthinker{}. \qwensmallnew{} supports 201 languages, but it is not disclosed which ones, so potentially there are some languages included here that it does not support.

For PolyMath, \mif{}, \mist{}, and \macaron, 90+\% of all languages from the original benchmarks are covered in Table~\ref{tab:benchmarks-stats}. For MGSM and GlobalPIQA, we evaluate on a subset, though it still spans a diverse range of high-resource and low-resource languages.

\section{Decoding Settings}\label{app:decoding}
We generated a single completion per example, and separated the reasoning trace from the final answer according to each model's output convention. \tinyayamulti{} and \tinyayaen{} encloses reasoning within \verb+<|START_THINKING|>+ and \verb+<|END_THINKING|>+. \qwensmallnew{}, and \mthinker{} use the \verb+<think>+\dots\verb+</think>+ convention; Magistral deployment represents reasoning using \verb+[THINK]+\dots\verb+[/THINK]+, and we use the system preamble recommended in the Magistral model card\footnote{\url{https://huggingface.co/mistralai/Magistral-Small-2506}}. All models were evaluated using a 32K-token context window.

\section{Benchmark results by language}

Tables~\ref{tab:app-mgsm} to \ref{tab:app-macaron_mcq} include per-language task accuracy and L2 reasoning rate of all models in Table~\ref{tab:main-results} for each benchmark, accompanied by the mean and std over languages. PolyMath numbers are reported separately across the medium, top, and high levels. In \Cref{sec:results}, PolyMath scores report a weighted average over the levels as $(2\,\text{medium} + 4\,\text{high} + 8\,\text{top}) / 14$.
\begin{table}[h]
\centering
\caption{Per-language task accuracy (Acc) and L2 reasoning rate (L2\%) on \textbf{MGSM} (35 languages, 0–100). Languages are ordered alphabetically by language code. Avg / Std are over languages; the highest average in each metric is bold.}
\scriptsize{%
\setlength{\tabcolsep}{2pt}
\resizebox{\textwidth}{!}{%
\begin{tabular}{l|cccc|cccccccccc}
\toprule
Language & \multicolumn{2}{c}{\tinyayaen{}} & \multicolumn{2}{c|}{\tinyayamulti{}} & \multicolumn{2}{c}{\qwensmallnew{}} & \multicolumn{2}{c}{\begin{tabular}{@{}c@{}}\qwensmallnew{}\\{\scriptsize(user prefix LF)}\end{tabular}} & \multicolumn{2}{c}{\begin{tabular}{@{}c@{}}\qwensmallnew{}\\{\scriptsize(thinking prefix LF)}\end{tabular}} & \multicolumn{2}{c}{\mthinker{}} & \multicolumn{2}{c}{\magistralsmall{}} \\
\cmidrule(lr){2-3}\cmidrule(lr){4-5}\cmidrule(lr){6-7}\cmidrule(lr){8-9}\cmidrule(lr){10-11}\cmidrule(lr){12-13}\cmidrule(lr){14-15}
 & Acc & L2\% & Acc & L2\% & Acc & L2\% & Acc & L2\% & Acc & L2\% & Acc & L2\% & Acc & L2\% \\
\midrule
amh & $61.2$ & $0.0$ & $54.8$ & $99.6$ & $51.8$ & $0.8$ & $51.2$ & $1.2$ & $41.2$ & $98.4$ & $4.8$ & $92.0$ & $8.0$ & $0.4$ \\
ar & $77.6$ & $0.0$ & $77.2$ & $100.0$ & $77.9$ & $52.6$ & $76.0$ & $60.8$ & $82.0$ & $99.6$ & $70.0$ & $100.0$ & $78.8$ & $1.2$ \\
bn & $60.0$ & $0.0$ & $63.2$ & $100.0$ & $49.2$ & $56.9$ & $47.2$ & $63.6$ & $3.2$ & $100.0$ & $47.5$ & $100.0$ & $51.4$ & $0.8$ \\
ca & $88.4$ & $0.0$ & $77.6$ & $97.2$ & $58.6$ & $15.2$ & $57.6$ & $13.6$ & $80.8$ & $100.0$ & $67.1$ & $99.6$ & $61.6$ & $68.8$ \\
cs & $78.8$ & $0.0$ & $71.6$ & $100.0$ & $76.5$ & $38.1$ & $71.5$ & $45.4$ & $71.6$ & $99.2$ & $69.1$ & $100.0$ & $81.6$ & $98.0$ \\
cy & $80.0$ & $0.0$ & $75.6$ & $87.8$ & $65.6$ & $0.8$ & $68.8$ & $3.2$ & $34.8$ & $98.4$ & $15.6$ & $99.2$ & $80.3$ & $23.8$ \\
de & $88.8$ & $0.0$ & $87.2$ & $100.0$ & $80.0$ & $41.6$ & $81.2$ & $46.2$ & $94.0$ & $100.0$ & $89.6$ & $100.0$ & $97.5$ & $100.0$ \\
el & $78.4$ & $0.0$ & $79.2$ & $99.6$ & $75.6$ & $71.5$ & $78.8$ & $78.4$ & $87.6$ & $100.0$ & $39.6$ & $100.0$ & $92.3$ & $97.6$ \\
en & $92.8$ & $100.0$ & $93.6$ & $100.0$ & $96.4$ & $100.0$ & $93.8$ & $99.6$ & $92.0$ & $100.0$ & $84.8$ & $97.6$ & $98.0$ & $99.6$ \\
es & $87.6$ & $0.0$ & $78.8$ & $100.0$ & $80.0$ & $58.4$ & $76.2$ & $65.7$ & $89.2$ & $95.2$ & $77.9$ & $100.0$ & $84.2$ & $96.8$ \\
eu & $69.6$ & $0.4$ & $63.6$ & $99.6$ & $58.6$ & $2.9$ & $50.0$ & $3.6$ & $30.8$ & $96.4$ & $22.1$ & $99.2$ & $73.1$ & $41.8$ \\
fr & $84.4$ & $0.0$ & $73.6$ & $100.0$ & $67.8$ & $19.2$ & $76.2$ & $25.0$ & $83.2$ & $97.6$ & $73.6$ & $100.0$ & $86.3$ & $95.2$ \\
gl & $82.0$ & $0.0$ & $75.6$ & $73.4$ & $70.8$ & $0.8$ & $61.6$ & $1.6$ & $84.0$ & $90.4$ & $72.2$ & $9.3$ & $75.2$ & $18.8$ \\
gu & $79.6$ & $0.0$ & $72.0$ & $100.0$ & $62.0$ & $62.8$ & $62.9$ & $63.3$ & $55.6$ & $100.0$ & $28.4$ & $100.0$ & $83.6$ & $1.6$ \\
hau & $60.0$ & $0.0$ & $56.8$ & $99.6$ & $28.1$ & $18.6$ & $27.6$ & $15.2$ & $9.6$ & $90.4$ & $0.8$ & $34.4$ & $12.0$ & $0.8$ \\
hi & $76.8$ & $0.0$ & $74.0$ & $100.0$ & $68.3$ & $36.5$ & $69.2$ & $42.9$ & $68.0$ & $99.6$ & $1.2$ & $100.0$ & $82.8$ & $0.0$ \\
hu & $69.6$ & $0.0$ & $66.8$ & $100.0$ & $60.3$ & $44.5$ & $64.4$ & $56.0$ & $73.6$ & $100.0$ & $61.6$ & $100.0$ & $76.8$ & $41.2$ \\
ja & $75.6$ & $0.0$ & $69.6$ & $100.0$ & $82.5$ & $26.1$ & $79.6$ & $36.3$ & $83.2$ & $100.0$ & $63.0$ & $100.0$ & $86.7$ & $6.7$ \\
km & $66.4$ & $0.0$ & $60.4$ & $100.0$ & $29.2$ & $17.3$ & $20.0$ & $15.6$ & $14.8$ & $100.0$ & $11.6$ & $100.0$ & $4.1$ & $3.5$ \\
ko & $68.4$ & $0.0$ & $72.4$ & $99.2$ & $62.3$ & $41.0$ & $67.6$ & $51.6$ & $86.4$ & $100.0$ & $62.0$ & $100.0$ & $80.0$ & $36.8$ \\
my & $58.8$ & $0.0$ & $48.4$ & $100.0$ & $4.5$ & $42.4$ & $5.6$ & $37.0$ & $1.2$ & $99.6$ & $12.0$ & $100.0$ & $53.2$ & $6.4$ \\
ru & $82.8$ & $0.0$ & $87.2$ & $100.0$ & $84.6$ & $43.7$ & $80.2$ & $58.0$ & $97.6$ & $100.0$ & $88.8$ & $100.0$ & $96.2$ & $97.5$ \\
sna & $48.8$ & $0.8$ & $47.6$ & $99.6$ & $2.5$ & $24.4$ & $2.8$ & $24.4$ & $0.0$ & $94.0$ & $2.0$ & $69.9$ & $12.4$ & $2.4$ \\
sr & $78.0$ & $0.0$ & $73.2$ & $94.5$ & $68.0$ & $39.7$ & $61.2$ & $45.6$ & $72.0$ & $98.0$ & $2.8$ & $86.3$ & $77.6$ & $96.0$ \\
sw & $82.0$ & $0.0$ & $80.8$ & $80.7$ & $61.5$ & $0.0$ & $62.2$ & $0.4$ & $27.6$ & $65.6$ & $4.4$ & $36.8$ & $84.4$ & $8.2$ \\
ta & $75.6$ & $0.0$ & $77.2$ & $100.0$ & $22.2$ & $58.9$ & $21.2$ & $58.0$ & $29.2$ & $99.2$ & $11.7$ & $100.0$ & $87.6$ & $2.8$ \\
te & $75.2$ & $0.0$ & $75.6$ & $100.0$ & $46.9$ & $69.5$ & $40.2$ & $68.4$ & $20.0$ & $100.0$ & $19.3$ & $100.0$ & $60.8$ & $50.8$ \\
th & $78.0$ & $0.0$ & $67.6$ & $100.0$ & $35.1$ & $41.3$ & $36.8$ & $44.8$ & $46.8$ & $98.0$ & $75.9$ & $100.0$ & $86.7$ & $8.8$ \\
ur & $82.8$ & $0.4$ & $81.6$ & $100.0$ & $28.1$ & $0.4$ & $24.9$ & $0.8$ & $64.4$ & $100.0$ & $52.4$ & $100.0$ & $89.2$ & $0.4$ \\
vi & $77.2$ & $12.0$ & $78.8$ & $100.0$ & $67.5$ & $65.0$ & $67.6$ & $68.8$ & $82.0$ & $99.2$ & $53.0$ & $100.0$ & $88.4$ & $29.6$ \\
wol & $17.2$ & $14.0$ & $20.8$ & $99.6$ & $4.6$ & $11.2$ & $4.4$ & $6.4$ & $3.6$ & $64.0$ & $1.2$ & $57.0$ & $7.2$ & $2.0$ \\
xho & $46.4$ & $0.0$ & $43.2$ & $99.6$ & $21.2$ & $29.4$ & $17.7$ & $25.7$ & $2.9$ & $93.8$ & $2.1$ & $87.1$ & $21.7$ & $4.0$ \\
yor & $50.0$ & $0.0$ & $54.0$ & $52.0$ & $25.9$ & $0.8$ & $26.4$ & $0.0$ & $1.2$ & $44.3$ & $1.2$ & $32.2$ & $5.6$ & $0.0$ \\
zh & $73.6$ & $0.0$ & $76.8$ & $98.8$ & $79.7$ & $1.2$ & $84.4$ & $3.2$ & $95.2$ & $100.0$ & $84.4$ & $100.0$ & $94.6$ & $99.2$ \\
zul & $48.8$ & $0.0$ & $48.4$ & $99.5$ & $17.8$ & $19.0$ & $16.4$ & $22.0$ & $4.0$ & $83.9$ & $3.7$ & $82.9$ & $19.6$ & $0.4$ \\
\midrule
\textbf{Avg} & $\mathbf{71.5}$ & $3.6$ & $68.7$ & $\mathbf{96.6}$ & $53.5$ & $32.9$ & $52.4$ & $35.8$ & $51.8$ & $94.4$ & $39.4$ & $88.1$ & $65.1$ & $35.5$ \\
Std & $15.2$ & $16.8$ & $14.5$ & $9.5$ & $25.4$ & $25.1$ & $25.8$ & $26.8$ & $34.6$ & $12.0$ & $32.1$ & $23.7$ & $31.3$ & $40.1$ \\
\bottomrule
\end{tabular}%
}
}
\label{tab:app-mgsm}
\end{table}

\begin{table}[h]
\centering
\caption{Per-language task accuracy (Acc) and L2 reasoning rate (L2\%) on \textbf{PolyMath (medium)} (18 languages, 0–100). Languages are ordered alphabetically by language code. Avg / Std are over languages; the highest average in each metric is bold.}
\scriptsize{%
\setlength{\tabcolsep}{2pt}
\resizebox{\textwidth}{!}{%
\begin{tabular}{l|cccc|cccccccccc}
\toprule
Language & \multicolumn{2}{c}{\tinyayaen{}} & \multicolumn{2}{c|}{\tinyayamulti{}} & \multicolumn{2}{c}{\qwensmallnew{}} & \multicolumn{2}{c}{\begin{tabular}{@{}c@{}}\qwensmallnew{}\\{\scriptsize(user prefix LF)}\end{tabular}} & \multicolumn{2}{c}{\begin{tabular}{@{}c@{}}\qwensmallnew{}\\{\scriptsize(thinking prefix LF)}\end{tabular}} & \multicolumn{2}{c}{\mthinker{}} & \multicolumn{2}{c}{\magistralsmall{}} \\
\cmidrule(lr){2-3}\cmidrule(lr){4-5}\cmidrule(lr){6-7}\cmidrule(lr){8-9}\cmidrule(lr){10-11}\cmidrule(lr){12-13}\cmidrule(lr){14-15}
 & Acc & L2\% & Acc & L2\% & Acc & L2\% & Acc & L2\% & Acc & L2\% & Acc & L2\% & Acc & L2\% \\
\midrule
ar & $55.2$ & $0.0$ & $39.5$ & $96.8$ & $77.2$ & $0.0$ & $78.4$ & $0.8$ & $62.4$ & $96.8$ & $63.4$ & $85.4$ & $60.0$ & $0.0$ \\
bn & $49.6$ & $0.8$ & $26.8$ & $99.2$ & $66.4$ & $0.8$ & $67.2$ & $0.0$ & $43.2$ & $92.0$ & $50.4$ & $100.0$ & $22.4$ & $2.4$ \\
de & $50.8$ & $0.0$ & $39.5$ & $98.4$ & $79.0$ & $0.0$ & $73.6$ & $0.8$ & $72.0$ & $95.2$ & $65.6$ & $98.4$ & $31.2$ & $96.8$ \\
en & $52.4$ & $100.0$ & $55.0$ & $99.2$ & $64.0$ & $100.0$ & $56.3$ & $99.2$ & $66.4$ & $100.0$ & $71.0$ & $91.1$ & $53.6$ & $100.0$ \\
es & $55.2$ & $0.8$ & $40.0$ & $89.6$ & $79.8$ & $0.0$ & $76.8$ & $0.0$ & $72.8$ & $91.2$ & $68.8$ & $100.0$ & $17.9$ & $87.8$ \\
fr & $49.6$ & $0.0$ & $40.8$ & $98.4$ & $75.2$ & $0.0$ & $75.2$ & $0.0$ & $72.8$ & $96.8$ & $66.1$ & $99.2$ & $15.3$ & $58.9$ \\
id & $51.6$ & $0.8$ & $45.6$ & $95.2$ & $76.6$ & $0.0$ & $77.6$ & $0.0$ & $72.8$ & $72.0$ & $65.3$ & $92.7$ & $25.6$ & $19.2$ \\
it & $48.0$ & $0.0$ & $39.2$ & $94.4$ & $74.4$ & $0.0$ & $74.4$ & $0.0$ & $68.8$ & $96.8$ & $71.8$ & $98.4$ & $16.8$ & $66.4$ \\
ja & $48.0$ & $0.0$ & $30.2$ & $99.2$ & $76.8$ & $0.0$ & $72.8$ & $0.0$ & $57.6$ & $97.6$ & $66.9$ & $99.2$ & $31.2$ & $35.2$ \\
ko & $52.1$ & $0.0$ & $40.0$ & $99.2$ & $76.0$ & $0.0$ & $72.8$ & $0.0$ & $58.4$ & $84.0$ & $61.6$ & $100.0$ & $41.6$ & $2.4$ \\
ms & $52.1$ & $0.8$ & $43.8$ & $92.8$ & $75.2$ & $0.8$ & $74.4$ & $0.8$ & $68.0$ & $84.0$ & $68.0$ & $5.6$ & $29.6$ & $6.4$ \\
pt & $53.3$ & $0.8$ & $32.8$ & $87.7$ & $76.0$ & $0.0$ & $76.6$ & $0.0$ & $70.4$ & $92.8$ & $68.3$ & $98.4$ & $21.1$ & $91.9$ \\
ru & $55.0$ & $0.0$ & $39.0$ & $97.6$ & $70.4$ & $0.0$ & $75.2$ & $0.0$ & $70.4$ & $97.6$ & $69.6$ & $100.0$ & $35.2$ & $48.0$ \\
sw & $46.8$ & $0.0$ & $36.0$ & $42.4$ & $72.0$ & $0.0$ & $67.2$ & $0.0$ & $18.4$ & $25.6$ & $35.2$ & $2.4$ & $14.9$ & $15.7$ \\
te & $45.2$ & $0.0$ & $28.3$ & $97.6$ & $55.2$ & $2.6$ & $60.8$ & $5.6$ & $32.0$ & $96.8$ & $45.5$ & $100.0$ & $16.5$ & $27.3$ \\
th & $51.6$ & $0.0$ & $30.6$ & $98.4$ & $68.8$ & $0.0$ & $67.2$ & $0.0$ & $56.0$ & $92.0$ & $61.8$ & $95.9$ & $36.0$ & $52.8$ \\
vi & $53.7$ & $0.0$ & $38.5$ & $97.5$ & $74.4$ & $0.0$ & $78.4$ & $0.8$ & $64.0$ & $94.4$ & $65.0$ & $100.0$ & $43.2$ & $14.4$ \\
zh & $51.2$ & $0.0$ & $36.1$ & $95.8$ & $64.0$ & $0.0$ & $70.2$ & $0.0$ & $72.8$ & $99.2$ & $70.4$ & $100.0$ & $64.0$ & $21.6$ \\
\midrule
\textbf{Avg} & $51.2$ & $5.8$ & $37.9$ & $\mathbf{93.3}$ & $\mathbf{72.3}$ & $5.8$ & $71.9$ & $6.0$ & $61.1$ & $89.2$ & $63.0$ & $87.0$ & $32.0$ & $41.5$ \\
Std & $2.8$ & $22.9$ & $6.5$ & $12.8$ & $6.2$ & $22.9$ & $6.0$ & $22.6$ & $14.9$ & $16.8$ & $9.4$ & $29.6$ & $14.9$ & $34.0$ \\
\bottomrule
\end{tabular}%
}
}
\label{tab:app-polymath-medium}
\end{table}

\begin{table}[h]
\centering
\caption{Per-language task accuracy (Acc) and L2 reasoning rate (L2\%) on \textbf{PolyMath (high)} (18 languages, 0–100). Languages are ordered alphabetically by language code. Avg / Std are over languages; the highest average in each metric is bold.}
\scriptsize{%
\setlength{\tabcolsep}{2pt}
\resizebox{\textwidth}{!}{%
\begin{tabular}{l|cccc|cccccccccc}
\toprule
Language & \multicolumn{2}{c}{\tinyayaen{}} & \multicolumn{2}{c|}{\tinyayamulti{}} & \multicolumn{2}{c}{\qwensmallnew{}} & \multicolumn{2}{c}{\begin{tabular}{@{}c@{}}\qwensmallnew{}\\{\scriptsize(user prefix LF)}\end{tabular}} & \multicolumn{2}{c}{\begin{tabular}{@{}c@{}}\qwensmallnew{}\\{\scriptsize(thinking prefix LF)}\end{tabular}} & \multicolumn{2}{c}{\mthinker{}} & \multicolumn{2}{c}{\magistralsmall{}} \\
\cmidrule(lr){2-3}\cmidrule(lr){4-5}\cmidrule(lr){6-7}\cmidrule(lr){8-9}\cmidrule(lr){10-11}\cmidrule(lr){12-13}\cmidrule(lr){14-15}
 & Acc & L2\% & Acc & L2\% & Acc & L2\% & Acc & L2\% & Acc & L2\% & Acc & L2\% & Acc & L2\% \\
\midrule
ar & $27.2$ & $0.0$ & $16.8$ & $100.0$ & $52.9$ & $0.0$ & $53.6$ & $0.0$ & $32.8$ & $95.2$ & $40.8$ & $92.0$ & $48.8$ & $0.0$ \\
bn & $26.4$ & $0.0$ & $10.4$ & $99.2$ & $44.8$ & $1.6$ & $41.6$ & $1.6$ & $21.0$ & $92.7$ & $22.6$ & $100.0$ & $24.0$ & $5.6$ \\
de & $31.2$ & $0.0$ & $18.4$ & $100.0$ & $48.0$ & $0.0$ & $56.8$ & $0.0$ & $40.0$ & $93.6$ & $40.0$ & $100.0$ & $15.2$ & $96.8$ \\
en & $24.4$ & $99.2$ & $26.6$ & $100.0$ & $34.4$ & $99.2$ & $31.9$ & $99.2$ & $30.4$ & $100.0$ & $52.9$ & $100.0$ & $40.8$ & $100.0$ \\
es & $29.0$ & $0.0$ & $13.6$ & $79.2$ & $52.0$ & $0.0$ & $50.4$ & $0.0$ & $46.3$ & $95.9$ & $45.6$ & $98.4$ & $12.8$ & $89.6$ \\
fr & $25.4$ & $0.0$ & $21.6$ & $100.0$ & $51.2$ & $0.0$ & $56.0$ & $0.0$ & $41.6$ & $93.6$ & $43.2$ & $99.2$ & $18.6$ & $73.4$ \\
id & $25.0$ & $0.0$ & $14.6$ & $97.6$ & $61.6$ & $0.0$ & $48.0$ & $0.0$ & $45.6$ & $78.4$ & $36.8$ & $100.0$ & $5.7$ & $33.1$ \\
it & $31.7$ & $0.0$ & $17.4$ & $99.2$ & $55.2$ & $0.0$ & $56.0$ & $0.0$ & $41.6$ & $93.6$ & $42.7$ & $100.0$ & $15.2$ & $56.0$ \\
ja & $27.9$ & $0.0$ & $10.7$ & $99.2$ & $52.0$ & $0.0$ & $53.6$ & $0.0$ & $31.2$ & $92.8$ & $37.6$ & $100.0$ & $12.9$ & $26.6$ \\
ko & $22.7$ & $0.0$ & $14.2$ & $99.2$ & $54.5$ & $0.0$ & $51.6$ & $0.0$ & $20.8$ & $84.0$ & $38.2$ & $100.0$ & $21.6$ & $2.4$ \\
ms & $24.2$ & $0.0$ & $18.5$ & $90.4$ & $57.6$ & $0.8$ & $54.0$ & $0.0$ & $38.4$ & $91.2$ & $39.0$ & $5.7$ & $19.2$ & $5.6$ \\
pt & $26.7$ & $0.0$ & $14.2$ & $90.8$ & $59.2$ & $0.0$ & $54.6$ & $0.0$ & $45.6$ & $88.8$ & $45.8$ & $98.3$ & $27.2$ & $93.6$ \\
ru & $31.7$ & $0.0$ & $19.2$ & $99.2$ & $56.8$ & $0.0$ & $55.2$ & $0.0$ & $46.4$ & $98.4$ & $42.0$ & $100.0$ & $22.6$ & $34.7$ \\
sw & $30.9$ & $0.0$ & $15.6$ & $70.4$ & $49.6$ & $0.0$ & $48.8$ & $0.0$ & $8.8$ & $30.4$ & $12.0$ & $5.6$ & $8.1$ & $13.0$ \\
te & $24.2$ & $0.0$ & $7.5$ & $97.6$ & $37.6$ & $0.8$ & $36.0$ & $3.2$ & $12.0$ & $95.2$ & $16.0$ & $99.0$ & $6.7$ & $23.7$ \\
th & $23.2$ & $0.0$ & $10.7$ & $98.4$ & $47.2$ & $0.0$ & $43.5$ & $0.0$ & $34.4$ & $89.6$ & $36.4$ & $100.0$ & $17.1$ & $24.4$ \\
vi & $27.2$ & $0.0$ & $20.3$ & $98.4$ & $52.8$ & $0.0$ & $58.4$ & $0.8$ & $44.0$ & $98.4$ & $40.8$ & $99.2$ & $20.0$ & $22.4$ \\
zh & $25.6$ & $0.0$ & $11.6$ & $94.2$ & $36.3$ & $0.0$ & $36.7$ & $0.0$ & $36.8$ & $94.4$ & $40.0$ & $100.0$ & $44.0$ & $52.4$ \\
\midrule
\textbf{Avg} & $26.9$ & $5.5$ & $15.6$ & $\mathbf{95.2}$ & $\mathbf{50.2}$ & $5.7$ & $49.3$ & $5.8$ & $34.3$ & $89.2$ & $37.4$ & $88.7$ & $21.1$ & $41.8$ \\
Std & $2.8$ & $22.7$ & $4.6$ & $7.9$ & $7.5$ & $22.7$ & $7.8$ & $22.7$ & $11.4$ & $15.1$ & $10.1$ & $29.4$ & $12.0$ & $34.0$ \\
\bottomrule
\end{tabular}%
}
}
\label{tab:app-polymath-high}
\end{table}

\begin{table}[h]
\centering
\caption{Per-language task accuracy (Acc) and L2 reasoning rate (L2\%) on \textbf{PolyMath (top)} (18 languages, 0–100). Languages are ordered alphabetically by language code. Avg / Std are over languages; the highest average in each metric is bold.}
\scriptsize{%
\setlength{\tabcolsep}{2pt}
\resizebox{\textwidth}{!}{%
\begin{tabular}{l|cccc|cccccccccc}
\toprule
Language & \multicolumn{2}{c}{\tinyayaen{}} & \multicolumn{2}{c|}{\tinyayamulti{}} & \multicolumn{2}{c}{\qwensmallnew{}} & \multicolumn{2}{c}{\begin{tabular}{@{}c@{}}\qwensmallnew{}\\{\scriptsize(user prefix LF)}\end{tabular}} & \multicolumn{2}{c}{\begin{tabular}{@{}c@{}}\qwensmallnew{}\\{\scriptsize(thinking prefix LF)}\end{tabular}} & \multicolumn{2}{c}{\mthinker{}} & \multicolumn{2}{c}{\magistralsmall{}} \\
\cmidrule(lr){2-3}\cmidrule(lr){4-5}\cmidrule(lr){6-7}\cmidrule(lr){8-9}\cmidrule(lr){10-11}\cmidrule(lr){12-13}\cmidrule(lr){14-15}
 & Acc & L2\% & Acc & L2\% & Acc & L2\% & Acc & L2\% & Acc & L2\% & Acc & L2\% & Acc & L2\% \\
\midrule
ar & $10.5$ & $0.0$ & $1.7$ & $100.0$ & $31.1$ & $0.0$ & $28.0$ & $0.0$ & $21.6$ & $97.6$ & $24.8$ & $90.4$ & $29.6$ & $0.0$ \\
bn & $7.2$ & $0.0$ & $0.8$ & $97.6$ & $27.1$ & $2.5$ & $28.8$ & $0.0$ & $5.6$ & $91.2$ & $16.0$ & $100.0$ & $10.4$ & $3.2$ \\
de & $5.7$ & $0.0$ & $3.2$ & $99.2$ & $26.4$ & $0.0$ & $28.8$ & $0.0$ & $24.0$ & $95.2$ & $24.4$ & $100.0$ & $8.0$ & $97.6$ \\
en & $5.0$ & $99.2$ & $2.5$ & $98.4$ & $9.8$ & $99.2$ & $8.3$ & $99.2$ & $9.6$ & $100.0$ & $30.4$ & $99.2$ & $22.4$ & $100.0$ \\
es & $5.6$ & $0.0$ & $5.7$ & $90.3$ & $27.4$ & $0.0$ & $24.8$ & $0.0$ & $20.8$ & $86.4$ & $26.4$ & $100.0$ & $6.5$ & $93.5$ \\
fr & $3.2$ & $0.0$ & $4.1$ & $100.0$ & $30.6$ & $0.0$ & $30.4$ & $0.0$ & $22.4$ & $85.6$ & $25.2$ & $98.4$ & $13.6$ & $77.6$ \\
id & $6.5$ & $0.0$ & $5.6$ & $92.0$ & $28.0$ & $0.0$ & $30.4$ & $0.0$ & $26.4$ & $80.8$ & $22.1$ & $98.4$ & $7.3$ & $38.7$ \\
it & $2.5$ & $0.0$ & $4.1$ & $99.2$ & $25.0$ & $0.8$ & $29.6$ & $0.0$ & $15.2$ & $96.8$ & $27.4$ & $99.2$ & $8.8$ & $66.4$ \\
ja & $8.3$ & $0.0$ & $1.6$ & $98.4$ & $35.5$ & $0.0$ & $29.6$ & $0.0$ & $11.2$ & $95.2$ & $26.8$ & $97.6$ & $5.6$ & $22.4$ \\
ko & $4.2$ & $0.0$ & $2.5$ & $96.8$ & $24.8$ & $0.0$ & $30.4$ & $0.0$ & $8.8$ & $84.0$ & $22.3$ & $100.0$ & $8.8$ & $3.2$ \\
ms & $5.6$ & $0.0$ & $5.8$ & $94.4$ & $32.0$ & $0.0$ & $31.2$ & $0.0$ & $16.0$ & $88.8$ & $27.1$ & $10.7$ & $9.0$ & $11.5$ \\
pt & $8.1$ & $0.0$ & $0.8$ & $89.6$ & $27.2$ & $0.0$ & $29.4$ & $0.0$ & $19.2$ & $91.2$ & $25.6$ & $100.0$ & $9.8$ & $92.6$ \\
ru & $4.9$ & $0.0$ & $1.6$ & $96.8$ & $22.4$ & $0.0$ & $33.9$ & $0.0$ & $23.2$ & $91.2$ & $32.2$ & $100.0$ & $8.8$ & $49.6$ \\
sw & $5.7$ & $0.0$ & $4.1$ & $72.0$ & $24.0$ & $0.0$ & $22.4$ & $0.0$ & $0.8$ & $28.0$ & $8.8$ & $8.0$ & $9.7$ & $14.5$ \\
te & $6.6$ & $0.8$ & $0.0$ & $99.2$ & $20.0$ & $0.0$ & $20.0$ & $0.8$ & $5.6$ & $90.4$ & $10.4$ & $100.0$ & $4.0$ & $23.4$ \\
th & $7.4$ & $0.0$ & $0.8$ & $97.5$ & $17.6$ & $0.0$ & $24.0$ & $0.0$ & $13.6$ & $81.6$ & $21.6$ & $98.4$ & $3.2$ & $20.8$ \\
vi & $7.2$ & $0.0$ & $0.8$ & $100.0$ & $32.8$ & $0.0$ & $29.6$ & $0.0$ & $20.8$ & $93.6$ & $25.0$ & $100.0$ & $8.8$ & $20.8$ \\
zh & $7.3$ & $0.0$ & $1.6$ & $97.6$ & $8.8$ & $0.0$ & $9.8$ & $0.0$ & $15.2$ & $98.4$ & $27.4$ & $100.0$ & $20.0$ & $49.2$ \\
\midrule
\textbf{Avg} & $6.2$ & $5.6$ & $2.6$ & $\mathbf{95.5}$ & $25.0$ & $5.7$ & $\mathbf{26.1}$ & $5.6$ & $15.6$ & $87.6$ & $23.6$ & $88.9$ & $10.8$ & $43.6$ \\
Std & $1.9$ & $22.7$ & $1.8$ & $6.5$ & $7.1$ & $22.7$ & $6.9$ & $22.7$ & $7.1$ & $15.5$ & $6.0$ & $28.2$ & $6.5$ & $34.7$ \\
\bottomrule
\end{tabular}%
}
}
\label{tab:app-polymath-top}
\end{table}

\begin{table}[h]
\centering
\caption{Per-language task accuracy (Acc) and L2 reasoning rate (L2\%) on \textbf{\mist{}} (25 languages, 0–100). MIST judge score (1–7) is linearly rescaled to the same Acc range. Languages are ordered alphabetically by language code. Avg / Std are over languages; the highest average in each metric is bold.}
\scriptsize{%
\setlength{\tabcolsep}{2pt}
\resizebox{\textwidth}{!}{%
\begin{tabular}{l|cccc|cccccccccc}
\toprule
Language & \multicolumn{2}{c}{\tinyayaen{}} & \multicolumn{2}{c|}{\tinyayamulti{}} & \multicolumn{2}{c}{\qwensmallnew{}} & \multicolumn{2}{c}{\begin{tabular}{@{}c@{}}\qwensmallnew{}\\{\scriptsize(user prefix LF)}\end{tabular}} & \multicolumn{2}{c}{\begin{tabular}{@{}c@{}}\qwensmallnew{}\\{\scriptsize(thinking prefix LF)}\end{tabular}} & \multicolumn{2}{c}{\mthinker{}} & \multicolumn{2}{c}{\magistralsmall{}} \\
\cmidrule(lr){2-3}\cmidrule(lr){4-5}\cmidrule(lr){6-7}\cmidrule(lr){8-9}\cmidrule(lr){10-11}\cmidrule(lr){12-13}\cmidrule(lr){14-15}
 & Acc & L2\% & Acc & L2\% & Acc & L2\% & Acc & L2\% & Acc & L2\% & Acc & L2\% & Acc & L2\% \\
\midrule
ar & $83.7$ & $5.0$ & $85.2$ & $99.0$ & $78.2$ & $12.1$ & $77.7$ & $24.0$ & $70.3$ & $99.0$ & $17.8$ & $98.0$ & $60.5$ & $29.3$ \\
bn & $88.0$ & $15.0$ & $88.8$ & $97.0$ & $78.2$ & $38.1$ & $79.7$ & $48.0$ & $35.8$ & $100.0$ & $36.5$ & $99.0$ & $68.8$ & $18.4$ \\
cs & $84.7$ & $6.0$ & $87.5$ & $100.0$ & $71.2$ & $12.2$ & $75.8$ & $25.0$ & $49.5$ & $96.0$ & $25.0$ & $100.0$ & $86.0$ & $96.9$ \\
de & $87.0$ & $15.0$ & $82.8$ & $100.0$ & $84.3$ & $13.4$ & $84.2$ & $28.0$ & $79.8$ & $77.0$ & $36.2$ & $100.0$ & $98.3$ & $99.0$ \\
el & $88.8$ & $6.0$ & $87.3$ & $100.0$ & $75.5$ & $22.2$ & $79.8$ & $39.0$ & $66.0$ & $99.0$ & $7.5$ & $99.0$ & $88.0$ & $97.0$ \\
en & $95.3$ & $100.0$ & $95.0$ & $100.0$ & $90.8$ & $98.0$ & $91.3$ & $99.0$ & $57.3$ & $100.0$ & $90.7$ & $99.0$ & $97.8$ & $100.0$ \\
et & $88.3$ & $7.0$ & $85.2$ & $98.0$ & $53.0$ & $17.9$ & $58.3$ & $30.0$ & $35.8$ & $96.0$ & $7.7$ & $100.0$ & $72.7$ & $41.0$ \\
fa & $91.7$ & $13.0$ & $88.2$ & $100.0$ & $89.5$ & $13.0$ & $92.7$ & $37.0$ & $77.2$ & $100.0$ & $37.0$ & $100.0$ & $55.3$ & $16.0$ \\
hi & $93.8$ & $19.0$ & $94.2$ & $100.0$ & $86.2$ & $15.6$ & $83.0$ & $33.0$ & $55.5$ & $99.0$ & $42.8$ & $100.0$ & $67.8$ & $14.0$ \\
hr & $83.0$ & $1.0$ & $86.7$ & $23.0$ & $75.5$ & $1.0$ & $79.5$ & $4.0$ & $65.7$ & $57.0$ & $31.0$ & $39.4$ & $82.7$ & $27.8$ \\
id & $88.3$ & $14.0$ & $90.5$ & $95.0$ & $89.2$ & $14.3$ & $86.0$ & $43.0$ & $84.3$ & $96.0$ & $52.2$ & $100.0$ & $89.8$ & $75.0$ \\
it & $87.3$ & $10.0$ & $87.8$ & $100.0$ & $87.2$ & $16.3$ & $86.7$ & $31.0$ & $83.5$ & $97.0$ & $41.7$ & $100.0$ & $99.0$ & $100.0$ \\
ja & $73.5$ & $2.0$ & $74.2$ & $99.0$ & $87.5$ & $5.1$ & $86.7$ & $15.3$ & $75.3$ & $99.0$ & $28.8$ & $99.0$ & $77.8$ & $39.0$ \\
ko & $74.3$ & $4.0$ & $79.8$ & $99.0$ & $89.5$ & $11.3$ & $89.7$ & $18.4$ & $79.7$ & $100.0$ & $19.3$ & $100.0$ & $70.2$ & $22.1$ \\
lt & $86.8$ & $6.0$ & $86.8$ & $99.0$ & $69.0$ & $19.2$ & $67.0$ & $27.0$ & $59.5$ & $97.0$ & $7.5$ & $100.0$ & $76.8$ & $60.0$ \\
mr & $91.0$ & $8.0$ & $88.8$ & $97.0$ & $54.8$ & $11.1$ & $69.5$ & $25.3$ & $36.2$ & $54.0$ & $23.0$ & $98.0$ & $63.8$ & $24.0$ \\
ro & $88.8$ & $17.0$ & $87.3$ & $93.0$ & $82.8$ & $18.2$ & $84.7$ & $22.0$ & $78.0$ & $97.0$ & $32.5$ & $100.0$ & $95.2$ & $95.0$ \\
ru & $84.5$ & $9.0$ & $89.8$ & $100.0$ & $90.0$ & $18.0$ & $90.7$ & $29.0$ & $84.3$ & $99.0$ & $33.3$ & $100.0$ & $94.5$ & $100.0$ \\
sr & $77.7$ & $5.0$ & $79.8$ & $94.0$ & $71.3$ & $6.1$ & $76.5$ & $21.9$ & $56.0$ & $95.0$ & $23.5$ & $86.0$ & $88.0$ & $73.0$ \\
sv & $86.7$ & $8.0$ & $87.0$ & $97.0$ & $82.2$ & $12.1$ & $78.5$ & $36.7$ & $69.0$ & $98.0$ & $34.3$ & $99.0$ & $93.3$ & $95.0$ \\
th & $73.5$ & $21.0$ & $73.2$ & $99.0$ & $83.5$ & $30.3$ & $85.8$ & $37.5$ & $76.2$ & $99.0$ & $14.3$ & $100.0$ & $71.3$ & $46.0$ \\
tr & $84.2$ & $12.0$ & $85.2$ & $100.0$ & $71.7$ & $16.0$ & $74.5$ & $31.2$ & $64.2$ & $99.0$ & $25.0$ & $100.0$ & $70.8$ & $30.0$ \\
uk & $88.3$ & $7.1$ & $89.0$ & $99.0$ & $84.3$ & $20.4$ & $89.3$ & $38.1$ & $74.7$ & $100.0$ & $33.3$ & $99.0$ & $90.5$ & $91.9$ \\
vi & $89.7$ & $42.0$ & $89.0$ & $100.0$ & $93.0$ & $16.2$ & $92.3$ & $53.1$ & $81.8$ & $98.0$ & $22.3$ & $100.0$ & $83.3$ & $63.0$ \\
zh & $83.3$ & $1.0$ & $85.3$ & $98.0$ & $90.0$ & $4.0$ & $88.3$ & $8.0$ & $85.3$ & $88.9$ & $87.3$ & $100.0$ & $95.0$ & $97.0$ \\
\midrule
\textbf{Avg} & $85.7$ & $14.1$ & $\mathbf{86.2}$ & $95.4$ & $80.3$ & $18.5$ & $81.9$ & $32.2$ & $67.2$ & $93.6$ & $32.4$ & $\mathbf{96.6}$ & $81.5$ & $62.0$ \\
Std & $5.7$ & $19.4$ & $5.0$ & $14.9$ & $10.4$ & $17.9$ & $8.3$ & $17.6$ & $15.3$ & $12.2$ & $20.0$ & $12.0$ & $12.7$ & $32.6$ \\
\bottomrule
\end{tabular}%
}
}
\label{tab:app-mist_oeg}
\end{table}

\begin{table}[h]
\centering
\caption{Per-language task accuracy (Acc) and L2 reasoning rate (L2\%) on \textbf{\mif{}} (29 languages, 0–100). Languages are ordered alphabetically by language code. Avg / Std are over languages; the highest average in each metric is bold.}
\scriptsize{%
\setlength{\tabcolsep}{2pt}
\resizebox{\textwidth}{!}{%
\begin{tabular}{l|cccc|cccccccccc}
\toprule
Language & \multicolumn{2}{c}{\tinyayaen{}} & \multicolumn{2}{c|}{\tinyayamulti{}} & \multicolumn{2}{c}{\qwensmallnew{}} & \multicolumn{2}{c}{\begin{tabular}{@{}c@{}}\qwensmallnew{}\\{\scriptsize(user prefix LF)}\end{tabular}} & \multicolumn{2}{c}{\begin{tabular}{@{}c@{}}\qwensmallnew{}\\{\scriptsize(thinking prefix LF)}\end{tabular}} & \multicolumn{2}{c}{\mthinker{}} & \multicolumn{2}{c}{\magistralsmall{}} \\
\cmidrule(lr){2-3}\cmidrule(lr){4-5}\cmidrule(lr){6-7}\cmidrule(lr){8-9}\cmidrule(lr){10-11}\cmidrule(lr){12-13}\cmidrule(lr){14-15}
 & Acc & L2\% & Acc & L2\% & Acc & L2\% & Acc & L2\% & Acc & L2\% & Acc & L2\% & Acc & L2\% \\
\midrule
ar & $51.0$ & $6.1$ & $49.8$ & $99.1$ & $30.2$ & $33.1$ & $25.7$ & $37.1$ & $39.2$ & $91.7$ & $27.6$ & $98.3$ & $36.0$ & $20.3$ \\
bn & $48.6$ & $10.2$ & $51.1$ & $98.3$ & $15.6$ & $40.5$ & $14.2$ & $40.3$ & $24.5$ & $96.2$ & $22.6$ & $99.4$ & $29.4$ & $13.7$ \\
cs & $52.9$ & $5.9$ & $51.6$ & $97.2$ & $22.2$ & $34.4$ & $19.8$ & $35.2$ & $27.0$ & $86.0$ & $28.6$ & $99.1$ & $58.1$ & $74.4$ \\
de & $55.3$ & $7.6$ & $52.0$ & $99.2$ & $24.7$ & $33.1$ & $27.4$ & $35.1$ & $41.4$ & $64.1$ & $34.0$ & $99.4$ & $68.1$ & $92.5$ \\
el & $53.2$ & $4.8$ & $47.8$ & $99.4$ & $26.1$ & $38.1$ & $20.3$ & $42.3$ & $30.3$ & $87.1$ & $19.8$ & $99.6$ & $63.9$ & $71.6$ \\
en & $71.2$ & $97.6$ & $72.7$ & $98.5$ & $37.0$ & $95.7$ & $34.6$ & $97.8$ & $34.6$ & $95.6$ & $57.1$ & $97.2$ & $78.9$ & $96.1$ \\
es & $62.9$ & $22.2$ & $57.7$ & $95.0$ & $35.4$ & $34.5$ & $32.4$ & $37.1$ & $55.6$ & $84.1$ & $37.7$ & $99.2$ & $69.3$ & $84.6$ \\
fr & $59.0$ & $12.0$ & $56.8$ & $99.1$ & $32.0$ & $32.9$ & $29.1$ & $33.0$ & $52.7$ & $78.9$ & $37.0$ & $99.1$ & $70.4$ & $86.3$ \\
he & $50.1$ & $2.0$ & $54.1$ & $98.1$ & $16.6$ & $35.0$ & $15.3$ & $35.1$ & $17.4$ & $90.4$ & $23.1$ & $95.0$ & $32.7$ & $29.0$ \\
hu & $56.6$ & $5.6$ & $52.7$ & $97.4$ & $18.0$ & $33.1$ & $16.4$ & $38.1$ & $24.2$ & $92.6$ & $23.2$ & $99.8$ & $54.1$ & $53.0$ \\
id & $60.3$ & $8.5$ & $60.0$ & $98.5$ & $31.8$ & $31.5$ & $28.8$ & $36.1$ & $53.2$ & $74.3$ & $30.8$ & $99.4$ & $66.5$ & $59.0$ \\
it & $57.5$ & $8.7$ & $56.7$ & $99.3$ & $30.5$ & $32.7$ & $28.4$ & $37.2$ & $48.4$ & $77.6$ & $35.9$ & $99.6$ & $69.0$ & $88.5$ \\
ja & $46.5$ & $1.1$ & $44.4$ & $98.3$ & $19.0$ & $21.9$ & $19.4$ & $28.6$ & $31.1$ & $87.8$ & $32.6$ & $98.9$ & $36.0$ & $13.6$ \\
ko & $45.7$ & $0.6$ & $38.0$ & $98.1$ & $23.3$ & $27.2$ & $21.6$ & $29.2$ & $38.5$ & $86.7$ & $25.1$ & $99.6$ & $29.6$ & $12.8$ \\
ms & $58.0$ & $15.6$ & $56.7$ & $94.6$ & $32.6$ & $34.3$ & $27.2$ & $36.6$ & $34.8$ & $75.0$ & $31.2$ & $23.3$ & $67.3$ & $20.0$ \\
ne & $53.3$ & $3.3$ & $50.2$ & $97.9$ & $10.6$ & $20.4$ & $7.0$ & $22.9$ & $11.1$ & $80.3$ & $23.3$ & $97.4$ & $33.5$ & $8.9$ \\
nl & $63.6$ & $14.1$ & $59.3$ & $97.2$ & $27.9$ & $34.9$ & $25.2$ & $35.7$ & $34.8$ & $80.8$ & $31.2$ & $99.2$ & $69.0$ & $81.7$ \\
pl & $43.4$ & $9.3$ & $40.2$ & $99.2$ & $21.0$ & $34.8$ & $19.9$ & $35.6$ & $25.1$ & $86.3$ & $29.3$ & $99.6$ & $49.7$ & $69.7$ \\
pt & $55.1$ & $26.5$ & $53.2$ & $97.4$ & $30.3$ & $33.9$ & $29.8$ & $32.4$ & $49.5$ & $83.0$ & $34.8$ & $99.8$ & $69.9$ & $81.9$ \\
ro & $51.6$ & $11.5$ & $47.9$ & $98.1$ & $26.2$ & $33.3$ & $19.4$ & $34.0$ & $30.9$ & $82.1$ & $25.1$ & $100.0$ & $57.8$ & $69.0$ \\
ru & $48.2$ & $4.3$ & $46.3$ & $99.1$ & $30.4$ & $35.2$ & $24.4$ & $38.0$ & $46.4$ & $84.3$ & $38.7$ & $98.0$ & $57.1$ & $78.9$ \\
sw & $53.3$ & $0.4$ & $52.4$ & $85.5$ & $9.9$ & $12.1$ & $8.0$ & $12.8$ & $1.9$ & $65.2$ & $15.2$ & $87.2$ & $35.5$ & $3.7$ \\
th & $49.9$ & $2.2$ & $51.4$ & $98.1$ & $13.2$ & $39.0$ & $11.1$ & $34.6$ & $25.5$ & $87.1$ & $23.8$ & $98.3$ & $34.2$ & $26.6$ \\
tr & $54.6$ & $6.1$ & $51.1$ & $99.4$ & $26.6$ & $35.6$ & $21.4$ & $39.2$ & $39.4$ & $90.4$ & $25.9$ & $99.3$ & $73.0$ & $19.6$ \\
uk & $46.4$ & $4.1$ & $47.2$ & $98.9$ & $23.1$ & $37.4$ & $20.2$ & $41.0$ & $37.1$ & $87.8$ & $29.5$ & $99.4$ & $47.5$ & $64.5$ \\
ur & $52.9$ & $12.4$ & $51.0$ & $98.7$ & $24.5$ & $33.4$ & $19.4$ & $33.9$ & $30.3$ & $89.8$ & $19.9$ & $98.9$ & $28.1$ & $13.9$ \\
vi & $52.3$ & $14.0$ & $51.5$ & $99.4$ & $31.2$ & $39.7$ & $29.8$ & $43.8$ & $43.6$ & $81.7$ & $27.9$ & $100.0$ & $45.1$ & $47.7$ \\
yo & $47.3$ & $1.3$ & $44.0$ & $71.6$ & $2.8$ & $15.6$ & $2.9$ & $17.8$ & $0.7$ & $55.7$ & $11.5$ & $48.6$ & $17.3$ & $9.5$ \\
zh & $49.5$ & $0.7$ & $44.4$ & $96.9$ & $31.4$ & $11.3$ & $29.1$ & $10.9$ & $51.8$ & $76.2$ & $45.3$ & $98.5$ & $58.2$ & $75.2$ \\
\midrule
\textbf{Avg} & $\mathbf{53.5}$ & $11.0$ & $51.5$ & $\mathbf{96.8}$ & $24.3$ & $33.6$ & $21.7$ & $35.6$ & $33.8$ & $82.7$ & $29.2$ & $94.2$ & $51.9$ & $50.6$ \\
Std & $6.0$ & $17.5$ & $6.6$ & $5.4$ & $8.2$ & $14.0$ & $7.7$ & $14.1$ & $14.0$ & $9.2$ & $8.9$ & $16.4$ & $17.0$ & $31.1$ \\
\bottomrule
\end{tabular}%
}
}
\label{tab:app-marcobenchmif}
\end{table}

\begin{table}[h]
\centering
\caption{Per-language task accuracy (Acc) and L2 reasoning rate (L2\%) on \textbf{GlobalPIQA} (59 languages, 0–100). Languages are ordered alphabetically by language code. Avg / Std are over languages; the highest average in each metric is bold.}
\scriptsize{%
\setlength{\tabcolsep}{2pt}
\resizebox{\textwidth}{!}{%
\begin{tabular}{l|cccc|cccccccccc}
\toprule
Language & \multicolumn{2}{c}{\tinyayaen{}} & \multicolumn{2}{c|}{\tinyayamulti{}} & \multicolumn{2}{c}{\qwensmallnew{}} & \multicolumn{2}{c}{\begin{tabular}{@{}c@{}}\qwensmallnew{}\\{\scriptsize(user prefix LF)}\end{tabular}} & \multicolumn{2}{c}{\begin{tabular}{@{}c@{}}\qwensmallnew{}\\{\scriptsize(thinking prefix LF)}\end{tabular}} & \multicolumn{2}{c}{\mthinker{}} & \multicolumn{2}{c}{\magistralsmall{}} \\
\cmidrule(lr){2-3}\cmidrule(lr){4-5}\cmidrule(lr){6-7}\cmidrule(lr){8-9}\cmidrule(lr){10-11}\cmidrule(lr){12-13}\cmidrule(lr){14-15}
 & Acc & L2\% & Acc & L2\% & Acc & L2\% & Acc & L2\% & Acc & L2\% & Acc & L2\% & Acc & L2\% \\
\midrule
amh\_ethi & $74.0$ & $17.3$ & $63.6$ & $100.0$ & $70.0$ & $0.0$ & $69.0$ & $0.0$ & $65.0$ & $98.0$ & $46.0$ & $96.0$ & $61.0$ & $0.0$ \\
arb\_arab & $63.0$ & $14.0$ & $68.0$ & $100.0$ & $76.0$ & $1.0$ & $73.0$ & $6.0$ & $66.0$ & $99.0$ & $57.0$ & $100.0$ & $69.0$ & $2.0$ \\
ben\_beng & $84.0$ & $60.0$ & $79.0$ & $100.0$ & $75.0$ & $1.0$ & $70.0$ & $5.0$ & $77.0$ & $100.0$ & $49.0$ & $100.0$ & $81.0$ & $17.0$ \\
bul\_cyrl & $78.0$ & $1.0$ & $81.0$ & $100.0$ & $91.0$ & $6.0$ & $89.0$ & $6.0$ & $89.0$ & $100.0$ & $73.0$ & $100.0$ & $94.0$ & $99.0$ \\
cat\_latn & $65.0$ & $33.0$ & $65.0$ & $100.0$ & $75.0$ & $2.0$ & $81.0$ & $1.0$ & $78.0$ & $100.0$ & $57.0$ & $100.0$ & $68.0$ & $79.4$ \\
ces\_latn & $65.0$ & $9.0$ & $68.0$ & $100.0$ & $76.5$ & $3.1$ & $75.0$ & $5.0$ & $73.0$ & $98.0$ & $52.0$ & $100.0$ & $80.0$ & $100.0$ \\
cmn\_hans & $70.0$ & $0.0$ & $58.0$ & $100.0$ & $83.8$ & $0.0$ & $80.0$ & $0.0$ & $83.0$ & $100.0$ & $75.0$ & $100.0$ & $67.0$ & $95.0$ \\
cmn\_hant & $61.0$ & $1.0$ & $54.5$ & $96.9$ & $74.0$ & $0.0$ & $77.0$ & $0.0$ & $74.0$ & $100.0$ & $62.0$ & $100.0$ & $70.0$ & $93.0$ \\
deu\_latn & $70.0$ & $48.0$ & $66.0$ & $100.0$ & $82.8$ & $1.0$ & $80.0$ & $3.0$ & $78.0$ & $99.0$ & $63.0$ & $100.0$ & $84.0$ & $98.0$ \\
ekk\_latn & $56.0$ & $26.0$ & $51.0$ & $100.0$ & $59.6$ & $0.0$ & $70.7$ & $1.0$ & $62.0$ & $100.0$ & $51.0$ & $99.0$ & $75.0$ & $3.0$ \\
ell\_grek & $61.0$ & $55.0$ & $55.0$ & $100.0$ & $69.7$ & $14.1$ & $64.0$ & $27.0$ & $66.0$ & $100.0$ & $52.1$ & $100.0$ & $70.7$ & $91.9$ \\
eng\_latn & $75.0$ & $100.0$ & $70.0$ & $100.0$ & $88.0$ & $100.0$ & $90.0$ & $100.0$ & $90.0$ & $100.0$ & $71.0$ & $100.0$ & $88.0$ & $99.0$ \\
fin\_latn & $71.0$ & $50.0$ & $63.0$ & $100.0$ & $86.0$ & $4.0$ & $83.0$ & $1.0$ & $86.0$ & $2.0$ & $54.0$ & $99.0$ & $90.0$ & $0.0$ \\
fra\_latn\_cana & $88.0$ & $14.0$ & $86.0$ & $100.0$ & $92.0$ & $4.0$ & $95.0$ & $17.0$ & $95.0$ & $99.0$ & $88.0$ & $100.0$ & $76.0$ & $94.0$ \\
fra\_latn\_fran & $60.0$ & $40.0$ & $65.0$ & $100.0$ & $75.8$ & $5.1$ & $82.0$ & $21.0$ & $80.0$ & $100.0$ & $67.0$ & $100.0$ & $70.0$ & $96.0$ \\
glg\_latn & $67.0$ & $10.0$ & $64.0$ & $98.9$ & $82.7$ & $0.0$ & $85.0$ & $1.0$ & $82.0$ & $99.0$ & $70.0$ & $1.0$ & $67.0$ & $56.0$ \\
guj\_gujr & $83.0$ & $5.0$ & $76.0$ & $100.0$ & $80.0$ & $22.0$ & $80.0$ & $19.0$ & $83.0$ & $100.0$ & $56.0$ & $100.0$ & $86.0$ & $0.0$ \\
hau\_latn & $69.0$ & $33.0$ & $77.5$ & $100.0$ & $59.0$ & $7.0$ & $62.0$ & $7.0$ & $65.0$ & $96.0$ & $48.0$ & $52.0$ & $60.0$ & $0.0$ \\
heb\_hebr & $55.0$ & $0.0$ & $59.0$ & $100.0$ & $68.7$ & $13.1$ & $75.0$ & $21.0$ & $68.0$ & $100.0$ & $62.0$ & $100.0$ & $82.0$ & $6.0$ \\
hin\_deva & $79.0$ & $24.0$ & $82.0$ & $100.0$ & $83.0$ & $2.0$ & $82.0$ & $0.0$ & $66.0$ & $100.0$ & $66.0$ & $100.0$ & $93.0$ & $0.0$ \\
hrv\_latn & $83.0$ & $10.2$ & $82.0$ & $97.0$ & $90.0$ & $0.0$ & $88.0$ & $1.0$ & $91.0$ & $95.0$ & $58.0$ & $85.0$ & $85.0$ & $82.0$ \\
hun\_latn & $82.0$ & $1.0$ & $77.0$ & $100.0$ & $92.0$ & $6.0$ & $96.0$ & $4.0$ & $90.0$ & $100.0$ & $52.0$ & $100.0$ & $90.0$ & $10.0$ \\
ibo\_latn & $62.0$ & $88.0$ & $65.0$ & $99.0$ & $64.7$ & $31.3$ & $59.0$ & $33.0$ & $52.0$ & $99.0$ & $55.6$ & $49.5$ & $53.0$ & $7.0$ \\
ind\_latn & $85.0$ & $9.0$ & $85.0$ & $100.0$ & $93.0$ & $15.0$ & $91.0$ & $36.0$ & $95.0$ & $100.0$ & $69.0$ & $100.0$ & $83.0$ & $37.0$ \\
ita\_latn & $64.0$ & $43.0$ & $66.0$ & $100.0$ & $87.9$ & $9.1$ & $82.0$ & $26.0$ & $82.0$ & $97.0$ & $62.0$ & $100.0$ & $82.8$ & $97.0$ \\
jav\_latn & $60.0$ & $17.5$ & $61.0$ & $100.0$ & $72.0$ & $7.0$ & $70.0$ & $4.0$ & $61.0$ & $97.0$ & $51.0$ & $0.0$ & $74.0$ & $0.0$ \\
jpn\_jpan & $75.0$ & $0.0$ & $76.0$ & $100.0$ & $91.9$ & $0.0$ & $89.0$ & $0.0$ & $92.0$ & $100.0$ & $66.0$ & $100.0$ & $91.0$ & $5.0$ \\
kor\_hang & $62.0$ & $0.0$ & $67.0$ & $100.0$ & $73.0$ & $0.0$ & $64.0$ & $2.0$ & $71.0$ & $98.0$ & $56.6$ & $100.0$ & $84.0$ & $2.0$ \\
lit\_latn & $76.0$ & $36.0$ & $71.0$ & $100.0$ & $86.0$ & $4.0$ & $76.0$ & $6.0$ & $82.0$ & $99.0$ & $48.0$ & $99.0$ & $89.0$ & $93.0$ \\
mar\_deva & $82.0$ & $68.0$ & $75.8$ & $100.0$ & $72.0$ & $2.0$ & $71.0$ & $1.0$ & $73.0$ & $66.0$ & $55.6$ & $100.0$ & $87.0$ & $0.0$ \\
nld\_latn & $76.0$ & $57.0$ & $73.0$ & $100.0$ & $78.0$ & $1.0$ & $81.0$ & $12.0$ & $75.0$ & $100.0$ & $66.0$ & $100.0$ & $61.6$ & $96.0$ \\
nob\_latn & $65.0$ & $37.0$ & $59.0$ & $100.0$ & $77.8$ & $4.0$ & $74.0$ & $23.0$ & $76.0$ & $100.0$ & $52.0$ & $100.0$ & $71.0$ & $90.0$ \\
pes\_arab & $68.0$ & $22.0$ & $77.0$ & $100.0$ & $81.0$ & $1.0$ & $83.0$ & $6.0$ & $86.0$ & $100.0$ & $57.0$ & $100.0$ & $80.0$ & $8.0$ \\
pol\_latn & $68.0$ & $68.7$ & $61.0$ & $100.0$ & $74.0$ & $14.0$ & $78.0$ & $29.0$ & $76.0$ & $100.0$ & $56.0$ & $100.0$ & $86.0$ & $100.0$ \\
por\_latn\_braz & $79.0$ & $95.0$ & $84.0$ & $100.0$ & $88.0$ & $7.0$ & $87.0$ & $23.0$ & $94.0$ & $99.0$ & $74.0$ & $100.0$ & $78.0$ & $95.0$ \\
por\_latn\_port & $70.0$ & $84.0$ & $72.0$ & $100.0$ & $80.0$ & $4.0$ & $77.0$ & $7.0$ & $80.0$ & $100.0$ & $54.0$ & $100.0$ & $80.0$ & $94.0$ \\
ron\_latn & $89.0$ & $0.0$ & $90.0$ & $100.0$ & $98.0$ & $20.0$ & $97.0$ & $38.0$ & $98.0$ & $100.0$ & $68.0$ & $100.0$ & $98.0$ & $100.0$ \\
rus\_cyrl & $74.0$ & $32.6$ & $72.0$ & $100.0$ & $86.0$ & $15.0$ & $90.0$ & $30.0$ & $88.0$ & $100.0$ & $58.6$ & $100.0$ & $76.8$ & $98.0$ \\
slk\_latn & $72.0$ & $0.0$ & $72.0$ & $100.0$ & $88.9$ & $3.0$ & $88.0$ & $2.0$ & $87.0$ & $100.0$ & $56.0$ & $100.0$ & $73.0$ & $89.0$ \\
slk\_latn\_sari & $68.0$ & $1.0$ & $50.0$ & $98.0$ & $63.6$ & $1.0$ & $62.0$ & $0.0$ & $58.0$ & $86.0$ & $46.0$ & $98.0$ & $65.0$ & $24.0$ \\
slv\_latn & $66.0$ & $0.0$ & $70.0$ & $100.0$ & $79.0$ & $1.0$ & $73.0$ & $0.0$ & $82.0$ & $98.0$ & $54.0$ & $99.0$ & $64.0$ & $45.0$ \\
slv\_latn\_cerk & $51.0$ & $0.0$ & $54.0$ & $99.0$ & $48.0$ & $3.0$ & $62.0$ & $0.0$ & $44.0$ & $61.0$ & $54.0$ & $97.0$ & $56.6$ & $60.6$ \\
spa\_latn\_mexi & $91.0$ & $93.0$ & $85.0$ & $100.0$ & $92.0$ & $8.0$ & $95.0$ & $19.0$ & $93.0$ & $98.0$ & $69.0$ & $100.0$ & $72.0$ & $97.0$ \\
spa\_latn\_peru & $95.0$ & $85.0$ & $93.0$ & $100.0$ & $95.0$ & $7.0$ & $95.0$ & $15.0$ & $97.0$ & $100.0$ & $88.0$ & $100.0$ & $74.0$ & $99.0$ \\
spa\_latn\_spai & $78.0$ & $72.0$ & $76.0$ & $100.0$ & $84.0$ & $3.0$ & $85.0$ & $10.0$ & $86.0$ & $100.0$ & $64.0$ & $100.0$ & $67.0$ & $100.0$ \\
srp\_cyrl & $63.0$ & $3.0$ & $70.0$ & $100.0$ & $88.0$ & $1.0$ & $84.0$ & $8.0$ & $85.0$ & $98.0$ & $54.0$ & $82.0$ & $87.0$ & $97.0$ \\
swe\_latn & $75.0$ & $20.0$ & $68.0$ & $100.0$ & $79.0$ & $11.0$ & $84.0$ & $11.0$ & $83.0$ & $99.0$ & $56.0$ & $100.0$ & $82.0$ & $96.0$ \\
swh\_latn & $85.0$ & $0.0$ & $79.0$ & $69.0$ & $75.0$ & $2.0$ & $75.0$ & $0.0$ & $65.0$ & $79.0$ & $49.0$ & $96.0$ & $78.0$ & $16.0$ \\
tam\_taml & $73.0$ & $20.0$ & $71.0$ & $100.0$ & $63.0$ & $58.0$ & $63.0$ & $51.0$ & $61.0$ & $100.0$ & $56.0$ & $100.0$ & $77.0$ & $0.0$ \\
tel\_telu & $83.0$ & $35.0$ & $75.0$ & $100.0$ & $66.7$ & $38.4$ & $59.0$ & $36.0$ & $55.0$ & $100.0$ & $55.0$ & $100.0$ & $79.0$ & $24.0$ \\
tgl\_latn & $77.0$ & $17.0$ & $71.0$ & $100.0$ & $77.0$ & $1.0$ & $77.0$ & $1.0$ & $67.0$ & $95.0$ & $53.0$ & $100.0$ & $83.0$ & $2.0$ \\
tha\_thai & $72.0$ & $19.0$ & $69.0$ & $100.0$ & $74.5$ & $36.7$ & $72.0$ & $32.0$ & $78.0$ & $98.0$ & $52.0$ & $100.0$ & $77.0$ & $2.0$ \\
tur\_latn & $67.0$ & $10.0$ & $73.0$ & $100.0$ & $86.0$ & $0.0$ & $86.0$ & $2.0$ & $84.0$ & $100.0$ & $58.0$ & $100.0$ & $85.0$ & $47.0$ \\
ukr\_cyrl & $74.0$ & $22.0$ & $77.0$ & $100.0$ & $85.0$ & $25.0$ & $83.0$ & $31.0$ & $83.0$ & $100.0$ & $66.7$ & $100.0$ & $90.0$ & $90.0$ \\
urd\_arab & $84.0$ & $70.0$ & $81.0$ & $100.0$ & $85.0$ & $9.0$ & $90.0$ & $4.0$ & $91.0$ & $100.0$ & $57.0$ & $100.0$ & $91.0$ & $1.0$ \\
vie\_latn & $66.0$ & $50.5$ & $78.0$ & $100.0$ & $83.0$ & $9.0$ & $85.0$ & $29.0$ & $87.0$ & $100.0$ & $57.0$ & $100.0$ & $77.0$ & $4.0$ \\
yor\_latn & $66.0$ & $0.0$ & $56.0$ & $43.0$ & $52.0$ & $4.0$ & $44.0$ & $0.0$ & $37.0$ & $47.0$ & $51.0$ & $72.0$ & $51.0$ & $0.0$ \\
zsm\_latn & $74.0$ & $4.0$ & $71.0$ & $100.0$ & $78.0$ & $12.0$ & $78.0$ & $33.0$ & $78.0$ & $99.0$ & $68.0$ & $33.0$ & $81.0$ & $1.0$ \\
zul\_latn & $76.0$ & $15.0$ & $68.0$ & $100.0$ & $69.7$ & $0.0$ & $63.0$ & $3.0$ & $65.0$ & $92.0$ & $63.0$ & $90.0$ & $67.0$ & $1.0$ \\
\midrule
\textbf{Avg} & $72.4$ & $29.6$ & $70.7$ & $\mathbf{98.3}$ & $\mathbf{78.7}$ & $9.5$ & $78.3$ & $13.7$ & $77.2$ & $94.7$ & $59.4$ & $92.3$ & $77.3$ & $49.8$ \\
Std & $9.4$ & $29.3$ & $9.6$ & $8.3$ & $10.5$ & $16.2$ & $10.9$ & $17.3$ & $13.1$ & $15.6$ & $9.0$ & $21.5$ & $10.5$ & $43.4$ \\
\bottomrule
\end{tabular}%
}
}
\label{tab:app-globalpiqa}
\end{table}

\begin{table}[h]
\centering
\caption{Per-language task accuracy (Acc) and L2 reasoning rate (L2\%) on \textbf{\macaron{}} (20 languages, 0–100). Languages are ordered alphabetically by language code. Avg / Std are over languages; the highest average in each metric is bold.}
\scriptsize{%
\setlength{\tabcolsep}{2pt}
\resizebox{\textwidth}{!}{%
\begin{tabular}{l|cccc|cccccccccc}
\toprule
Language & \multicolumn{2}{c}{\tinyayaen{}} & \multicolumn{2}{c|}{\tinyayamulti{}} & \multicolumn{2}{c}{\qwensmallnew{}} & \multicolumn{2}{c}{\begin{tabular}{@{}c@{}}\qwensmallnew{}\\{\scriptsize(user prefix LF)}\end{tabular}} & \multicolumn{2}{c}{\begin{tabular}{@{}c@{}}\qwensmallnew{}\\{\scriptsize(thinking prefix LF)}\end{tabular}} & \multicolumn{2}{c}{\mthinker{}} & \multicolumn{2}{c}{\magistralsmall{}} \\
\cmidrule(lr){2-3}\cmidrule(lr){4-5}\cmidrule(lr){6-7}\cmidrule(lr){8-9}\cmidrule(lr){10-11}\cmidrule(lr){12-13}\cmidrule(lr){14-15}
 & Acc & L2\% & Acc & L2\% & Acc & L2\% & Acc & L2\% & Acc & L2\% & Acc & L2\% & Acc & L2\% \\
\midrule
brazil & $43.0$ & $15.0$ & $48.0$ & $98.0$ & $55.0$ & $12.0$ & $54.0$ & $27.0$ & $61.6$ & $96.0$ & $42.0$ & $100.0$ & $44.0$ & $66.0$ \\
china & $60.8$ & $0.0$ & $48.5$ & $96.9$ & $62.9$ & $2.1$ & $70.1$ & $3.1$ & $68.0$ & $97.9$ & $60.8$ & $100.0$ & $65.0$ & $8.2$ \\
egypt & $39.4$ & $20.2$ & $34.3$ & $99.0$ & $33.3$ & $7.1$ & $26.3$ & $4.0$ & $31.3$ & $100.0$ & $27.3$ & $99.0$ & $40.4$ & $3.0$ \\
ethiopia & $34.7$ & $0.0$ & $34.7$ & $100.0$ & $4.1$ & $1.0$ & $0.0$ & $2.0$ & $1.0$ & $97.9$ & $21.4$ & $82.7$ & $22.4$ & $0.0$ \\
georgia & $23.2$ & $50.5$ & $21.2$ & $40.4$ & $26.3$ & $11.1$ & $13.1$ & $18.2$ & $27.3$ & $99.0$ & $36.4$ & $100.0$ & $45.5$ & $6.1$ \\
greece & $53.0$ & $21.6$ & $39.0$ & $100.0$ & $42.0$ & $26.0$ & $39.0$ & $33.0$ & $47.0$ & $100.0$ & $31.0$ & $100.0$ & $57.0$ & $10.0$ \\
india & $54.0$ & $20.0$ & $47.0$ & $100.0$ & $51.0$ & $42.0$ & $46.0$ & $21.0$ & $53.5$ & $96.0$ & $29.0$ & $100.0$ & $78.0$ & $5.0$ \\
indonesia & $49.5$ & $62.8$ & $48.4$ & $100.0$ & $43.6$ & $44.2$ & $36.2$ & $43.6$ & $41.0$ & $92.6$ & $36.8$ & $100.0$ & $61.0$ & $11.6$ \\
italy & $61.2$ & $9.2$ & $57.1$ & $99.0$ & $50.0$ & $18.4$ & $51.0$ & $19.4$ & $70.4$ & $96.9$ & $37.8$ & $100.0$ & $55.1$ & $93.9$ \\
japan & $51.5$ & $2.0$ & $49.5$ & $100.0$ & $41.4$ & $14.1$ & $48.0$ & $16.3$ & $52.5$ & $98.0$ & $33.3$ & $100.0$ & $77.1$ & $0.0$ \\
kyrgyzstan & $26.0$ & $5.0$ & $20.4$ & $81.6$ & $29.0$ & $2.0$ & $13.1$ & $1.0$ & $21.0$ & $97.0$ & $35.0$ & $0.0$ & $46.5$ & $1.0$ \\
mexico & $43.4$ & $9.2$ & $41.4$ & $100.0$ & $53.5$ & $12.1$ & $47.5$ & $12.1$ & $59.6$ & $96.0$ & $39.4$ & $100.0$ & $38.4$ & $67.7$ \\
morocco & $36.4$ & $10.1$ & $39.0$ & $100.0$ & $28.0$ & $6.0$ & $14.0$ & $5.0$ & $36.4$ & $100.0$ & $32.0$ & $100.0$ & $36.4$ & $2.0$ \\
nigeria & $43.6$ & $0.0$ & $48.9$ & $61.7$ & $20.2$ & $0.0$ & $5.4$ & $0.0$ & $0.0$ & $4.3$ & $23.4$ & $0.0$ & $43.6$ & $0.0$ \\
philippines & $41.4$ & $11.3$ & $39.4$ & $100.0$ & $34.3$ & $3.0$ & $24.2$ & $2.0$ & $27.3$ & $85.9$ & $25.2$ & $100.0$ & $57.6$ & $0.0$ \\
south\_africa & $38.0$ & $4.0$ & $38.0$ & $100.0$ & $19.0$ & $6.0$ & $2.0$ & $6.0$ & $13.0$ & $75.0$ & $23.0$ & $76.0$ & $35.0$ & $1.0$ \\
thailand & $41.4$ & $26.5$ & $35.4$ & $100.0$ & $25.2$ & $23.2$ & $14.1$ & $23.2$ & $37.8$ & $91.8$ & $28.3$ & $100.0$ & $45.3$ & $11.6$ \\
tunisia & $34.3$ & $9.3$ & $31.0$ & $100.0$ & $35.0$ & $2.0$ & $20.0$ & $3.0$ & $26.0$ & $100.0$ & $30.0$ & $98.0$ & $41.8$ & $1.0$ \\
turkey & $46.0$ & $6.2$ & $43.0$ & $100.0$ & $46.0$ & $21.0$ & $41.0$ & $24.0$ & $46.0$ & $97.0$ & $39.4$ & $99.0$ & $61.6$ & $0.0$ \\
yemen & $31.3$ & $21.2$ & $32.0$ & $100.0$ & $41.0$ & $3.0$ & $39.0$ & $6.0$ & $34.0$ & $100.0$ & $23.0$ & $99.0$ & $40.4$ & $6.1$ \\
\midrule
\textbf{Avg} & $42.6$ & $15.2$ & $39.8$ & $\mathbf{93.8}$ & $37.0$ & $12.8$ & $30.2$ & $13.5$ & $37.7$ & $91.1$ & $32.7$ & $87.7$ & $\mathbf{49.6}$ & $14.7$ \\
Std & $10.1$ & $16.0$ & $9.2$ & $15.2$ & $14.0$ & $12.6$ & $19.2$ & $12.0$ & $19.7$ & $20.7$ & $8.8$ & $29.9$ & $13.8$ & $26.4$ \\
\bottomrule
\end{tabular}%
}
}
\label{tab:app-macaron_mcq}
\end{table}

\section{Doomlooping score scatter plots}
\begin{figure}[!p]
    \centering
    \includegraphics[width=1.0\linewidth]{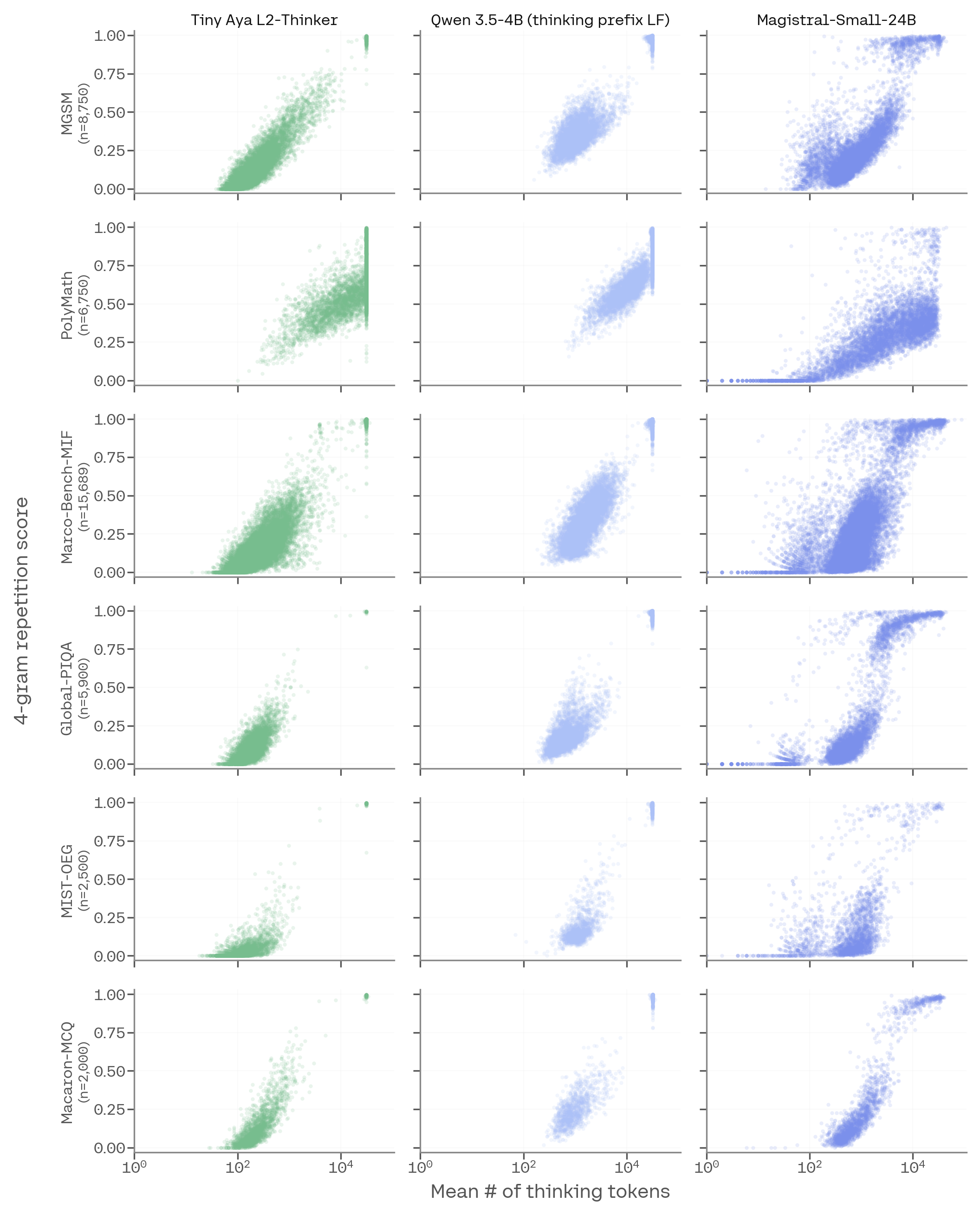}
    \caption{Per-sample doomlooping vs.\ thinking length. Each point is one non-empty thinking trace (log reasoning length vs.\ 4-gram repetition score where higher suggests more redundant traces). $n$ is the number of samples across all languages. \qwensmallnew{}  with thinking-prefix language forcing produces systematically longer traces, while \tinyayamulti{} generates more efficient thinking traces, spending extra tokens only on hard math set (PolyMath). \magistralsmall{} samples are more scattered and show weaker correlations between 4-gram repetition score and length of thinking traces.   
}
    \label{fig:doomloop_scatter}
\end{figure}

Figure~\ref{fig:doomloop_scatter} shows per-sample doomlooping scores vs.\ thinking length. Each point is one non-empty thinking trace across all languages in the respective task. Mean number of thinking tokens is on a log scale and 4-gram repetition score is used as a proxy for doomlooping where higher suggests more redundant traces. $n$ is the number of samples across all languages. \qwensmallnew{}  with thinking-prefix language forcing produces systematically longer traces: almost every sample is well above 100 tokens, and a large fraction pile up at the max-token budget on every benchmark. \tinyayamulti{} adapts length to the task: most traces sit near $10^{2}$ tokens on MGSM, \mif{}, GlobalPIQA, \mist{}, and \macaron{}, and only stretch---often to the token cap---on the hard math set (PolyMath). For \tinyayamulti{} and \qwensmallnew{} the link between longer traces and doomlooping score is tight; for \magistralsmall{} it is weaker and the cloud is more diffuse.

\section{Ablations}\label{app:ablations}

Our analyses in \Cref{sec:controlled_exps} assume a single model trained by joint mixing. We now justify that choice, comparing mixing against two strategies that fragment the process---sequentially adapting an English-only reasoner, and merging separately trained specialists---and examining how each fails when it cannot reason in the target language. For a controlled comparison we reuse the ten-language, five-region setup and specialist models of \Cref{sec:lang_scale}.

\subsection{Joint mixing gives the best accuracy--L2 trade-off}
\label{sec:training_strategies_tradeoff}

\begin{figure}[!p]
    \centering
    \includegraphics[width=0.8\linewidth]{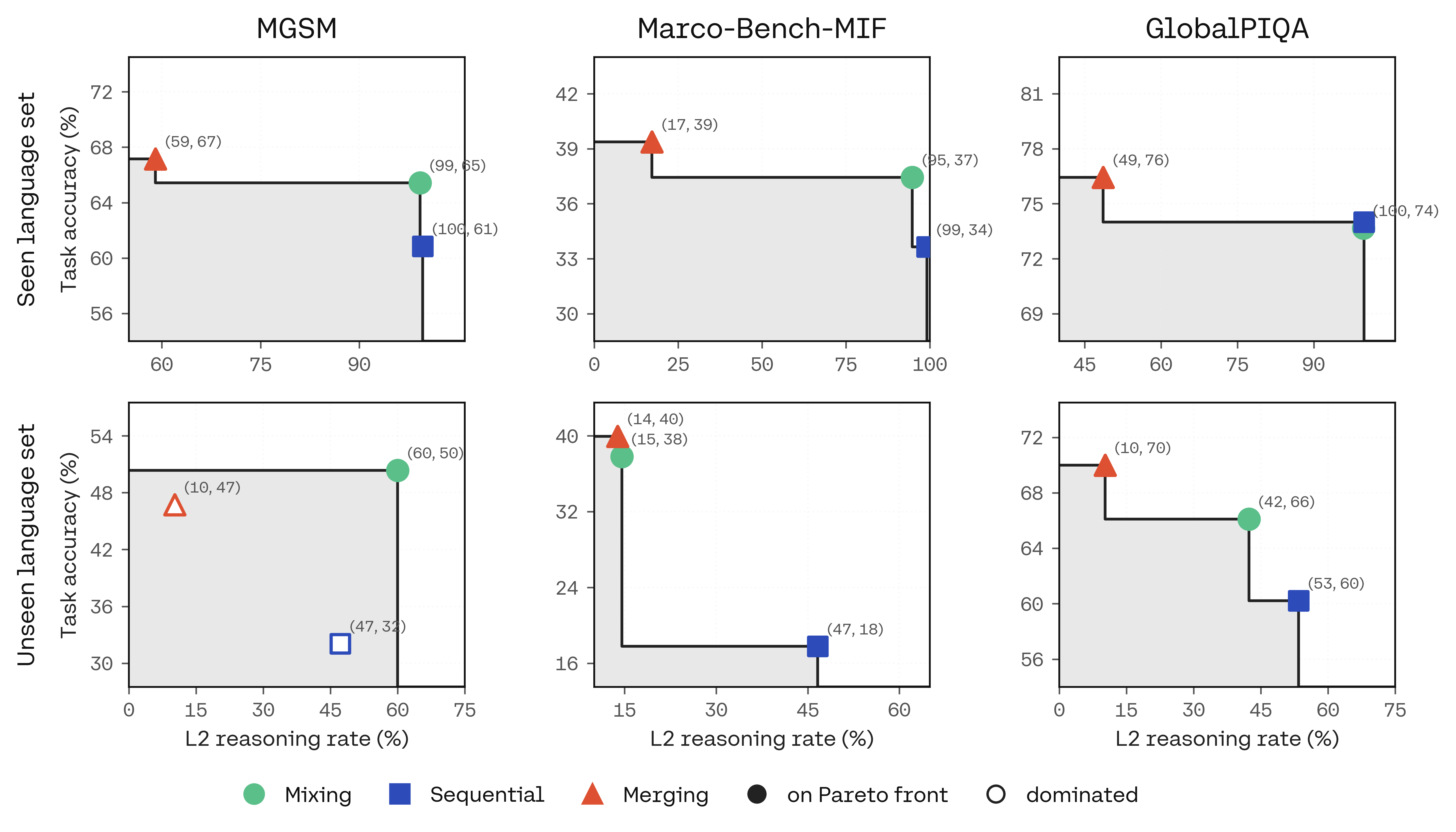}
    \caption{Pareto front of post-training strategies (mixing, merging, and sequential adaptation) for the task accuracy vs.\ L2 reasoning rate trade-off. Mixing is never dominated.}
    \label{fig:training_strategies_pareto}
\end{figure}

In this section we ask: does the timing of multilingual supervision matter? We compare three ways of combining English reasoning with L2 supervision under a fixed training budget. \emph{Mixing} fine-tunes the base model on English reasoning and all ten languages' L2 data at once (All-Mixed). \emph{Sequential} adaptation first fine-tunes on the English mix to obtain an English-only reasoner, then fine-tunes on L2 data (All-Seq). \emph{Merging} fine-tunes the regional specialists of \Cref{sec:lang_scale} separately and averages their weights linearly (Spec-mrg). \Cref{fig:training_strategies_pareto} reports task accuracy and L2 reasoning rate on seen and unseen languages per benchmark; numbers below are ordered MGSM/\mif{}/GlobalPIQA.

The three strategies trace an accuracy--L2 reasoning rate trade-off (\Cref{fig:training_strategies_pareto}). Merging wins on task accuracy but collapses in-language reasoning: averaging weights cancels each specialist's language conditioning, so the merged model reasons in English regardless of the prompt (\Cref{fig:unseen-thinking-fallback}), its seen-language in-language rate falling to 59/17/49 against 99/95/100 for mixing. Sequential adaptation trades the other way: it reaches for the prompt's language most often on held-out inputs, but forgets part of the base English capability, lowering accuracy. Mixing alone is never dominated: it achieves near-ceiling in-language reasoning on seen languages, task accuracy within a few points of merging, and on unseen languages it gives up some in-language reasoning for higher accuracy than sequential and a more predictable fallback (\Cref{sec:fallback-lang}); on MGSM's unseen languages it is best on both axes.

\subsection{Joint mixing fails predictably; sequential scatters}
\label{sec:fallback-lang}

\begin{figure}[!p]
\centering
\includegraphics[width=\linewidth]{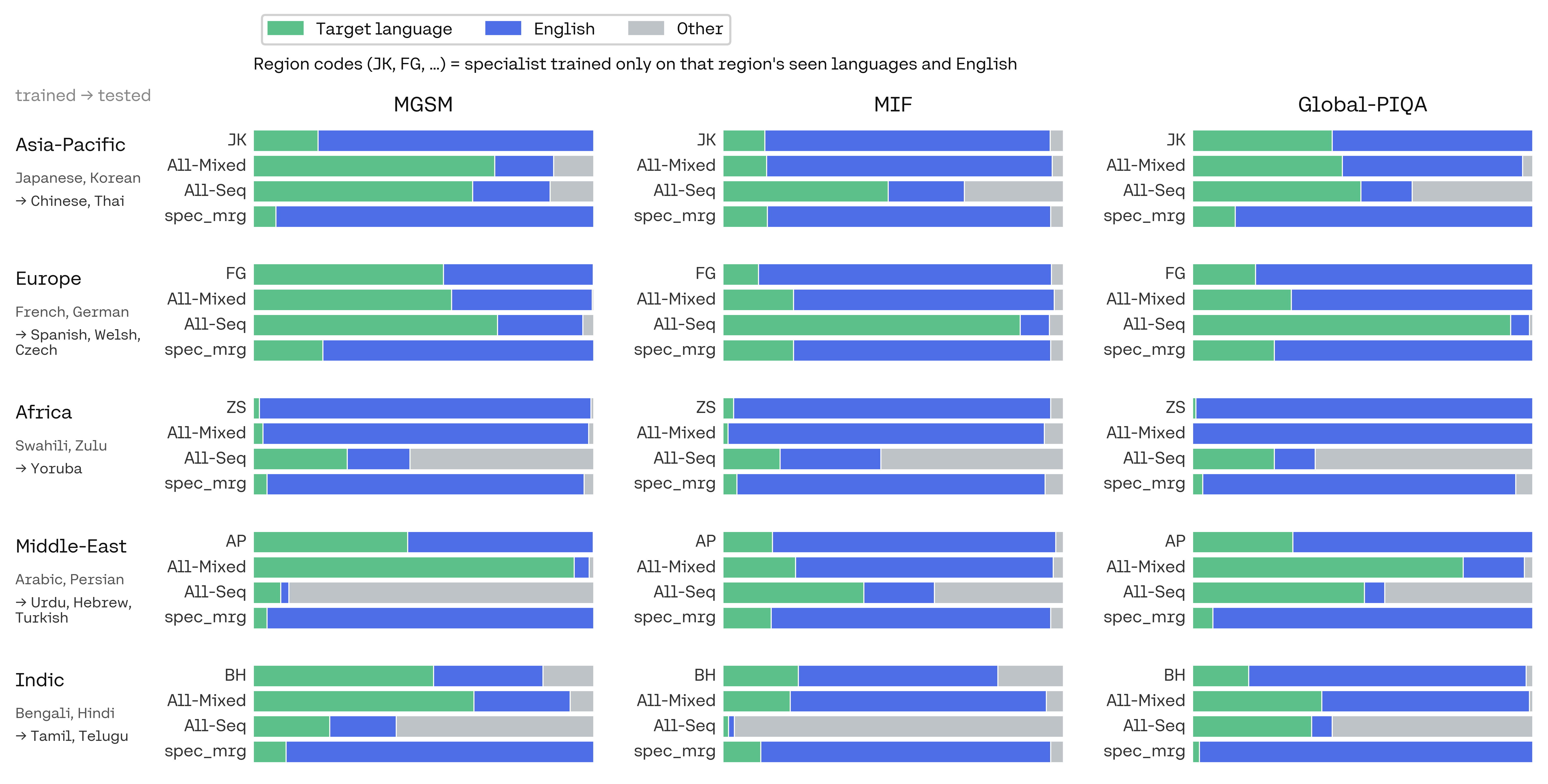}
\caption{Thinking-language distribution on unseen languages, by region (rows) and benchmark (columns). Each cell stacks four bars---region specialist, All-Mix, sequential, and linearly merged specialists---split into target language, English, and other. Joint training falls back to English predictably; sequential training scatters the fallback across other languages; the merged model reasons almost entirely in English.}
\label{fig:unseen-thinking-fallback}
\end{figure}

A model's fallback language---which language it reasons in when it cannot reason in the user's language---determines how auditable the failure is: a consistent fallback can be anticipated downstream, whereas unpredictable switching cannot. \Cref{fig:unseen-thinking-fallback} breaks down the thinking language on \emph{unseen} languages across the three benchmarks and five regions for four models. Joint mixing falls back stably to English; the region specialists behave similarly but reason in-language far less often (consistent with \Cref{sec:lang_scale}), and the merged model reasons primarily in English. Sequential adaptation instead scatters its fallback across other languages seen during later fine-tuning, so its thinking language on unseen inputs is far harder to predict---the fragility quantified in \Cref{sec:training_strategies_tradeoff}. Joint mixing thus keeps English as a stable, auditable fallback while conditioning on the input; sequential adaptation creates competing associations that resist prediction and correction.

\clearpage
\section{Reasoning Errors}\label{sec:reasoning_errors}
Below are examples of reasoning traces and generations for an example prompt from the \mist{} dataset. \tinyayamulti{} successfully reasons and generates answer in the given prompt language (Korean) whereas \magistralsmall{} reasons primarily in English, and generates an answer in a mix of German, English, and Korean. \tinyayamulti{} achieves a higher overall score on the task while maintaining perfect language consistency in both reasoning and generation. Additionally \magistralsmall{} has a much more verbose reasoning trace compared to \tinyayamulti{} whilst having lower overall scores. \tinyayamulti{} in particular has higher scores in metrics of coherence and naturalness compared to \magistralsmall{}. Note for brevity we annotate traces with [...] but otherwise these traces would additionally be longer and more verbose or repetitive.\\\\
\vspace{-25pt}
\begin{center}
\begin{promptbox}{Reasoning Trace Errors - \magistralsmall{} Code-Switches Frequently, while \tinyayamulti{} does not}
\textbf{Prompt:} "Think in the same language as the prompt. 결혼식 초대장을 거절하는 게 무례한 일인가요? 여행 경비가 부담돼 참석하기 어려운 상황인데 금전적인 이유를 언급하지 않고 초대장을 거절하는 메시지를 쓰고 싶어요. 도와주실 수 있을까요?"\\\\
\textbf{Translated Prompt:} "Think in the same language as the prompt. Is it rude to decline a wedding invitation? I am unable to attend due to the burden of travel expenses, but I want to write a message declining the invitation without mentioning financial reasons. Could you help me?"\\\\
\textbf{\magistralsmall{} Reasoning Trace:} "Okay, the user is asking if declining a wedding invitation is rude, especially when the reason is financial difficulty due to travel costs. They want help drafting a message that doesn't mention money directly. Let me start by considering the cultural context—Korean weddings often place a lot of importance on attendance and relationships. Declining might be seen as rude, so the message needs to be polite and respectful.\texttt{\textbackslash n\textbackslash n}First, I need to think about possible reasons to decline that don't involve money. Maybe health, prior commitments, or other personal reasons. Since the issue is travel costs, perhaps they can say they have other plans or are unable to travel at that time.\texttt{\textbackslash n\textbackslash n}Now, how to phrase this politely? In Korean culture, it's important to express gratitude and regret. So, the message should start by thanking the couple for the invitation, then express regret, and give a vague but respectful reason.\texttt{\textbackslash n\textbackslash n}Here’s a draft:\texttt{\textbackslash n\textbackslash n}"안녕하세요 [호칭]님,\texttt{\textbackslash n}[결혼식 날짜]에 개최되는 결혼式에 초대해 주시 Thank you for the invitation to your wedding on [date]. We are truly honored and grateful for your thoughtfulness.\texttt{\textbackslash n\textbackslash n}Unfortunately, due to prior commitments, we will not be able to attend the wedding. We sincerely apologize for any inconvenience this may cause.\texttt{\textbackslash n\textbackslash n}Please know that we will be celebrating this special day with you in spirit. We wish you both a lifetime of happiness and love."\texttt{\textbackslash n\textbackslash n}But the user wants to avoid mentioning money, so perhaps saying “due to unforeseen circumstances” or “personal matters” might be better.\texttt{\textbackslash n\textbackslash n}Let me refine this:\texttt{\textbackslash n\textbackslash n}"안녕하세요 [호칭]님,\texttt{\textbackslash n}[결혼식 날짜]에 개최되는 결혼식 초대를 감사히 받았습니다. 진심으로 감사드리며, 이 중요한 순간을 함께할 수 있을 줄 알고 excitement을 느꼈습니다.\texttt{\textbackslash n\textbackslash n}다만, 현재 상황상 참여가 difficulty게됩니다. deeply regret that we cannot be there to celebrate with you in person.\texttt{\textbackslash n\textbackslash n}please understand that our absence is not a reflection of our feelings for you. We will be cheering for your happiness from afar and wish you both all the best in your new journey together.\texttt{\textbackslash n\textbackslash n}감사합니다."\texttt{\textbackslash n\textbackslash n}But maybe this is a bit long. Perhaps a more concise version would be better.\texttt{\textbackslash n\textbackslash n}Let me try again:\texttt{\textbackslash n\textbackslash n}"안녕하세요 [호칭]님,\texttt{\textbackslash n}[결혼식 날짜] 결혼식 초대를 감사히 받았습니다. 진심으로 감사드리며, 이 중요한 날을 함께할 수 있을 줄 알고 joy를 느꼈습니다.\texttt{\textbackslash n\textbackslash n}다만, 현재 상황상 참석하기 difficult가 있습니다. deeply regret this and apologize for any disappointment.\texttt{\textbackslash n\textbackslash n} please know that we will be thinking of you both on your special day and wishing you a lifetime of love and happiness.\texttt{\textbackslash n\textbackslash n}감사합니다."\texttt{\textbackslash n\textbackslash n}This seems more concise and polite. The reason is left vague, which fits the user's request to not mention money.\texttt{\textbackslash n\textbackslash n}Now, to present this to the user, I'll suggest this template and maybe provide a few variations to choose from." \\\\
\textbf{\magistralsmall{} Generation:} "결혼식 초 Großes zu verpassen ist oft unangenehm, aber es gibt respektvolle Möglichkeiten, dies auszudrücken, ohne die finanziellen Gründe zu erwähnen. Hier sind ein paar Vorschläge für Ihre Absage:\texttt{\textbackslash n\textbackslash n}1. **Vage, aber höfliche Begründung:**\texttt{\textbackslash n}```\texttt{\textbackslash n}안녕하세요 [호칭]님,\texttt{\textbackslash n\textbackslash n}[결혼식 날짜]에 개최되는 결혼식 초대를 감사히 받았습니다. 진심으로 감사드리며, 이 중요한 날을 함께할 수 있을 줄 알고 기쁘게 생각했습니다.\texttt{\textbackslash n\textbackslash n}다만, 현재 상황상 참석이 difficult가 있습니다. deeply regret this and apologize for any disappointment caused.\texttt{\textbackslash n\textbackslash n}please understand that our absence is not a reflection of our affection for you. We will be celebrating your special day in spirit and wish you both a lifetime of happiness.\texttt{\textbackslash n\textbackslash n}감사합니다.\texttt{\textbackslash n}```\texttt{\textbackslash n\textbackslash n}2. **Persönliche Verpflichtungen als Grund:**\texttt{\textbackslash n}```\texttt{\textbackslash n}안녕하세요 [호칭]님,\texttt{\textbackslash n\textbackslash n}[결혼식 날짜] 결혼식 초대를 감사히 받았습니다. 진심으로 감사드리며, 이 날을 함께할 수 있을 줄 알고 joy를 느꼈습니다.\texttt{\textbackslash n\textbackslash n}다만, 이미 예약된 중요한 일정으로 인해 참석하기 difficult가 있습니다. deeply regret that we cannot be there to celebrate with you.\texttt{\textbackslash n\textbackslash n} please know that we will be thinking of you both and wishing you all the best on your special day.\texttt{\textbackslash n\textbackslash n}감사합니다.\texttt{\textbackslash n}```\texttt{\textbackslash n\textbackslash n}3. **Kurze und herzliche Version:**\texttt{\textbackslash n}```\texttt{\textbackslash n}안녕하세요 [호칭]님,\texttt{\textbackslash n\textbackslash n}[결혼식 날짜] 결혼식 초대를 감사히 받았습니다. 진심으로 감사드리며, 이 날을 함께할 수 있을 줄 알고 joy를 느꼈습니다.\texttt{\textbackslash n\textbackslash n}다만, 현재 situation로 인해 참석하기 difficult가 있습니다. deeply regret this and apologize for any inconvenience.\texttt{\textbackslash n\textbackslash n}please accept our warmest wishes for your happiness and a beautiful future together.\texttt{\textbackslash n\textbackslash n}감사합니다.\texttt{\textbackslash n}```\texttt{\textbackslash n\textbackslash n}Tipp: Fügen Sie persönliche Worte hinzu, z.B. wenn Sie den Brautpaar gut kennen, um die Botschaft herzlicher zu gestalten. Beispiel: "Wir haben Ihre Beziehung immer bewundert und wünschen Ihnen alles Gute für Ihre gemeinsame Zukunft."\texttt{\textbackslash n\textbackslash n}Möchten Sie eine bestimmte Version anpassen oder einen anderen Stil bevorzugen?"\\\\
\textbf{Mean Score:} 1.5\\
\textbf{Accuracy:} 1\\
\textbf{Coherence:} 1\\
\textbf{Naturalness:} 1\\
\textbf{Instruction Following:} 3\\\\
\textbf{\tinyayamulti{} Reasoning Trace:} "사용자가 거절 편지를 작성해 달라고 요청했습니다. 거절하는 것도 무례하다고 생각할 수 있지만, 재정적 문제로 인해 참석할 수 없다는 정당한 이유에 대한 정중한 거절 편지입니다. 이는 허용됩니다. 정책 위반이 아닙니다. 정중한 메시지를 제공하세요." \\\\
\textbf{\tinyayamulti{} Generation:} "제목: [결혼 날짜] 행사 초대장에 대한 답변

[신부/신랑 이름]님께,

이 메시지가 잘 전달되길 바랍니다.

[날짜]에 예정된 [결혼 장소/ events 장소]에서의 결혼식 초대장에 대해 연락드린 것은 정말 기쁜 일이었습니다. 이 특별한 날의 모든 세부 사항을 세심하게 준비하고 계신 모습에 감탄합니다.

안타깝게도, 저는 그 행사에 참석할 수 없게 되었습니다. 예상치 못한 재정적 상황이 발생하여, 참석 fees에 대한 부담으로 인해 이 초대를 거절해야 하는 어려운 결정을 내렸습니다. 저는 이 결정이 완전히 개인적이며, 결혼식에 대한 존중이나 감사의 표현은 아니라는 점을 명확히 하고 싶습니다.

이 초대를 받아주신 것에 대해 진심으로 감사드립니다. 이 특별한 날의 기억이 여러분과 신부/신랑님께 평화와 기쁨을 안겨주길 바랍니다. 다른 행사에서 기꺼이 여러분을 환영합니다.

진심으로,

[당신의 이름]
[선택 사항: 가족/친구 이름]
[선택 사항: 연락처 정보]
" \\\\
\textbf{Mean Score:} 3\\
\textbf{Accuracy:} 1\\
\textbf{Coherence:} 5\\
\textbf{Naturalness:} 5\\
\textbf{Instruction Following:} 1\\\\
\end{promptbox}
\end{center}
Below are more examples of reasoning traces and generations for an example prompt from the \mist{} dataset. \tinyayamulti{} successfully reasons and generates answer in the given prompt language (Korean) whereas \qwensmallnew{} reasons primarily in English, and also fails to generate an answer due to doomlooping. \tinyayamulti{} achieves a higher overall score on the task while maintaining perfect language consistency in both reasoning and generation.

\begin{center}
\begin{promptbox}{Reasoning Trace Errors - Doomlooping in \qwensmallnew{} compared to none in \tinyayamulti{}}
\textbf{Prompt:} "Think in the same language as the prompt. 선생님은 우정이 중요하다고 말씀하셨는데 친구 없이도 행복하게 살 수 있는지 궁금해요. 지금 친구가 많지는 않지만 그래도 괜찮은 것 같아요."
\\\\
\textbf{Translated Prompt:} "Think in the same language as the prompt. You said that friendship is important, but I wonder if it is possible to live happily without friends. I don't have many friends right now, but I think that's okay.
"\\\\
\textbf{\qwensmallnew{} Reasoning Trace:} ""Thinking Process:\texttt{\textbackslash n\textbackslash n}1.  **Analyze the Request:**\texttt{\textbackslash n}   *   **Input Language:** Korean ("선생님은 우정이 중요하다고 말씀하셨는데 친구 없이도 행복하게 살 수 있는지 궁금해요. 지금 친구가 많지는 않지만 그래도 괜찮은 것 같아요.")\texttt{\textbackslash n}   *   **Task:** Respond to the user in the same language (Korean).\texttt{\textbackslash n}   *   **Context:** The user is asking about the importance of friendship (based on what their teacher said) and whether it's possible to be happy without friends. They mention they don't have many friends but feel okay about that.\texttt{\textbackslash n}   *   **Tone:** Empathetic, supportive, thoughtful, and conversational.\texttt{\textbackslash n\textbackslash n}2.  **Determine the Core Message:**\texttt{\textbackslash n}   *   [...] Validate their current state: "지금 친구가 많지 않더라도 괜찮은 것 같아요. 혼자만의 시간을 즐기거나, 가족이나 가족과 소통하는 것만으로도 충분히 만족할 수 있죠." (Having few friends is okay too. Enjoying alone time or communicating with family can be enough.)\texttt{\textbackslash n}   *   Emphasize self-love: "자신에게도 친구가 되는 것, 곧 자기 자신과 잘 지내는 것도 중요한 우정일 수 있어요." (Being friends with yourself is also important friendship.)\texttt{\textbackslash n}   * [...] Wait, I need to check if there are any specific constraints. "Think in the same language as the prompt." -> Output in Korean.\texttt{\textbackslash n}   Okay.\texttt{\textbackslash n\textbackslash n}    Let's produce the response.\texttt{\textbackslash n}   (Wait, I should check if I need to be more concise or detailed. The prompt is short, so a moderate length response is good.)\texttt{\textbackslash n\textbackslash n}    Okay, ready to generate.\texttt{\textbackslash n\textbackslash n}    Wait, one more thing. The prompt says "Think in the same language as the prompt." [...] Okay, I will write it naturally.\texttt{\textbackslash n}   "선생님의 말씀은 정말 귀중하죠. 우정이 중요한 건 맞아요. 하지만 친구가 없어도 행복하게 살 수 있다는 건, 사실 충분히 가능해요."\texttt{\textbackslash n}   "지금 친구가 많지 않더라도 괜찮은 것 같아요. 중요한 건 친구의 수보다, 마음의 평온과 자신감일 거예요."\texttt{\textbackslash n}   "혼자만의 시간을 즐기거나, 가족과 소통하는 것만으로도 충분히 행복할 수 있죠."\texttt{\textbackslash n}   "결국 행복은 외부의 관계보다 내면의 평온에서 더 많이 찾아올 수 있죠."\texttt{\textbackslash n}   "지금의 마음이 편안하다면, 그 상태를 존중하고 응원해 드릴게요."\texttt{\textbackslash n\textbackslash n}    Okay, I will combine these.\texttt{\textbackslash n}   "선생님의 말씀은 정말 귀중하죠. 우정이 중요한 건 맞아요. 하지만 친구가 없어도 행복하게 살 수 있다는 건, 사실 충분히 가능해요."\texttt{\textbackslash n}   "지금 친구가 많지 않더라도 괜찮은 것 같아요. 중요한 건 친구의 수보다, 마음의 평온과 자신감일 거예요."\texttt{\textbackslash n}   "혼자만의 시간을 즐기거나, 가족과 소통하는 것만으로도 충분히 행복할 수 있죠."\texttt{\textbackslash n}   "결국 행복은 외부의 관계보다 내면의 평온에서 더 많이 찾아올 수 있죠."\texttt{\textbackslash n}   "지금의 마음이 편안하다면, 그 상태를 존중하고 응원해 드릴게요."\texttt{\textbackslash n\textbackslash n}    Wait, I should not repeat myself.\texttt{\textbackslash n}   Okay, final version:\texttt{\textbackslash n}   선생님 말씀은 정말 귀중하죠. 우정이 중요한 건 맞아요. 하지만 친구가 없어도 행복하게 살 수 있다는 건, 사실 충분히 가능해요.\texttt{\textbackslash n\textbackslash n}    지금 친구가 많지 않더라도 괜찮은 것 같아요. 중요한 건 친구의 수보다, 마음의 평온과 자신감일 거예요. 혼자만의 시간을 즐기거나, 가족과 소통하는 것만으로도 충분히 행복할 수 있죠.\texttt{\textbackslash n\textbackslash n}    결국 행복은 외부의 관계보다 내면의 평온에서 더 많이 찾아올 수 있죠. 지금의 마음이 편안하다면, 그 상태를 존중하고 응원해 드릴게요. 친구가 필요할 때 찾아오는 관계도 좋지만, 지금처럼 혼자서도 충분히 빛날 수 있다는 걸 믿어주셔도 괜찮아요.\texttt{\textbackslash n\textbackslash n} [...] Wait, I need to make sure I don't violate any policies. No issues.\texttt{\textbackslash n}   Okay, final check.\texttt{\textbackslash n}   Input: Korean.\texttt{\textbackslash n}   Output: Korean.\texttt{\textbackslash n}   Content: Supportive response to the user's question about happiness without friends.\texttt{\textbackslash n}[...] ready.\texttt{\textbackslash n}   Wait, I should check if I need to add a title. No.\texttt{\textbackslash n}   Okay, just the text.\texttt{\textbackslash n}   Okay, done.\texttt{\textbackslash n}   Wait, I should check if I need to add a signature. No.\texttt{\textbackslash n}   Okay, just the text.\texttt{\textbackslash n}   Okay, done.\texttt{\textbackslash n}   Wait, I should check if I need to add any disclaimer. No.\texttt{\textbackslash n}   Okay, just the text.\texttt{\textbackslash n}   Okay, done.\texttt{\textbackslash n}   Wait, I should check if I need to add any formatting. No.\texttt{\textbackslash n}   Okay, just the text.\texttt{\textbackslash n}   Okay, done.\texttt{\textbackslash n}   Wait, I should check if I need to add any emojis. No.\texttt{\textbackslash n}   Okay, just the text.\texttt{\textbackslash n}   Okay, done.\texttt{\textbackslash n}   Wait, I should check if I need to add any links. No.\texttt{\textbackslash n}   Okay, just the text.\texttt{\textbackslash n}   Okay, done.\texttt{\textbackslash n}   Wait, I should check if I need to add any images. No.\texttt{\textbackslash n}   Okay, just the text.\texttt{\textbackslash n}   Okay, done.\texttt{\textbackslash n}   Wait, I should check if I need to add any audio. No.\texttt{\textbackslash n}   Okay, just the text.\texttt{\textbackslash n}   Okay, done.\texttt{\textbackslash n}[...] Okay, done.\texttt{\textbackslash n}   Wait, I should check if I need to add any production. No.\texttt{\textbackslash n}   Okay, just the text.\texttt{\textbackslash n}   Okay, done.\texttt{\textbackslash n}   Wait, I should check if I need to add any construction. No.\texttt{\textbackslash n}   Okay, just the text.\texttt{\textbackslash n}   Okay, done.\texttt{\textbackslash n}   Wait, I should check if I need to add any architecture. No.\texttt{\textbackslash n}   Okay, just the text.\texttt{\textbackslash n}   Okay, done.\texttt{\textbackslash n}   Wait, I should check if I need to add any design. No.\texttt{\textbackslash n}   Okay, just the text.\texttt{\textbackslash n}   Okay, done.\texttt{\textbackslash n}   Wait, I should check if I need to add any planning. No.\texttt{\textbackslash n}   Okay, just the text.\texttt{\textbackslash n}   Okay, done.\texttt{\textbackslash n}   Wait, I should check if I need to add any strategy. No.\texttt{\textbackslash n}   Okay, just the text.\texttt{\textbackslash n}   Okay, done.\texttt{\textbackslash n}   Wait, I should check if I need to add any management. No.\texttt{\textbackslash n}   Okay, just the text.\texttt{\textbackslash n}   Okay, done.\texttt{\textbackslash n}   Wait, I should check if I need to add any leadership. No.\texttt{\textbackslash n}   Okay, just the text.\texttt{\textbackslash n}   Okay, done.\texttt{\textbackslash n}   \textbackslash{} [...]
should check if I need to add any smartphone. No.\texttt{\textbackslash n}   Okay, just the text.\texttt{\textbackslash n}   Okay, done.\texttt{\textbackslash n}   Wait, I should check if I need to add any phone. No.\texttt{\textbackslash n}[...] Okay, just the text.\texttt{\textbackslash n}   Okay, done.\texttt{\textbackslash n}   Wait, I should check if I need to add any tablet. No.\texttt{\textbackslash n}   Okay, just the text.\texttt{\textbackslash n}   Okay, done.\texttt{\textbackslash n}   Wait, I should check if I need to add any phone. No.\texttt{\textbackslash n}   Okay, just the text.\texttt{\textbackslash n}   Okay, done.\texttt{\textbackslash n}   Wait, I should check if I need to add any mobile. No.\texttt{\textbackslash n}   Okay, just the text.\texttt{\textbackslash n}   Okay, done.\texttt{\textbackslash n}   Wait, I should check if I need to add any smartphone. No.\texttt{\textbackslash n}   Okay, just the text.\texttt{\textbackslash n}   Okay, done.\texttt{\textbackslash n}   Wait, I should check if I need to add any phone. No.\texttt{\textbackslash n}   Okay, just the text.\texttt{\textbackslash n}   Okay, done.\texttt{\textbackslash n}   Wait, I should check if I need to add any device. No.\texttt{\textbackslash n}   Okay, just the text.\texttt{\textbackslash n}   Okay, done.\texttt{\textbackslash n}   Wait, I should check if I need to add any gadget. No.\texttt{\textbackslash n}   Okay, just the text.\texttt{\textbackslash n}   Okay, done.\texttt{\textbackslash n}   Wait, I should check if I need to add any tool. No.\texttt{\textbackslash n}   Okay, just the text.\texttt{\textbackslash n}   Okay, done."
" \\\\
\textbf{\qwensmallnew{} Generation:} None \\
\textbf{Mean Score:} 1\\
\textbf{Accuracy:} 1\\
\textbf{Coherence:} 1\\
\textbf{Naturalness:} 1\\
\textbf{Instruction Following:} 1\\\\
\textbf{\tinyayamulti{} Reasoning Trace:} "사용자는 우정이 중요하다고 말한 선생님이 있지만 친구 없이도 행복하게 살 수 있는지 궁금해하며, 친구가 많지는 않지만 그래도 괜찮은지 묻고 있습니다. 이 질문은 관계에 관한 것으로, 우정의 가치, 자기 성취, 사회적 연결에 대한 통찰을 제공해야 합니다. 지원적인 답변을 제공하고, 균형 잡힌 관점을 제시하며, 자기 성취, 개인적 가치, 대체 관계(가족, 취미, 자기 계발)에 대해 이야기할 수 있습니다. 지침을 제공하고, 필요에 따라 추가 질문을 할 수도 있습니다. 친근한 말투를 사용하세요.
" \\\\
\textbf{\tinyayamulti{} Generation:} "선생님이right라고 하셨죠: 친구는 성장, 지원, 즐거움을 위한 귀중한 기회입니다. 하지만 “친구”가 당신의 삶에서 가장 중요한 관계일 필요는 없습니다. 친구 없이도 행복하고, 의미 있으며, 성취를 느낄 수 있습니다. 아래에는 이를 가능하게 하는 이유와 친구를 통한 행복을 위한 구체적인 방법을 간략히 설명한 몇 가지 아이디어가 제시되어 있습니다.

---

\#\# 1. **행복은 선택의 여지가 없습니다**

행복은 외부 조건에 따라 결정되는 것이 아닙니다. 우리는 모두 고유한 내적 자질, 가치관, 성향을 가지고 있으며, 이는 우리가 가장 큰 만족감을 느끼는 방식을 결정합니다. 어떤 사람들은 깊은 우정에서 위안을 찾지만, 다른 사람들은 다음과 같은 것에서 기쁨을 얻습니다:

- **개인적 관심사** (예: 음악, 예술, 과학, 스포츠)
- **창의적 표현** (글쓰기, 그림 그리기, 요리)
- **자기 계발 프로젝트** (독서, 코딩, 요가, 자원봉사)
- **신체 활동** (걷기, 요가, 춤, 운동)
- **마음챙김 및 정신 건강 관리** (일기 쓰기, 명상, 명상 앱)

당신이 가장 공감을 느끼는 “공간”이 무엇인지 파악하면, 친구를 얻지 못하더라도 그곳에서 행복을 얻을 수 있습니다.

---

\#\# 2. **대체 연결고리**

\#\#\# **가족 및 지역 사회**
- **가족** (부모, 형제자매, 조부모)은 역사, 사랑, 지원의 풍부한원을 제공합니다. 심지어 거리가 멀어도 정기적인 연락은 감정적 안전감을 줄 수 있습니다.
- **지역 사회** (자원봉사 단체, 수업, 클럽)는 목적의식과 소속감을 형성합니다. “동료”가 될 수 있지만 친구는 아닌 사람들과의 관계를 통해 관계를 구축할 수 있습니다.

\#\#\# **직업 및 학업**
- **동료**는 협업, 멘토링, 상호 책임감을 제공합니다. 심지어 비공식적인 대화도 통찰력을 제공할 수 있습니다.
- **지도자 또는 멘토**는 종종 깊은 공감과 성장에 대한 헌신을 가지고 있으며, 이는 우정과 유사한 관계를 형성할 수 있습니다.

\#\#\# **디지털 및 가상 관계**
- **온라인 커뮤니티** (서로 지원하거나 공유 관심사를 가진 그룹)는 친구와 유사한 역할을 할 수 있습니다.
- **가상 취미** (함께 게임을 하는 것, 온라인 수업, 원격 여행)는 물리적인 거리를 극복할 수 있습니다.

---

\#\# 3. **“친구”의 가치를 재평가해 보세요**

질문하신 선생님의 말을 생각해 보세요: *“우정이 중요하다고 하셨지만 친구 없이도 행복하게 살 수 있나요?”* 답은 **예입니다**. 다음 점을 고려해 보세요:

| 우정이 제공하는 것 | 대체로 충족되는 부분 |
|----------------------|-----------------------------|
| **감정적 지원** | 신뢰할 수 있는 가족 구성원, 치료사, 또는 지역 사회를 통해 |
| **시선과 검증** | 자기 계발을 통해 (독서, 취미, 개인적 목표) |
| **공감과 공유된 언어** | 동일한 취미, 활동, 또는 가치관을 가진 친구들을 통해 |
| ** growth and challenge** | 새로운 도전을 통해 (운동, 학습, 자원봉사) |

당신이 가장 필요로 하는 것이 *지원*이라면, 이는 친구를 통해 얻을 수도 있고 그렇지 않을 수도 있습니다. *흥미*와 *즐거움*이 필요하다면, 이는 예술, 음식, 여행, 코딩 등 다양한 분야에서 발견될 수 있습니다.

---

\#\# 4. **“친구”를 위한 작은 공간 만들기**

친구가 이제 더 많은 공간을 차지하더라도, 그 공간을 위한 **의식적인 공간**을 만들 수 있습니다:

1. **정기적인 “동료” 시간** – 매주 1시간씩 취미나 토론 그룹을 위해 시간을 내세요. 이 시간은 “친구”와 “동료”의 경계를 모호하게 만듭니다.
2. **소규모 그룹** – 소규모 (3~5명)의 고밀도 그룹은 더 깊은 유대를 형성하기 쉽습니다. 심지어 “비친구”와도 관계를 구축할 수 있습니다.
3. **수신 및 수신 균형** – 양쪽 이득이 되는 상호작용을 우선시하세요: 함께 학습하기, 협업 프로젝트, 또는 서로의 취미를 가르치기.
4. **경계 설정** – 모든 관계가 건강하게 유지되도록, 친구 관계에도 개인적 공간과 경계를 허용하세요.

---

\#\# 5. **행복에 대한 간단한 자기 점검**

다음 질문에 답해 보세요 (5분 소요):

1. **내 삶에서 가장 큰 기쁨을 주는 활동은 무엇인가요?**
2. **그 활동에서 가장 큰 성취를 느끼는 것은 무엇인가요?**
3. **그 활동에서 가장 큰 “친구”와 어떻게 연결할 수 있을까요?**
4. **그 활동에서 가장 큰 “동료”는 누구인가요?**
5. **그 활동에서 가장 큰 “비친구”는 누구인가요? (가족, 치료사, 지역 사회 등)**

당신이 답할 때마다, 친구를 대체할 수 있는 *내부* 자원을 발견하게 될 것입니다.

---

\#\# 6. **도움이 필요한 경우**

우정이 부족하다고 느끼거나 고립감을 느낀다면, 이는 일반적인 인간 경험입니다. 상담사, 치료사, 또는 신뢰할 수 있는 멘토와 대화하면, 이러한 감정을 탐구하고 대처 방법을 개발하는 데 도움이 되는 구조화된 공간을 제공할 수 있습니다.

---

\#\#\# **결론**

- **우정은 필수적이지만** 그 존재 여부와 상관없이 행복을 이룰 수 있습니다.
- **행복은 다양한 형태를 가질 수 있습니다**; 친구는 그 형태 중 하나일 뿐입니다.
- **당신의 가치를 반영하는 대체 관계와 취미를 발견하세요**.
- **그 관계에 작은 의식을 가하세요**, 친구 관계와 개인적 성장을 모두 키울 수 있습니다.

선생님의 조언을 가치 있게 여기고, 자신의 필요에 맞는 대체 “친구”를 찾아보세요. 가장 중요한 것은, 어떤 관계의 형태든 **당신을 완전히 지지하고, 성장시키며, 행복하게 하는** 관계라는 것입니다. 

특정 취미, 활동, 또는 관심사에 대해 더 알고 싶다면, 그 분야에서 행복을 위한 아이디어를 함께 모색해 보세요!
" \\\\
\textbf{Mean Score:} 6.75\\
\textbf{Accuracy:} 7\\
\textbf{Coherence:} 7\\
\textbf{Naturalness:} 6\\
\textbf{Instruction Following:} 7\\\\
\end{promptbox}
\end{center}

\end{document}